\documentclass[oneside,openright]{muthesis} 

\pdfoutput=1
\usepackage{enumerate}
\usepackage[chapter]{algorithm} 
\usepackage{algorithmic} 
\usepackage{amsmath, amssymb, mathtools} 
\usepackage{amsfonts}
\usepackage{nicefrac}
\usepackage{graphics}
\usepackage{graphicx}
\usepackage{wrapfig}
\usepackage{subfig}
\usepackage{caption}
\usepackage{paralist}
\usepackage{hyphenat}
\usepackage{multirow}
\usepackage{multicol}
\usepackage{rotating}
\usepackage{array}
\usepackage{booktabs} 

\usepackage{subfiles} 
\usepackage{setspace}

\begin{document}

\begin{titlepage}

\begin{center}

\vspace*{2.0cm}

\HRule \\[0.4cm]
{ \huge \bfseries Hybrid Evolutionary Computation for Continuous Optimization}\\[0.4cm]
\HRule \\[1.5cm]

\textsc{\Large Technical Memorandum} 2011-v.01\\[0.5cm]

School of Computer Science, University of Manchester, Kilburn Building, Oxford Road, MANCHESTER, M13 9PL

\vspace*{2.0cm}

\begin{minipage}{0.4\textwidth}
\begin{flushleft} \large
\emph{Author:}\\
Hassan A \textsc{Bashir}
\end{flushleft}
\end{minipage}
\begin{minipage}{0.4\textwidth}
\begin{flushright} \large
\emph{Supervisor:} \\
Dr.~Richard \textsc{Neville}
\end{flushright}
\end{minipage}

\vfill
{\large November 9, 2011}

\end{center}
This material represents the opinion of the author only and does not necessarily represent the opinion of the School or the University. While every attempt has been made to ensure the accuracy of this publication, neither the author, the School or the University can accept liability for mistakes that may exist. Please inform the author of any mistakes so they can be corrected in further printings.
\end{titlepage}


\newpage
\mbox{}
\pagebreak

\begin{abstract}
Hybrid optimization algorithms have gained popularity as it has become apparent there cannot be a universal optimization strategy which is globally more beneficial than any other. Despite their popularity, hybridization frameworks require more detailed categorization regarding: the nature of the problem domain, the constituent algorithms, the coupling schema and the intended area of application.

This report proposes a hybrid algorithm for solving small to large-scale continuous global optimization problems. It comprises evolutionary computation (EC) algorithms and a sequential quadratic programming (SQP) algorithm; combined in a collaborative portfolio. The SQP is a gradient based local search method. To optimize the individual contributions of the EC and SQP algorithms for the overall success of the proposed hybrid system, improvements were made in key features of these algorithms. The report proposes enhancements in: i) the evolutionary algorithm, ii) a new convergence detection mechanism was proposed; and iii) in the methods for evaluating the search directions and step sizes for the SQP local search algorithm.

The proposed hybrid design aim was to ensure that the two algorithms complement each other by exploring and exploiting the problem search space. Preliminary results justify that an adept hybridization of evolutionary algorithms with a suitable local search method, could yield a robust and efficient means of solving wide range of global optimization problems.

Finally, a discussion of the outcomes of the initial investigation and a review of the associated challenges and inherent limitations of the proposed method is presented to complete the investigation. The report highlights extensive research, particularly, some potential case studies and application areas.

\addcontentsline{toc}{chapter}{Abstract}

\end{abstract}

\listofalgorithms
\listoffigures
\listoftables
\tableofcontents




\chapter[Introduction]{Hybrid Evolutionary Computation for Optimization of Continuous Problems}
\label{IntroChap}

\section{Introduction}

As a \emph{vital} aspect for successful achievement of our everyday goals, optimization arises naturally in our daily lives. It deals with the task of selecting the \emph{best} out of the many possible decisions encountered in a typical real-life environment. For instance, searching for a shortest and/or fastest route to school or workplace is an everyday affair that requires dealing with optimization problem. A manufacturer seeking to boost production rate while cutting down the cost of production is also faced with an optimization task. In essence, optimization encompasses our routine need for maximizing gain, profit, quality, etc. or minimizing loss, cost, energy, time, etc. and it is as a result a prime facet of our everyday endeavour.

Of interest is the fact that optimization has over the years become a subject that is widely used in sciences, engineering, management and economics, and in the industry. This has led to the growing need for thorough understanding of optimization problems and their solution methods. As a research field in particular, optimization has been expanding in all directions at an astonishing rate during the last few decades and it has attracted extra attention from both academic and industrial communities. 

The recent growth in the development of new algorithmic and modelling techniques and in the theoretical background has largely led to the rapid diffusion of optimization into other disciplines. The striking emphasis on the interdisciplinary nature of the field has shifted it from being a mere tool in applied and computational mathematics to all areas of engineering, medicine, economics and other sciences. As pointed out by Yuqi He of Harvard University, a member of the US National Academy of Engineering:

\begin{quote}
\textit{``Optimization is a cornerstone for the development of civilization"} \cite{OptimTheoryPracticBookWen2006}.
\end{quote}

Formally, the subject is involved in determining optimal solutions for problems which are defined mathematically. It often requires the assessment of the problem's optimality conditions, model construction, building the algorithmic method of solution, establishment of convergence theories and series of experimentations with various categories of problems.

Although optimization problems include but not limited to the continuous problems, discrete or combinatorial problems, multiobjective optimization problems etc., the focus of this work was on the global linear/nonlinear continuous optimization problems potentially subject to some constraints and bounds. This was due to the fact that there are a great many applications from various domains that can be formulated as continuous optimization problems. For instance in:

\begin{itemize}
\item Controlling a chemical process or a mechanical device to optimize performance or meet standards of robustness;
\item Designing an investment portfolio to maximize expected return while maintaining an acceptable level of risk;
\item Finding an optimal trajectory for an aircraft or a robot arm;
\item Computing the optimal shape of an automobile or aircraft component in a manufacturing and process plant;
\item Scheduling tasks such as school time tabling or operations in manufacturing plants to maximize production level within the limited available resources while meeting the required quality standards and satisfying customer demands, etc.
\end{itemize}

Worth noting is that all these situations share the following three important aspects:
\begin{enumerate}
\item \textbf{Objective}: Also called an overall goal, it is a measure used to assess the extent to which the ultimate target in the activity is being realized and it is technically termed as the objective function which is typically modelled mathematically.
\item \textbf{Constraints}: This reflects the requirements within which the quest to optimizing the objective must be limited. It can be a limitation due to resource, time or space and or acceptable error levels or tolerance.
\item \textbf{Design variables}: This constitutes the set of all possible choices that must be made to ensure successful realization of the overall objective while satisfying the constraints. These implicit choices are technically referred to as decision or design variables and are the parameters around which the optimization task can be formulated. Obviously, any parameter that does not affect the objective or the constraints is not considered as a part of the design variables.
\end{enumerate}

Several solution techniques exist for the different types of the aforementioned optimization problems. However, the classical solution approach involves the use of numerical algorithms that have originated ever since the invention of the popular simplex algorithm for linear programming by Dantzig \cite{OptimTheoryPracticBookWen2006} in the late 1940s. Thereafter, many numerical algorithms such as gradient-based methods, conjugate gradient methods, Newton and quasi-Newton methods have evolved into powerful techniques for solving large scale nonlinear optimization problems. This category of algorithms is classified as exact or complete methods and constitutes the start-of-the-art approaches for solving various types of optimization problems in diverse fields. 

Subsequent to the development of the exact methods, a number of solution methods that are based on various \emph{heuristics} are developed. This category of algorithms also called approximate algorithms can be successfully applied to a wide range of optimization problems with little or no modifications in order to adapt to any specific problem. The term \emph{metaheuristics} originally coined by \cite{Blum2008paper88} is a generic term that was introduced to delineate a universal algorithmic framework designed to solve different optimization problems based on probabilistic decisions made during the search process. These approximate methods are usually easier to implement than their exact counterparts like the classical gradient-based algorithms. Although the approximate algorithms are mainly stochastic in nature, their main difference to pure random search is that randomness is guided in an intelligent and/or biased manner \cite{Blum2008paper88}. 

A large number of algorithms established on different theoretical paradigms and backgrounds such as the evolutionary computations (EC) like genetic algorithms (GA), genetic programming (GP) and evolutionary strategy (ES), simulated annealing, tabu search, ant colony optimization, artificial immune system, scatter search, estimation of distribution algorithms, multi-start and iterated local search algorithms, to mention a few, are typical examples of metaheuristics that fall into the category of approximate algorithms.

It has become evident that \cite{Pelikan2010paper79}, many real-world large scale optimization problems elude acceptable solutions via simple exact methods or even the approximate metaheuristics when applied independently. Therefore, in the recent years, researchers have become increasingly interested in the concepts that are not limited to the use of a single traditional algorithm, but combine various algorithmic ideas from different branches of artificial intelligence, operations research and computer science \cite{Blum2008paper88}. The combination of such algorithms is what is referred to as \emph{hybrid algorithms} or \emph{hybrid metaheuristics}. A skilful hybridization of algorithms is believed to provide a more flexible and efficient solution method that is suitable for large scale real-world problems. 

In fact, the need for hybrid algorithms surfaces and gains popularity after competing research communities have waived their traditional stance and believe in the invincibility of some classes of algorithms and philosophies that were regarded as generally the \emph{best}. It has become apparent that there cannot be a general optimization strategy which is globally better than any other. This argument was initially resolved---to \textit{some} degree---following the proposal by Wolpert and Macready of the well known \emph{no free lunch} (NFL) theorem \cite{WolpertMacreadyNFL1997}. The NFL theorem proved that on average over all possible functions\slash problems, the performance of all search\slash optimization methods that satisfy certain conditions is the same. Hence, as declared by \cite{Almeida2006paper86, Blum2010paper87} among others, the primary motivation behind the notion of hybridizing algorithms was to come up with robust systems that harness the benefits of the individual algorithms while discarding their inherent weaknesses.

\section{Motivations}
Despite the growing interest in the area of hybrid algorithms, more needs to be done to address matters of crucial importance vis-\`a-vis:
\begin{itemize}
\item Establishing a proper categorization of the hybrid strategies based on the expected precision or solution quality required for any given problem instance in the intended area of application.
\item Assessing based on the overall optimization goal the composition of the hybrid scheme as to whether it should comprise of algorithms from only approximate metaheuristics, exact algorithms or a mixture of the two.
\item Ascertaining when and why the identified approaches should be combined in an interleaved, paralleled or sequential manner. 
\item Enhancing the capabilities of the individual algorithms prior to hybridization, specifically focusing on the identified key features of the algorithms that are expected to play major roles in the hybridized system. 
\item Identifying at what stages of the solution process the key features of the algorithms can effectively be exploited to optimally benefit from the hybridization scheme. For instance, ensuring proper convergence assessment and maintenance of useful level of diversity at different stages of a typical EC algorithm. 
\item  Use of the well-known measures of problem difficulties \cite{Pelikan2010paper79} to judge the complexity of the problem categories upon which the hybrid algorithms are expected to be applied.
\item Developing hybrids that combine approximate algorithms with the state-of-the-art of exact optimization techniques like the sequential quadratic programming (SQP) algorithm\footnote{SQP is a gradient-based local search algorithm that is guaranteed to yield a solution for every finite size instance of constrained optimization problem in bounded time.}. This type of hybridization scheme is also called \emph{memetic algorithms} \cite{Fogarty1994paper52}. It is believed that although the approach can be very successful in practice, so far not much work exists in this direction \cite{Raidl2003paper92, ReevesRow2004GABookPrincplnPers}.
\end{itemize}

\section{Aims and Objectives}
The aim of this research was; to analyse and elucidate the current trend in hybridization of algorithms for optimization; to propose a novel hybrid optimization method that combines EC algorithm (for global searching) and an interior point method (IPM) based SQP algorithm (for local searching) to address large scale global optimization problems. And ultimately, to extend and apply the proposed system to deal with complex practical optimization problems such as dynamic optimization problems and control of feedback systems like PID tuning\footnote{PID stands for proportional-integral-derivative and PID controller is a generic control loop feedback mechanism that is widely used in industrial control systems. As one of the most commonly used feedback controllers, PID requires optimal tuning of its parameters.} in an efficient manner. 


The first objective of this report was to examine evolutionary computation algorithms (GA in specific) and their hybrids. We conduct an in-depth investigation on the parameterization aspect of EC algorithms. We investigate the effect of elitism in EC algorithms and propose a novel adaptive elitism method based on the overlapping population technique. We further analyse the convergence criteria of EC algorithms, model and quantify the individual contributions of the genetic operators involved in the evolution dynamics using extended Price's equation. We also demonstrate how to use the resulting Price's equation model as a criterion to measure and fully assess the convergence of the proposed EC algorithm. 

Secondly, we review local search optimization algorithms giving emphasis to quasi-Newton based numerical techniques. We investigate various methods for approximating and updating the Hessian matrix. We then extend the IPM based SQP algorithm \cite{XimingHassanSQPISDA} that uses BFGS Hessian approximation\footnote{In numerical optimization, the Broyden-Fletcher-Goldfarb-Shanno (BFGS) method permits approximation of the Hessian matrix  using rank-one updates specified by gradient evaluations (or approximate gradient evaluations) \cite{NocedalWright2006Book}.} to use exact Hessians so as to effectively solve complex constrained nonlinear optimization problems. 

The third objective was to study the technique of automatic differentiation (AD) for exact gradient and Hessian calculations. We investigate both the forward and reverse accumulation methods and then design an automatic differentiation tool based on the operator overloading principle. We implement the AD tool using object oriented design principle in Matlab environment. We then demonstrate how to adopt the AD tool to boost the capabilities of the proposed local search algorithm. 

Finally, we examine the current trends in the design and applications of hybridization methods for system optimization. And for the various techniques reported in the literature, we adopt the proposal in \cite{Almeida2006paper86} to devise a generalization that categorizes the hybrid systems based on the types of the combined algorithms and how they are combined in view of the overall optimization goal. We then design a hybrid system that combines the proposed global and local algorithms in a collaborative, batched and weakly-coupled manner with a built-in self-checking procedure for validation. The technique is hoped to ensure improvements not only in the efficiency\footnote{Efficient: The overhead as a result of the combination of the two algorithms will be minimized.}, but also in the robustness\footnote{Robust: It will ensure convergence to the optimal solution for wider range of problems with different levels of difficulties.} of the proposed hybrid system.

\section{Hypotheses}
In the following, we recast the aims of this research into the following hypotheses. Therefore, our overall objective is to verify the following:

\begin{description}
	\item[\#H1:] Hybrid global and local search methodologies provide good search strategies.
	\item[\#H2:] Specific types of local search algorithms (e.g. SQP/IPM)\cite{XimingHassanSQPISDA} are efficient in locating local optima.
	\item[\#H3:] Local optimization methods alone may not provide fast convergence to the global optimal solution.
	\item[\#H4:] Hybridization of global and local optimization algorithms should provide fast convergence to the optimal solution.
	\begin{description}
		\item[\#H4.1:] The global and local algorithms can serve as a means to validate each other's result.
	\end{description}
\end{description}

\section{Scope and Limitations}
The scope of this work was to deal with continuous optimization problems that are either local or global in nature. Hereby, the report is limited to problems that can be mathematically modelled in form of differentiable functions with at least second derivatives available.

\section{Chapter Organization and Summary}
Besides the introduction in this chapter, chapter 2 and 3 focus mainly on the principles and dynamics of evolutionary computation algorithms. Chapter 2 provides an in-depth review on the current trend and challenges militating against the development and simulation of evolutionary processes, particularly, the parameterization aspect of evolutionary algorithms in examined. This chapter lays out the evolutionary paradigm that will be used throughout this report. 

Chapter 3 further analyse the convergence characteristics of evolutionary algorithms and presents a fundamentally new way of perceiving the individual roles of evolutionary operators/processes towards the success of the evolution.  It then empirically analyse the efficacy of crossover in convergence detection in evolutionary computation. The chapter provides a foundation that will aid establishing a new hybrid strategy for the proposed system.

Chapter 4 investigates the framework of local optimization algorithms with particular emphasis on the gradient-based methods. The design of the sequential quadratic programming (SQP) algorithm and interior point method is investigated. The chapter then presents how an algorithmic approach for effective evaluation of derivatives could improve the convergence characteristics of the local search SQP algorithm.

In chapter 5, various techniques for hybridizing optimization algorithms are examined and a chronological taxonomy of various categories of hybrid algorithms is presented. The chapter presents a novel approach for hybridizing the EC algorithm with the SQP algorithm. A series of experiments undertaken to evaluate the proposed hybrid system are then analyzed.

Finally, Chapter 6 summarises the current work, discusses, in general, the outcome of our initial investigations. The chapter then concludes by pinpointing the open questions that will guide our further research in this direction.
\singlespacing
\bibliographystyle{ieeetr}






\chapter{Evolutionary Computation Algorithms--An Overview}
\label{ECAlgoChap}

In this part of the report, a review focusing on the foundation, development processes, mechanics and simulation of evolutionary computation (EC) algorithms will be provided. Details of parameterization aspect of EC will be investigated. Emphasis will be given on the rise in the earlier notion of standard parameter sets to the current trends of adaptive and dynamic systems that lead to the development of improved genetic algorithms. The chapter will conclude with the proposal of a novel adaptive elitism technique. 

\section{Introduction}
Evolution is a process that originated from the biologically inspired neo-Darwinian paradigm \cite{Fogel1997paper15} (i.e. the principle of survival of the fittest). It is believed to be a collection of stochastic processes that act on and within populations of species. These processes include reproduction, mutation, competition and selection \cite{Huxley1963paper15ref11}. In the late 1950s, evolution was understood as an optimization process that naturally shapes and maintains the balance in the existence and progress of individuals' life. As reported in \cite{Atmar1994paper17}, a salient rule of thumb of evolution as have come to be understood is that \emph{"Darwinian evolution is essentially an optimization technique. It is not a predictive theory, nor is it a tautology"}. Thus, as in most optimization processes, the solution point(s) are discovered via a trial and error search process. 

The far reaching impact of the idea of evolution has gone beyond the classical boundaries of biological thoughts. In what is termed as evolutionary computation (EC), the process of evolution has now become an optimization tool that can be simulated and applied in solving complex engineering problems. 

Evolutionary computation algorithms are designed to mimic the intrinsic mechanisms of natural evolution and progressively yield improved solutions to a wide range of optimization problems. This is evident because, the success of these algorithms is always not directly inclined to the domain knowledge specific to any problem. 

The three popular evolutionary computation algorithms that stand out are \emph{genetic algorithm} (GA), \emph{evolutionary strategies} (ES) and \emph{evolutionary programming} (EP). These techniques are all built around the common principles of natural evolution and rose almost independently of each other. They are strongly interrelated and differ mainly in the data structures used to represent individual solutions, the types of genetic alterations on current individuals to create new ones (i.e. genetic operations such as reproduction and mutation) and the techniques for selection after competition. 

In the original implementation of genetic algorithms, their data structure enforces representation of candidate solutions as binary vectors. In their distinct nature, evolutionary programming algorithms use finite state machines for representing candidate solutions, whereas in evolutionary strategies solution points are directly represented as real valued vectors. With the growing interrelations among these techniques, their minor differences blurred especially with regards to the choice of data structure and genetic operators. Recently, a number of experimental results have shown that \cite{Raidl2005paper89, Atmar1994paper17}, problem dependent representation of candidate solutions can significantly improve the effectiveness of the overall optimization process thereby avoiding the problem of mapping between various representations. 

As highlighted in the previous chapter, the aim of this work was to use genetic algorithms as global optimization method. Thus, subsequent treatment of the evolutionary computation literature will focus mainly on the evolution principles of genetic algorithms.

\section[Background and Process Dynamics of Evolutionary Computations]{\nohyphens{Background and Process Dynamics of\\ Evolutionary Computation}}
Genetic algorithms are evolutionary based algorithms originally inspired by Holland in the 1970s and the principles of which is extensively disseminated in his book \textit{Adaptation in Natural and Artificial Systems} \cite{Holland1975}. Although Holland's contribution to the development of the original ideas has been quite remarkable, history has shown that quite a number of researchers working on the same area have also contributed immensely in the design and development of these techniques. In late 1960s, an independent work by Schewefel and Rechenberg \cite{ReevesRow2004GABookPrincplnPers} led to their proposal of the technique of evolutionary strategies. Parallel to that Fogel \cite{Fogel1963GABookref81, Fogel1966GABookref82} and his colleagues implemented the idea of evolutionary programming which also is based on natural evolution principles. Hitherto the work of Goldberg \cite{Mitsuo1997GAnEngDesingBook} who researched and extensively outlined the typical form of the genetic algorithm used today, prior proposals were mainly mutation and selection based without incorporation of the recombination operator. Detailed historical background on genetic algorithms can be found in the excellent collection by David Fogel \cite{FogelGABookref79}. 

Genetic algorithms have proven to provide a heuristic means of solving complex optimization problems that require a robust solution method. Recently, they have been successfully applied in the areas of computing and industrial engineering such as vehicle routing \cite{Wan1999paper49}, scheduling and sequencing \cite{Man1999paper59}, network design and synthesis \cite{Queiroz2009paper43, Grimbleby1999paper55}, reliability design \cite{Bavafa2009paper37}, facility layout and location \cite{Xingbo2008paper32}, to mention a few.

\begin{figure}[hbtp]
	\centering
	\includegraphics[scale=0.40]{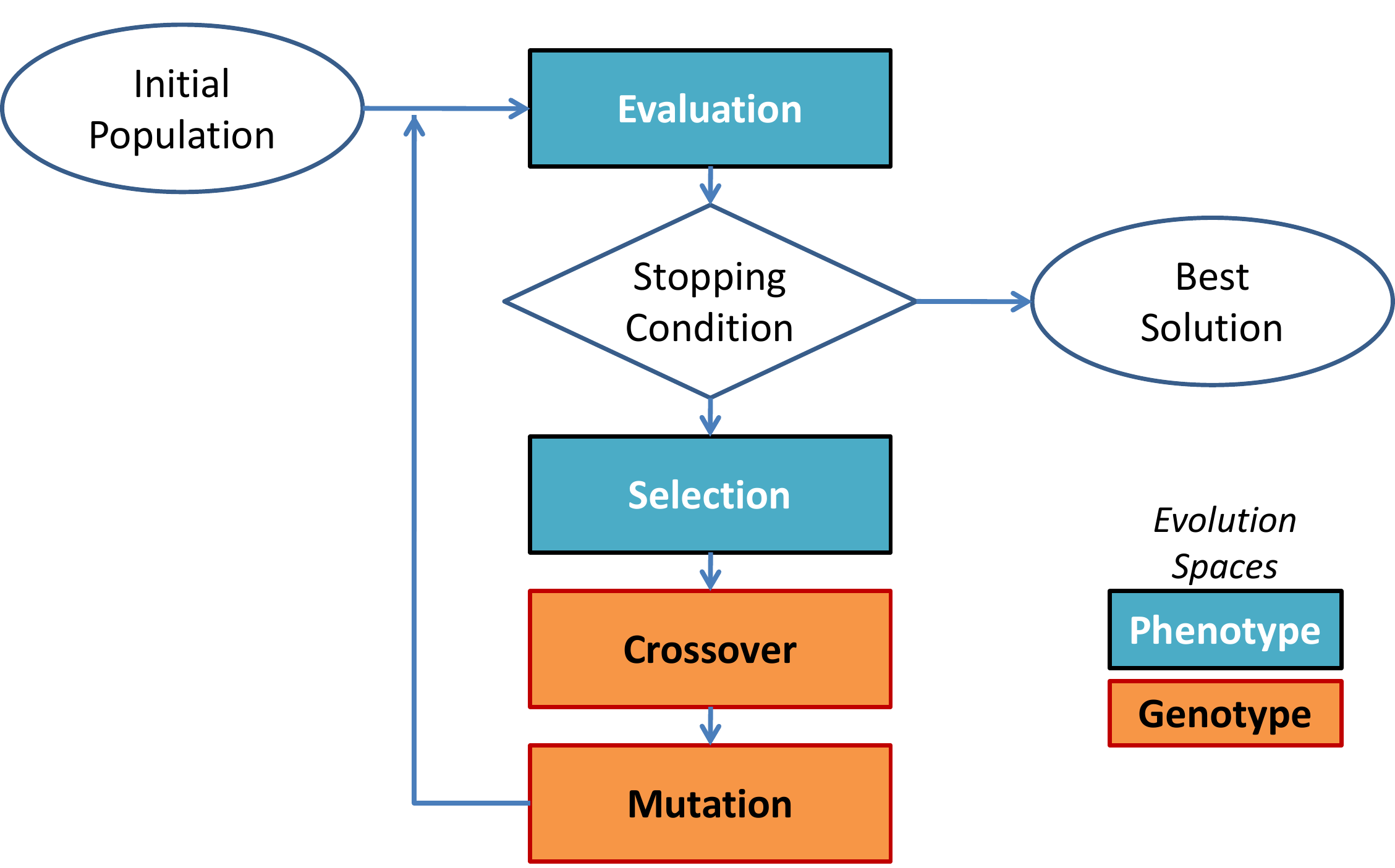}
	\caption{Flowchart of a Typical Genetic Algorithm}
	\label{GAFlowchart}
\end{figure}

Contrary to the traditional optimization methods, as depicted by figure \ref{GAFlowchart}, genetic algorithm is an iterative procedure that starts with an initial fixed set or pool of candidate solutions called \emph{population}. A candidate solution point is called an \emph{individual} and represents a possible solution to the problem under consideration. An individual is represented by a computational data structure called a \emph{chromosome}. Usually, a chromosome is encoded as a string of symbols of finite-length called \emph{genes}. The possible values a gene can take correspond to the \emph{allele}. As proclaimed by \cite{Pengfei2010paper6, GoldbergHolland1988}, a chromosome can be a binary bit string or any otherwise representation. 

The chromosomes in the initial population are usually created randomly or via a simple heuristic construction. During each iteration step, called a \emph{generation}, a stochastic selection process is applied on the initial population to choose better solutions following an evaluation that is based on some measures of \emph{fitness}. Chromosomes that survive through the selection process constitute a new set called \emph{parents} and are qualified to take part in the remaining stages of the evolution process. 

In order to explore other areas of the search space, the parent chromosomes undergo recombination and/or mutation operations and generate a new set of chromosomes called \emph{offspring}. The recombination entails exchange of characteristics by merging two parent chromosomes using a \emph{crossover} operator, while mutation operation is a genetic alteration of a randomly chosen parent chromosome by a \emph{mutation} operator.

A new generation of chromosomes is then formed by selecting from either the combined pool of parents and offspring or the offspring pool based on a prescribed fitness measure. Fitter chromosomes have higher chances of being selected and the average fitness of the population is expected to grow with successive generations. The process continues until a termination criterion is met or it ultimately converges to the best chromosome which hopefully represents the optimum or suboptimal solution to the problem. Notice how figure \ref{GAFlowchart} categorizes the key GA components on the basis of the evolution space they operate, details on this will be given in  section \ref{SimEvolutionSec}.

\subsection{A Generalized Model for Genetic Algorithm}

Based on the foregoing discussion on GA dynamics, without loss of generality, the evolution processes involved in a typical genetic algorithm can be modelled as shown in Algorithm \ref{Algorithm1}. For any generation $t$, the parameters $P(t),\: Q_{s}(t),\: Q_{r}(t)$  and $Q_{m}(t)$ used in this algorithm respectively represent the population at the initial generation, at the end of selection, and after recombination and mutation operations.

\begin{algorithm}
\caption{A Canonical Model of Genetic Algorithm}
\label{Algorithm1}
\begin{algorithmic}
\STATE $\textbf{begin}$\\
	$\qquad$ $t\leftarrow0;$\\
	$\qquad$ initialize $P(t);$\\
	$\qquad$ evaluate $P(t);$\\
	$\qquad$ $\textbf{while}$ $\textit{not termination}$ $\textbf{do}$\\
		\STATE $\qquad \qquad  Q_{s}(t)\leftarrow$ select $P(t);$\\
		$\qquad \qquad  Q_{r}(t)\leftarrow$ recombine $Q_{s}(t);$\\
		$\qquad \qquad  Q_{m}(t)\leftarrow$ mutate $Q_{r}(t);$\\
		$\qquad \qquad$  evaluate $Q_{m}(t);$\\
		$\qquad \qquad  P(t+1)\leftarrow$ select $Q_{m}(t)\cup P_{m}(t);$\\
		$\qquad \qquad  t\leftarrow t+1;$\\
	$\qquad$ $\textbf{end while}$\\
$\textbf{end}$\\
\end{algorithmic}
\end{algorithm}

Because of their simple and stochastic nature, GAs require only the evaluation of the objective function but not its gradients. Such a derivative-free nature relieved GAs of the computational burden of evaluating derivatives especially when dealing with complex objective functions where derivatives are difficult to compute. The randomness in GAs improves their versatility in escaping the trap of suboptimal solution which is the major drawback of gradient based optimization techniques. Goldberg \cite{Goldberg1989Book} summarises the following key features of GAs that made them robust optimization search methods. 

\begin{itemize}
\item Genetic algorithms search from a population of solutions, not a single solution;
\item The genetic operations (i.e. recombination and mutation) work on the encoded solution set, not the solution themselves;
\item The evolution operation (i.e. selection) uses a fitness measure rather than derivative or other auxiliary knowledge;
\item The progress of the process relies on probabilistic transition rules, not deterministic rules.
\end{itemize}

\section[Simulation of Evolution: Phenotype and Genotype Spaces]{Simulation of Evolution: Phenotype and\\ Genotype Spaces}
\label{SimEvolutionSec}
\begin{figure}[hbtp]
	\centering
	\includegraphics[scale=0.50]{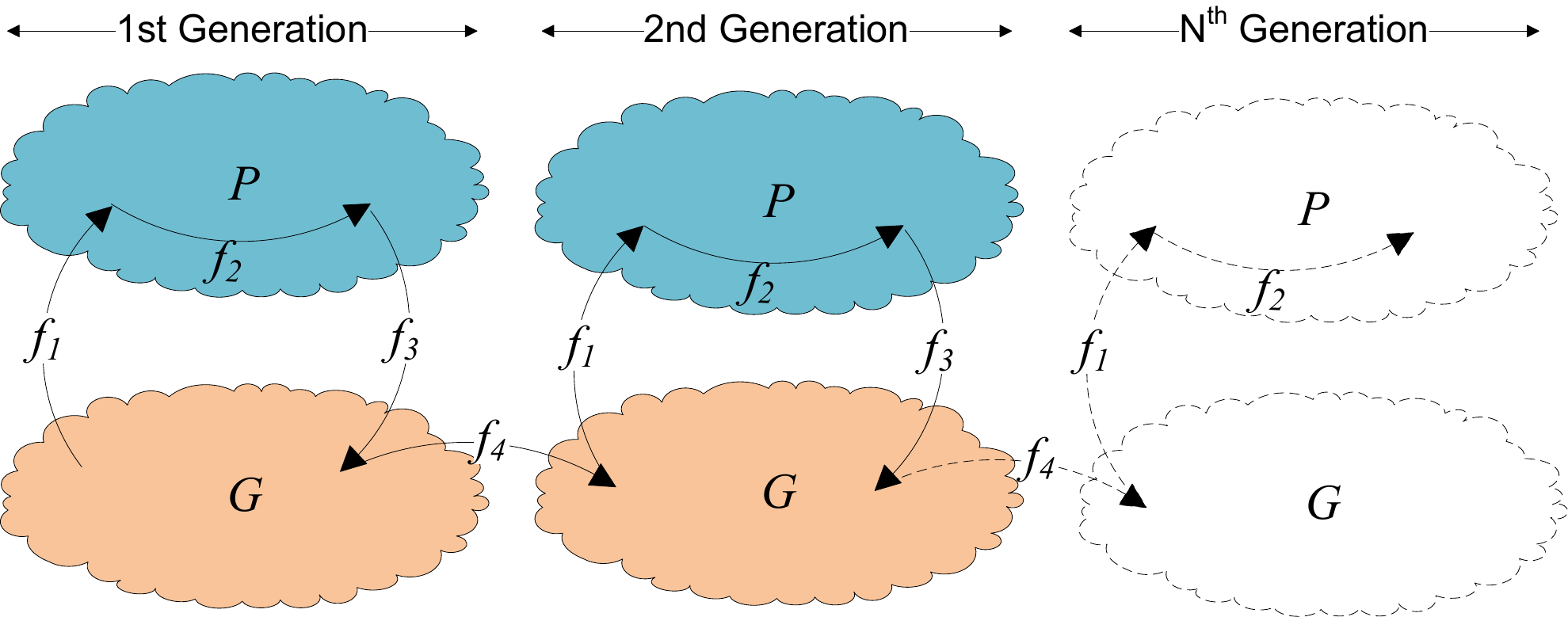}
	\caption{Simulation of Genotype-Phenotype Mapping Functions}
	\label{GAMappingSimChart}
\end{figure}

In spite of the simplicity in the informational physics of the processes governing evolutionary system, it has always been an area of misunderstanding to clearly delineate which part of the evolution occurs at what space. Atmar \cite{Atmar1994paper17} argued that it is only possible to adopt and successfully simulate the process of natural evolution for engineering purposes if the physics of evolution is well understood and the sequence of causation is represented appropriately. Formally, evolutionary system inherently runs in two distinct spaces: \emph{phenotypic} and \emph{genotypic} spaces \cite{Lewontin1974Book}. The phenotype space $\mathcal{P}$  represents the behavioural or physical characteristics of an individual or chromosome, whereas the genotype space $\mathcal{G}$  is the encoding space and represents the exact genetic makeup of a chromosome. Figure \ref{GAMappingSimChart} shows a simple simulation of evolution processes within and across generations depicting various genotype-phenotype mapping functions. 

Very often, simulated evolution mimic the natural evolution by creating the initial population as a set of chromosomes encoded in a genotype space. Thus, the process usually begins in a genotype space with $g_{i}\in \mathcal{G}$  population that evolve over generations and finally ends with a solution set $p_{i}\in \mathcal{P}$  in a phenotype space as described in the following steps. 

\begin{enumerate}[i)]
\item The first mapping function $f_{1}:g\rightarrow p$  decodes from $\mathcal{G}$  to $\mathcal{P}$  space such that each $g_{i}$  is translated into phenotype $p_{i}$  and get evaluated. Thus, the function shifts the evolution from the genotype to the phenotype space. 
\item The second mapping function $f_{2}:p\rightarrow p$  describes the selection operation. It is the process of choosing individuals for reproduction and it occurs entirely in the phenotypic space.
\item The third mapping function $f_{3}:p\rightarrow g$  describes the genotypic representation. It is the process of encoding the genotype prior to reproduction and shifts the evolution back to genotype space.
\item The fourth mapping function $f_{4}:g\rightarrow g$  is the reproduction function. It is where the variation operations such as recombination and mutation take place. It incorporates the rules of random and directed coding alteration during the reproduction process. This process entirely happens in the genotype space and it is where the transition from current generation to the next occurs.
\end{enumerate}

Lewontin \cite{Lewontin1974Book} stressed that although the distinction between evolutionary spaces $\mathcal{P}$ and $\mathcal{G}$ is sometimes \emph{illusory}, it is very important to clearly understand which part of the evolutionary process take place at which state space. Failure to clearly delineate the two spaces led to the confusion that surrounds the theory of evolutionary dynamics.

\section{Initialization of Evolutionary Computation Algorithms}
A number of questions need to be answered in order to properly set up an EC model for any optimization problem under consideration. Of primary importance are the choices of suitable data structure (i.e. chromosome representation and encoding), the method of creating the initial population and its size. In the following we will consider binary encoding in EC and its mapping method between the genotype and phenotype spaces. 

\subsection[Representation in Evolutionary Algorithms]{Representation in Evolutionary Computations}

Defining a proper representation scheme for an EC algorithm is crucial to its overall performance with regard to efficiency and robustness. Holland's original idea \cite{Holland1975} advocates the use of binary representation and was based on the motive of ensuring the genetic processes operate in a domain (i.e. space) that is distinct from that of the original problem. This will ultimately enhance the robustness of evolutionary algorithms by making them more problem-independent. Furthermore, binary representation can ease the task of design and implementation of the major evolutionary reproduction operators.

As a crucial part in EC algorithms, \cite{Boyabaltli2007paper95} categorizes the parameterization aspect of EC algorithms into two groups: \textit{structural} and \textit{numerical} and argue that representation constitute the major part of the structural group. The following section will show how the mapping between a genotypic space (encoded in binary) and a phenotypic space (in real-valued) can be achieved with any level of precision.

\subsection{Real-Binary Encoding and Mapping functions}
Since our goal is to design a hybrid EC algorithm for solving continuous linear/nonlinear problems where in most cases the design variables are usually real-valued, integer or mixture of the two, a real-coded binary representation will make a good choice. Consider the problem shown in equation \eqref{SchwefelFunction} \cite{Salomon1996paper109}. It is a continuous optimization problem that requires maximizing the function over the search space $v=\left\{ x\in\mathbb{R}: x=\left\{ -5.0,\:5.0\right\} \right\} $. The optimal solution $x^{*}$ is a real number in the range $\left[-5.0,\:5.0\right]$. This problem can adequately be encoded if the \textit{range} of the design variables and the \textit{precision} requirement is known.

\begin{equation}
\label{SchwefelFunction}
f(x)=\sum_{i=1}^{n}x_{i}\sin\left(\sqrt{\left|x_{i}\right|}\right);\quad-5.0\leq x_{i}\leq5.0;\quad n=1,2,... 
\end{equation}

The first step is to encode the problem domain data from the phenotype space into a sensible formulation for the EC process (i.e. the genotype space). For any variable $a_{i}\leq x_{i}\leq b_{i}:a,b\in\mathbb{R}$, assuming the precision requirement $p$ is two places after decimal, i.e. $p=10^{-2}$, then, length for the binary bits $l_{i}$ required to map the real variable $x_{i}$ into a corresponding binary variable $x'_{i}$ can be derived from:

\begin{equation}
\label{bitLengthEqn}
2^{l_{i}-1}\leq(b_{i}-a_{i})\times\frac{1}{p}\leq2^{l_{i}}: l_{i}\in\mathbb{N}
\end{equation}
Thus, for the problem in \eqref{SchwefelFunction}, we have:
\[
2^{l_{i}-1}\leq(5-(-5))\times\frac{1}{10^{-2}}\leq2^{l_{i}}
\]
\[
2^{l_{i}-1}\leq1000\leq2^{l_{i}}
\]
\[
2^{9}\leq1000\leq2^{10}
\]
Hence, the required bit length for the variable $x'_{i}$ is $l_{i}=10$.

Now, for any multidimensional function having $x_{1},x_{2},...,x_{n}$ real variables, if each of these variables is mapped to its corresponding binary variable $x'_{1},x'_{2},...,x'_{n}$, then, for a population of size $N$, every individual binary chromosome $\tilde{x}_{k}$ is obtained by concatenating all the binary variables as follows:
\begin{equation}
\tilde{x}_{k}=x'_{1}\mid x'_{2}\mid...\mid x'_{n}\::\: k=1,2,...,N
\end{equation}
Hence, the length of the resulting binary chromosome $\tilde{x}_{k}$ is $L$ which is equal to the sum of the bit length $l_{i}$ of all the $n$ binary variables $x'_{i}$, such that
\begin{equation}
L=\sum_{i=1}^{n}l_{i}.
\end{equation}

It is worth mentioning at this point that the precision requirement for the decision variables may differ from one variable to another within a given problem. Thus, in a general case, if a decision variable is defined in the range $a_{i}\leq x_{i}\leq b_{i}:a,b\in\mathbb{R},$ in order to map it to a binary string of length $l_{i}$, the precision is:

\begin{equation}
p_{i}=\frac{b_{i}-a_{i}}{2^{l_{i}}-1}.
\end{equation}

Having successfully encoded the problem into binary (i.e. the genotype space), decoding chromosomes back to the phenotype space is a reverse process and it is necessary for evaluating their fitness before selection. This process entails the following two steps:

\begin{description}
\item [{First:}] Decomposing the binary chromosome $\tilde{x}$ into its constituent binary variables $x'_{i}$. This requires splitting the $L$ bits of $\tilde{x}$ into a chunk of $l_{i}$ bits corresponding to the $x'_{i}$ binary variables. Then, the corresponding real variables $x{}_{i}$ are derived via binary to decimal transformation of the $l_{i}$ bits of $x'_{i}$, such that:

\begin{equation}
x_{i}=\sum_{j=1}^{l_{i}}b_{j}2^{l_{i}-j}:i=1,2,...,n.
\end{equation}

where $b_{j}$ are the binary bits of $x'_{i}$, $l_{i}$ is its length and $n$ is the total number of these variables.

\item [{Second:}] Mapping the obtained real variables $x{}_{i}$ to conform to their originally defined ranges $a_{i}\leq x_{i}\leq b_{i}$, such that:
\begin{equation}
\label{MappingForVarRange}
x_{i}=a_{i}+\frac{b_{i}-a_{i}}{2^{l_{i}}-1}x'_{i}:\forall i=1,2,...,n
\end{equation}
\end{description}

Hence, for the problem under consideration \eqref{SchwefelFunction}, we have
\[
x_{i}=-5.0+\frac{5.0-(-5.0)}{2^{10}-1}x'_{i}.
\]
\[
x_{i}=-5.0+0.009775x'_{i}.
\]

\subsubsection*{Example}
Supposing for the problem in \eqref{SchwefelFunction}, the value of the first variable is $x{}_{1}=3.0$, and its precision requirement $p{}_{1}=10^{-2}$, then, based on equation \eqref{bitLengthEqn}, the required bit length to represent $(-5.0\leq x_{1}\leq5.0:x{}_{1}=3.0)$ in binary is determined to be $l{}_{1}=10$. Hence, the 10 bits binary equivalent of $x{}_{1}$ is:
\[
x'_{1}=0000000011.
\]
Suppose after the parent chromosome of $x'_{1}$ undergoes genetic reproduction (i.e. crossover and mutation), the value of $x'_{1}$ get transformed to $x'_{1}=1011000010$. Then, in order to derive the corresponding phenotype value, this is first converted to a decimal value $x_{1}=706$, and then mapped to its prescribed domain/range according to equation \eqref{MappingForVarRange} to obtain the true phenotype value of $x_{1}$ as follows:
\[
x_{1}=-5.0+0.009775x'_{i}=-5.0+0.009775\times706=-4.37
\]
\begin{flushright}
$\Box$
\end{flushright}

Although this may seldom happen, situations arise where binary representation is not only promising but is also the natural choice. The \textit{knapsack} problem in operations research is a typical example. The 0-1 knapsack problem consists of a set of $n$ items to be packed into a knapsack of size $K$ units. If each item has a weight $w_{i}$ and is of size $k_{i}$ units, then the goal is to maximize the weight for a given subset $I$ of the items such that:

\[\max\sum_{i\in I}w_{i}:\:\sum_{i\in I}k_{i}\leq K.\] 

Reeves et al. \cite{ReevesRow2004GABookPrincplnPers} have shown that the knapsack problem can be reformulated as an integer programming problem and a solution can be represented as a binary string of length $n$. In such case, there will be no distinction between the genotype and the phenotype and thus completely eliminating the need for mapping functions. 

In the past, the general view in the EC community regarding problem's data structure and the choice of suitable EC algorithm was to match the problem to an appropriate EC algorithm. Evolutionary strategies were designed based on real valued representation and are therefore used for continuous problems. Genetic algorithms were primarily designed for discrete optimization and thus originally use binary representation as a norm. However, many researchers \cite{Blum2008paper88, Raidl2005paper89, LanceChambers1995Book} have pointed out that this is not the case at the moment as every one of these algorithms is been successfully used with all kinds of representations for various optimization problems.

\subsection{Other representations in the Literature}
Many types of representations for genetic algorithms are echoed in the literature for different problem domains. Special cases arise where the binary representation is inadequate or even unsuitable for the problem under investigation. Greenhalgh et al. \cite{Greenhalgh2000paper12} argue that although Goldberg's \cite{Goldberg1989Book} notion of implicit parallelism in genetic processing favours binary representation, practitioners report better performance with non-binary representations in many applications \cite{Davis1989}. Rees et al. \cite{Rees1999GABookref212} extend their results from binary to alphabets of cardinality of powers of $2$ (i.e. $2^{x}$) and uphold the use of higher cardinality representation by deriving an upper bound for the required number of iterations for such higher cardinality GAs to visit all individuals in a population. 

Thus, in some situations, use of problem dependent representations is necessary. For instance, the \textit{rotor stacking} problem originally described by \cite{McKee1987GABookref162} is a typical discrete non-binary problem that requires higher cardinality ($k$-ary) to be properly represented. For a set of $n$ rotors having $q$ holes to be stacked, a straightforward representation is to create a candidate solution with a fixed length of $n$-bits with cardinality $q$-ary corresponding to the number of holes in a rotor. This means that for a problem of stacking $10$ rotors each having $3$ holes, a candidate solution is a string of length $10$ that is made from a $3-$ary dataset. It is interesting to note that for $q=2$, rotor stacking problem reduces to binary and therefore, binary representation would be the best choice. 

Very often, apparent representation schemes exist that can best suit the problem to be modelled. Optimization of permutation problems is a typical example where there is a natural choice for representation. Here, the representation can directly be defined over the range of all the possible permutations. A typical example is the work on flowshop sequencing scheduling problem by \cite{Reeves1995paper7}. Flowshop sequencing is a permutation problem in which $n$ jobs are to be processed on $m$  machines over a certain time limit. The objective was to find the permutation of jobs that will minimize the total time required to complete all the jobs (i.e. the makespan). For any job $i$ on machine $j$ with a job permutation set $\left\{ J_{1},\, J_{2},\:...,\: J_{n}\right\}$, instead of developing a complex representation, the authors directly represent the schedule as a $k-$ary integer problem and develop suitable genetic operators that will ensure generation of feasible solutions. 

In a similar approach, Man et al. \cite{Man1999paper59} proposed a non-binary representation for a combinatorial optimization problem of scheduling partially ordered tasks in a multiple processor environment. The goal was to schedule an optimal execution of $\tau_{i}\in T$  set of tasks with each requiring a duration $d_{i}$  on a set of $p_{i}\in P$  processors. Each processor can only execute one task at a time and the entire problem is subject to a set of temporal ordering constraints $\mathcal{O}$. The authors developed an interesting problem specific representation that best suit the problem's requirement and use specialized genetic operators for the reproduction operation. 

Elsewhere, Chambers \cite{LanceChambers1995Book} proposed a generalized model for scheduling problems in which a scheduling strategy is parameterized and used in matching various loop characteristics to system environment. The various parameters for the generalized loop scheduling strategy are concatenated into a binary chromosome representing a candidate solution for the problem. The binary representations are decoded into integer values during the simulation process. The author argued that since GA does not impose any specific rules in encoding chromosomes, the quality of the resultant solutions does not depend on the arrangement of the parameters within the chromosomes. 

As noted by \cite{Queiroz2009paper43}, majority of the researches on network configuration and distribution systems have adopted direct representation for the state of the network \cite{Zhu2002paper43ref13, Sivanagaraju2006paper43ref14, Mendoza2006paper43ref15}. This entails setting the bits of the chromosomes to the status of the switches (i.e. open or close state) in the network with each chromosome having a length equal to the number of the switches. The advantage is that no extra decoding task is required as the design made the genotype to map directly to the phenotype. However, with this representation, genetic operators often yield infeasible solutions which require a repair mechanism. This adds a computational overhead and intuitively offsets the paradigm of natural evolution. 

Realizing the critical role of chromosome representation in the overall success of GA, Queiroz et al. \cite{Queiroz2009paper43} suggest a tree-like representation for the network reconfiguration problem of finding a topology that will minimize technical losses throughout a given planning period. They adopted the so-called network random keys (NRK) \cite{Rothlauf2002paper43ref17} representation for minimum spanning tree problem. NRK is an arc-based representation that can exploit the sparsity in the distribution networks graphs. The representation defines chromosomes to be of length equal to the number of arcs in the network and to consist of integer weights corresponding to each arc. The authors used minimum spanning tree algorithm to map the chromosome into a tree for evaluation and argued that the proposed design ensures production of feasible solution by crossover and mutation operators. Although the formulation yields a suitable data structure for the evolution operators in genetic algorithm, the authors admit that it leads to a larger optimization problems than was needed to identify the best configuration for the fixed demands due to apparent increase in the number of variables. 

Worth mentioning at this point is the assertion that designing representation schemes that easily map the genotype to phenotype is very essential as it limits the overhead caused by complex mapping functions \cite{Back1997paper16}. Very often, complex encoding functions tend to introduce additional nonlinearities, discontinuities and multimodalities to the optimization problem. This can hinder the search process substantially thereby making the combined objective function more complex than that of the original problem. 

Elsewhere, Radcliff et al. \cite{Fogarty1994paper52} introduced the concept of allelic-representation and described how it distinctively differs from the traditional genetic representation. The authors present formalizations for both the genetic and allelic representations and use it to model a typical travelling salesman problem (TSP). They argue that unlike the former, the latter representation can always yield feasible solutions following the action genetic operations. For a search space $\mathcal{P}$ (of phenotypes) and a representation space $\mathcal{G}$ (of genotypes), given any solution in $\mathcal{P}$, a representation function $\rho(\mathcal{P},\mathcal{G})$:
\begin{equation}
\rho:\:\mathcal{{P}}\rightarrow\mathcal{G}
\end{equation}
returns the chromosome in $\mathcal{G}$ that represents it. The representation function is injective such that there is a one-to-one \textit{mapping} between any solution point in the phenotype $\mathcal{P}$ to any chromosome represented in the genotype $\mathcal{G}$. In the context of genetic representation, a formal allele is formulated as an ordered pair consisting of a gene and one of its possible values, such that a chromosome:

\[\chi=\left(x_{1},x_{2},...,x_{n}\right)\] 
has alleles $\left(1,\, x_{1}\right),\left(2,\, x_{2}\right),...,\left(n,\, x_{n}\right)$. Thus, in an allelic representation, instead of being a vector, a chromosome is a set whose elements are drawn from some universal set $\mathcal{A}$. 

A rather recent application of GA on feature selection problem by \cite{IlSeok2004paper56} demonstrates how binary representation can be used to appropriately represent chromosomes. The author noted that feature selection problems have exponential search space making genetic algorithms the natural choice for their optimization. A string of binary digits is used to represent a feature with values of $1$ and $0$ indicating a selected and removed feature respectively. Thus, a chromosome $D$  has a length equal to the size of the feature set and each gene in the chromosome carries the status of the feature. A chromosome $01001010$ has the second, fifth and seventh features selected with the rest turned off. Controlled mutation and crossover operators were used to ensure generation of feasible chromosomes. 

Nevertheless, a common point of consensus in the field of hybrid metaheuristics is that use of problem specific representation is viewed by many \cite{Blum2008paper88, Raidl2005paper89, IlSeok2004paper56, Raidl2003paper92} as an act of hybridizing genetic algorithms. This kind of hybridization is believed to crucially affect the performance of the EC algorithm. Details on hybridization techniques will be presented in later sections. 

Table \ref{GARepresentationMethods} compares and summarises the various representation techniques reviewed from various domains of the evolutionary computation. It highlights the cases when phenotype-genotype mapping functions is necessary and pinpoints a suitable category of the genetic operators to adopt.

\begin{table}[htb]
\begin{center}
\caption[Representation methods in Genetic Algorithms]{A Summary of Commonly used Representation Methods in Genetic Algorithms}
\label{GARepresentationMethods}
\begin{tabular}{| p{68pt} | p{75pt} | p{95pt} | >{\centering}p{62pt} | p{50pt}| }
\hline

\multirow{3}{*}{\textbf{Category}} & \textbf{Example \mbox{Problems}}  & \textbf{Representation Type}  & \textbf{Phenotype-Genotype mapping} &  \textbf{\mbox{Genetic} \mbox{Operators} Type}\\
\hline

\multirow{7}{68pt}{\textbf{Discrete \mbox{Binary}}}   & Knapsack \;problem \cite{ReevesRow2004GABookPrincplnPers}      & \multirow{2}{95pt}{Direct binary}   & \multirow{2}{*}{No}  & \multirow{2}{*}{Generic}\\ \cline{2-5}
     &  Network \;distribution \cite{Zhu2002paper43ref13, Sivanagaraju2006paper43ref14, Mendoza2006paper43ref15}  & \multirow{3}{95pt}{Direct binary}   & \multirow{3}{*}{No}  & \multirow{3}{*}{Generic}\\  \cline{2-5}
     & Feature \;selection  \cite{IlSeok2004paper56}    & \multirow{2}{95pt}{Direct binary}   & \multirow{2}{*}{No}  & \multirow{2}{*}{Generic}\\ \hline

\multirow{6}{68pt}{\textbf{Discrete \mbox{Non-Binary}}} & \multirow{3}{75pt}{Rotor stacking \cite{McKee1987GABookref162}} & Higher cardinality $q-$ary representation, $q>2$  & \multirow{3}{*}{Yes}  & \multirow{3}{50pt}{Problem  \mbox{specific}}\\  \cline{2-5}
 & Network \mbox{reconfiguration} \cite{Queiroz2009paper43} &  NRK, using minimum spanning tree \mbox{algorithm}  & \multirow{3}{*}{Yes} & \multirow{3}{50pt}{Problem \;specific}\\ \hline

\multirow{4}{*}{\textbf{Permutation}}         & Flowshop \mbox{sequencing \cite{Reeves1995paper7}} & Direct integer, range of permutation & \multirow{2}{*}{No}  & Problem \;specific\\ \cline{2-5}
  & TSP problem \cite{Fogarty1994paper52} & Allelic, using \mbox{ordered} pairs  & \multirow{2}{*}{Yes} & Problem \;specific\\ \hline

\multirow{5}{*}{\textbf{Combinatorial}} & Process scheduling problem \cite{Man1999paper59} & \multirow{2}{95pt}{Direct integer}    & \multirow{2}{*}{No}  & Problem \;specific\\ \cline{2-5}
                    &  Parameterized scheduling \mbox{strategy \cite{LanceChambers1995Book}} & \multirow{3}{95pt}{Binary encoding}   & \multirow{3}{*}{Yes} & \multirow{3}{*}{Generic}\\ \hline

\multirow{3}{68pt}{\textbf{Non-discrete}} &  Continues              & Real encoding     & No  & Generic\\ \cline{3-5}  
                                       &  linear/nonlinear      &  Binary encoding  & Yes & Generic\\ \cline{3-5}  
                                       &  problems              & Gray code         & Yes & Generic\\  
\hline
\end{tabular}
\end{center}
\end{table}

Ultimately, although problem specific representations have received wide acceptance in the EC community, they still have both their merits and demerits and should be used with great caution. This is because, although using them may improve the performance of an EC algorithm, the improvement is usually limited to only that specific problem. It also risks losing the problem independence nature of EC algorithms which is what make them robust and widely applicable. 

\subsection{EC Population: Creation and Sizing}
Evolutionary computation algorithms enjoy global search capabilities mainly due to their population based nature. The initial population in a typical genetic algorithm is mainly created randomly and of fixed size. For some problems where domain knowledge is cheaply available, simple heuristic constructions allow creation of suitable initial population or via \textit{seeding} process in which some supposedly good solution are injected into an initially random population. 

If we consider the initial population as representing a set of points in the search space of all possible populations, then, evolving over one generation effectively shifts the initial population to a different set of points in the search space. Thus, this action of evolution can be seen as a dynamic process that build-up the quality of the initial population that was randomly created. 

At low population sizes, a GA makes many decision errors and the quality of convergence suffers, but larger population sizes allow GA to easily discriminate between good and bad building blocks. And as suggested by \cite{Goldberg1989paper99}, it is the parallel processing and recombination of these building blocks that lead to deriving quick solution of even large and deceptive problems. Empirical investigations by De Jong \cite{DeJong1975} have shown that for a standard GA having binary representation, population sizes of $50-100$  are sufficient for wide range of optimization problems. 

In spite of the several theoretical viewpoints to the choice of population size, the underlying trade off between efficiency and effectiveness remains. For a given string length, larger population sizes facilitate exploration of the problem’s search space but can impair the efficiency of the search process. On the other hand, too small population size would not permit adequate exploration of the promising areas of the search space and may risk convergence to a suboptimal solution. Hence, determination of appropriate \emph{optimal} population size still remains an open area of further research \cite{RoweTheoreticalAspectsGABook2001}. 

In an attempt to establish the relationship between population size and string length, using the idea of schemata Goldberg \cite{Goldberg1985GABookref93} had earlier suggested an exponential growth in population size with respect to string length. This was later denounced after a number of empirical investigations by Schaffer \cite{SchafferCaruana1989GABookref254} and Grefensette \cite{Grefenstette1986paper94} which show that a linear relation is sufficient. Since string length significantly increases with even a slight increase in problem size and/or parameter precision, a point of further argument remains what could be regarded as the \textit{minimum} population size for a realistic evolutionary search. 

An interesting finding by \cite{Reeves1993GABookref213} reveals that at the very least, there should be one instance of every allele at each locus in the whole population of strings. This sets a minimum requirement for every point in the search space to be reachable from the initial population by a recombinative genetic algorithm (i.e. a GA having only crossover operator). 

For a typical binary representation, the probability that at least one allele is present at each locus was found by \cite{Reeves1993GABookref213} to be

\begin{equation}
\label{PopSizingEqn1}
P=\left(1-\left(\frac{1}{2}\right)^{N-1}\right)^{l}\approx\exp\left(\frac{-l}{2^{N-1}}\right),
\end{equation}
where $N$  is population size, $l$  is the strength length. 

Thus, for $99.9\%$ confidence interval, i.e. $P\geq99.9\%$ , the minimum population size $N$  can be evaluated as:

\begin{equation}
\label{PopSizingEqn2}
N\approx\left\lceil 1+\frac{\log\left(\frac{-l}{\ln P}\right)}{\log2}\right\rceil \approx\left\lceil 1+\frac{\log\left(999.5\times l\right)}{\log2}\right\rceil.  
\end{equation}

\begin{figure}[hbtp]
	\centering
	\includegraphics[scale=0.75]{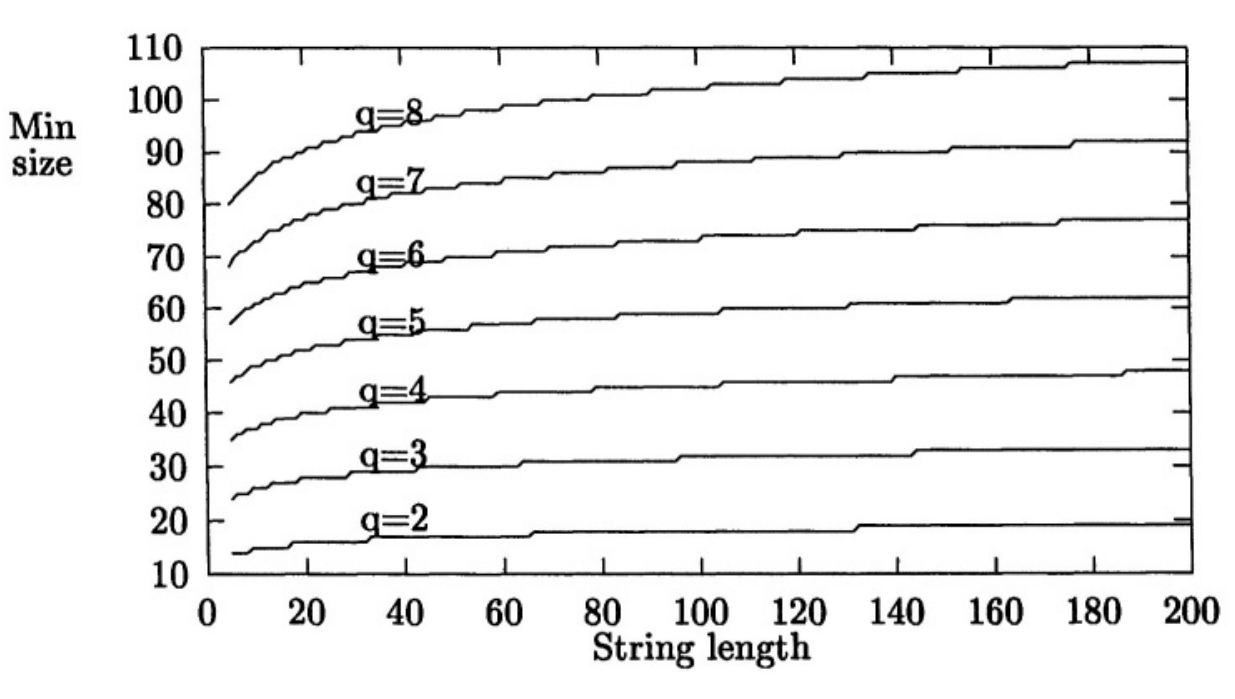} 
	\caption[Minimal Population Sizes for $q-$ary Representations]{Minimal Population sizes $N$ for $99.9\%$ confidence interval of having all allele in the initial population \cite{ReevesRow2004GABookPrincplnPers}, where $q$ is the arity of the representation, e.g. $q=2$ for binary.}
	\label{PopSizingChart}
\end{figure}

Worth noting is that the expression for $N$  here does not set the optimum value for the population size. It however prescribed a threshold value below which the population may not guarantee adequate exploration of the problem space by a genetic algorithm. Thus, in the experimentations presented in the later chapters, a much larger population size is used to avoid undersampling the search space. Nevertheless, extended application of formulation \eqref{PopSizingEqn2} for higher cardinality representations can be found in \cite{Reeves1993GABookref213} and an interesting plot for the threshold values of the minimum population sizes $N$  for higher $q-$ary representations is shown in figure \ref{PopSizingChart}. 

As can be seen from the family of curves in figure \ref{PopSizingChart}, the minimum population size required when a binary representation (i.e. when $q=2$) is used with up to a string length of $200$ bits is not more than $20$ individuals. However, this threshold grows as the cardinality of the problem's representation increases. We must remark here that the values shown by the curves are not the optimal values for the population size, neither are they sufficient for a realistic global search, but they are necessary and can be used to justify why it is possible for GA to converge to the optimal solution with an extraordinarily small population size. 

The study of evolutionary properties of a typical population under the influence of genetic operators resulted into an interesting development of a dynamical system model for the space of all possible population. A population sizing equation was proposed by \cite{Goldberg1989paper99} that facilitates accurate statistical decision making among competing building blocks\footnote{The phrase Building Block (BB) was used by Holland \cite{Holland1975} to represent the short, low order (or low defining-length) schemata. Building Block Hypothesis (BBH) suggests that GA performs adaptation efficiently by combining and processing these short, low order schemata (BB) which have above average fitness \cite{Goldberg1989Book}.} in a population-based search methods like GA. Parallel to that Rowe \cite{RoweTheoreticalAspectsGABook2001} proposes a mathematical model for a population and use it to analyse the effect of selection, mutation and crossover operators. The model permits investigation of the probability distribution of the next population, predicting the expected next population and analysing the long-term behaviour of the population. 

Assuming binary representation with string length $l$  having a search space $\mathcal{S}=2^{l}$, a population $P$ of size $N$ can be represented as a vector comprising of the proportions of each element $\mathcal{S}_{i}$  in the search space $\mathcal{S}$ such that:

\begin{equation}
\label{PopVectorEqn}
P=\left(p_{0},\, p_{1},\,...,\, p_{\mathcal{S}-1}\right);\; p_{i}\in\mathbb{R}:p_{i}=\frac{S_{i}}{N}   
\end{equation}

Supposing a population of size $N=10$  contains one copy of $00$, three copies of $01$, two copies of $10$ and four copies of $11$, then, the corresponding population vector $P$ can be represented as

\[P=\left(0.1,\,0.3,\,0.2,\,0.4\right)\]

Rowe argued that the population vector $P$  which is an element of vector space $\mathbb{R}^{\mathcal{S}}$ satisfies the following three important properties that qualifies it as a \emph{simplex} and can be denoted by $\Lambda$. First, a number of population vectors are added together will yield another vector which is also in the search space $\mathbb{R}^{\mathcal{S}}$. Second, since each element $p_{i}\in P$  is a proportion, then, $0\leq p_{i}\leq1$. And finally, the sum of all elements in a given population vector is $1$, i.e.

\[\sum_{i=0}^{n-1}p_{i}=1.\]

As GA progresses from one generation to another, in spite of the randomness induced by stochastic evolutionary operators, the dynamical model makes it possible to predict the \textit{expected} next population since the probability distribution of the population vector $P$  is always a member of the set $\Lambda$. Consequently, if the population size is large enough (i.e. as $N\rightarrow\infty$), the chances of having the next population been the expected one grows. In the limit, when the population size is infinite, the next population can accurately be predicted thereby turning the search process into a deterministic one. Because of this important property, the dynamical system model is often referred to as \emph{infinite population model}. And in essence, this concept is now used to predict the behaviour of finite populations typically used in genetic algorithms. Details can be found in \cite{RoweTheoreticalAspectsGABook2001}.

\section[The Selection Process in Evolutionary Algorithms]{The Selection Process in Evolutionary Algorithms}

As argued by many \cite{Hancock1994paper97, Blum2008paper88}, selection is the driving force in evolutionary algorithms. Although evolution as a whole is seen as a set of processes that operate on chromosomes represented in the genotype space, selection process works on the original form of the encoded problem, i.e. in the phenotype space. According to Darwinian notion of natural selection \cite{Atmar1994paper17}, so long as the genes (encoding space) and the behaviour (phenotype space) are separate entities, as they are almost always are, selection optimizes the functional behaviour not the underlying code. Thus, selection methods in simulated evolution also imitate the process of natural selection by choosing chromosomes that encode successful structures to undergo reproduction more often than others that do not. Successful chromosomes hereby refer to those individuals that after been evaluated happen to have higher fitness values relative to their counterparts. We must remark here that some researchers \cite{Fogel2008paper90} in the area of evolutionary computation do take issue with this idea and raise questions like, \textit{``does natural selection always favours those behavioural strategies that seek to minimize expected loss?"} 

In practice, there are varieties of selection methods in evolutionary computing, but not all of them directly use the fitness as the only criteria for carrying out the selection operation. The choice of appropriate selection method can be difficult as it involves deciding on crucial parameters like selection pressure, selection intensity, growth rate, takeover time etc. which consequently dictate the mode and rate of convergence of the EC. Thus, selection plays an important role in the parameterization of EC and as \cite{Goldberg1991paper93} asserts, selection is critical to the overall success of an EC algorithm. 

\subsection{Fitness Proportionate Selection methods}
\label{FPSselectionSection}
Roulette wheel selection (RWS) is the simplest and most commonly used fitness proportionate (FPS) means of selection in EC. In this technique, an individual is selected in proportion to its fitness on the evaluation function, relative to the average fitness of the entire population. In other words, RWS has a probability distribution such that the probability of choosing an individual is always directly proportional to its fitness. This probability intuitively corresponds to the area of a sector of a roulette wheel (or the angle subtended by the sector at the centre of the wheel). The larger the sector, the higher the chances of having the individual been selected. Hence, the name of the roulette wheel selection method. 

For a simple construction of the roulette wheel, consider a population of size $N$  consisting of a set of chromosomes $\left\{ x_{i}\right\} :\: i=1,\,2,\,...,\, N$.  Let the fitness evaluation function be $f$, then, the total fitness of the population $F$ is

\begin{equation}
	F=\sum_{i=1}^{N}f(x_{i}).
\end{equation}

Therefore, the proportion of the fitness of chromosome $x_{i}$  corresponds to its probability, $p(x_{i})$:

\begin{equation}
p(x_{i})=\frac{f(x_{i})}{F}:\: i=1,\,2,\,...,\, N
\end{equation}
and its cumulative probability $q(x_{i})$ is:

\begin{equation}
q(x_{k})=\sum_{j=1}^{k}p(x_{j}):\; k=1,\,2,\,...,\, N;\; q(x_{k})\in\left[0,\,1\right].
\end{equation}

Selection of $N$  chromosomes requires spinning the wheel $N$-times which corresponds to $N$-times sampling (with replacement) from a pseudo-random sequence $rand\,\in\left[0,\,1\right]$.  After every spin, chromosome $x_{k}$  is selected if the random number falls within the interval of its cumulative probability and that of its predecessor, i.e. if $q(x_{k-1})\leq rand\leq q(x_{k})$.  It is not difficult to notice that chromosomes with higher fitness values will have larger cumulative probabilities and hence higher chances of been selected. 

Although RWS scheme enjoys a great level of simplicity, it suffers from at least three critical problems:

\begin{enumerate}[(i)]
	\item \textbf{Sampling error}: Because RWS uses pseudo-random numbers to choose an individual for reproduction, the expected number of times an individual is  selected may significantly differ from the actual value as a result of high \textit{stochastic variability}. In the worst case, the tendency of not having higher (best) fit individuals being selected by the roulette wheel increases.
	\item \textbf{Scaling}: In fitness proportionate selection methods, for any given evaluation function, if an individual has a fitness value of $2$ and another has $4$, the method will allocate twice chances of reproduction to the latter individual than the former. A mere shift of their fitness values by 1 (i.e. to fitness values of 3 and 5) will change the selection preference of the second over the first from twice to only $1.67$-times. Another possible drawback of this lack of scaling in RWS is that at the early stages of evolution, majority of the individuals have fairly lower levels of fitness, if any individual happens to have relatively higher fitness value the selection method will favour it too much to the extent that the search process may prematurely converge to a sub optimal solution.
	\item \textbf{Selection pressure}: For the reason described in (ii) above, as the average fitness of the entire population increases, selection pressure falls. Thus, chances of selecting the best fit individual over the average (or worst) individuals dramatically drop. This may lead to stagnation of the entire search process.
\end{enumerate}

In summary, it is easier for FPS schemes to distinguish between values like ($5$ and $10$) than ($105$ and $110$). Thus, while too high genetic variation among individuals within a population results in excessive selection pressure and risk premature convergence, too little of this genetic variation collapses selection pressure and halts the evolution in FPS selection methods.

A proposal by Baker \cite{Baker1987GABookref15} to address the sampling error caused by the excessive stochastic variability inherent in the RWS method is to use a multi-armed roulette wheel having equally spaced $n$-arms. Spinning the wheel ones will corresponds to sampling $n$  pseudo-random numbers thereby allowing $n$  selections at once. Thus, Hancock et al. \cite{Hancock1994paper97} argue that this is a systematic random sampling and is superior from the statistical point of view over the traditional RWS method. Another variety of RWS method can be found in \cite{Hongze2009paper8} where RWS is used in job scheduling in a setup that favours lower fit individuals against those with higher fitness values. 

The commonly used measures \cite{Hancock1994paper97, LanceChambers1995Book, ReevesRow2004GABookPrincplnPers} to tackle the lack of scaling in RWS include windowing and scaling (linear, sigma). Indeed these techniques can to some degree serve as a remedy to the scaling problem in RWS as empirically proven by \cite{Hancock1994paper97}. However, in addition to the overhead involved, they principally fail to eliminate the sampling error inherent in FPS selection schemes. This is no doubt one of the reasons why, we can roughly argue from our literature survey that whenever an FPS selection method like RWS is utilized, \textit{elitist} strategy will be explicitly adopted to avoid the risk of losing promising individuals due to sampling errors.

\subsection{Ranking methods}
Ranking is another method developed to tackle the problems in FPS selection schemes. Originally proposed by Baker \cite{Baker1985paper97ref1} it entails ranking the chromosomes in order of their fitness values. Although ranking can lead to loss of some information regarding the actual fitness of the chromosomes, it successfully eliminates the need of rescaling and yields a simple and efficient selection algorithm. Ranking technique assigns new fitness value $f$ to a ranked chromosome $i$ inversely proportional to its rank $N$ in the population. The best individual is assigned a fitness value $s:\:1\leq s\leq2$,  the worse is assigned a value $2-s$  while the intermediate individuals are assigned fitness by interpolation according to the following linear ranking equation.

\begin{equation}
\label{LinearRankEqn}
	f(i)=s-\frac{2i(s-1)}{N-1}.
\end{equation}

Apart from linear ranking, exponential ranking is another variant of ranked selection method designed to give more diverse population. The method favours worst individuals at the expense of the above average ones. Both linear and exponential ranking methods facilitate maintenance and tunability of the selection pressure throughout the evolution process. One major drawback \cite{Hancock1994paper97} of ranking methods is the loss of information about the actual fitness values of the chromosomes. This can negatively alter the correlation between fitness and chances of reproductive success.

\subsection{Tournament selection methods}
\label{TournamentSelectionSection}
Tournament selection method is a non-direct fitness proportionate selection scheme. It was inspired by the natural mating contest in which a group of individuals compete for reproduction. Out of a population of size $N$, $k$  individuals are selected at random for the contest. An individual with higher fitness value wins the contest and is forwarded to the reproduction phase. The process is repeated until the required number of individuals is selected. 

It can be observed that this technique may have several varieties since for instance; selecting the contestants can be carried out \textit{with} or \textit{without} replacement. It is worth noting that running a series of tournaments with replacement risks having the higher (best) fit individuals not been selected for any of the contests. This can primarily give rise to sampling errors in tournament selection that is done with replacement. Further, the tournament size $k$  can be varied with the consequence that larger $k$ leads to increased selection pressure. Hence, this method gives improved control over the selection pressure and is immune to any scaling problem that usually lead to collapse of selection pressure as the average fitness of the population grows. 

A complete tournament selection cycle that yields $N$ chosen individuals will require each chromosome taking part in a contest $k$  times. Naturally, the best individual will win all its $k$  contests, an average individual is expected to win half of its $k$  contest and at the other extreme, the worse individual will lose all its contests. This kind of setup is what is referred to as the \textit{strict} tournament selection and is the most commonly used method. There is however other variants like the so-called \textit{soft} tournaments where in any contest, the fitter individual only wins the tournament with a probability $p<1$. This method is also called stochastic tournament selection since for example, instead of the best individual to be selected $k$ times, it is now reduced to $kp$. A value of $p=0.5$ will turn the process into a random selection with no preference given to the best individual. It is clear that even with $p>0.5$, stochastic tournament selection will occasionally suffer from possible sampling errors. 

A tournament selection of size $k=2$ is called binary tournament and it was analytically shown that \cite{Rudolph2000paper98, Hancock1994paper97}, the dynamics of a binary strict tournament selection resembles that of a specific form of linear ranking where the best ranked individual is assigned a new fitness value of $2$ (i.e. $s=2$ in equation \eqref{LinearRankEqn}). On the other hand, a stochastic tournament selection has been shown to resemble a linear ranking with $s<2$. And finally, it is not a coincidence that a stochastic tournament with $p=0.5$ is equivalent to a linear ranking having $s=2$. 

Other selection schemes include truncation selection \cite{Thierens1994paper13}, steady state GAs or Genitor \cite{Goldberg1991paper93, Hancock1994paper97}, and the $(\lambda,\,\mu)$  and $(\lambda+\mu)$  methods originally inherited from evolutionary programming and evolutionary strategies \cite{ReevesRow2004GABookPrincplnPers, Goldberg1991paper93}. Discussions on elitism and its varieties will follow in section \ref{ElitismSec} of this chapter. 

As previously highlighted, when investigating selection schemes it is imperative to understand the crucial parameters that govern the choice of appropriate selection scheme. Some of these parameters are said to have originated from field of quantitative genetics \cite{Thierens1998paper96} and are presented below:

\begin{enumerate}[i.]
	\item \textbf{Response to selection} $R(t)$: As the name implies, this measure quantifies the difference in the average fitness of population between two successive generations. In other words, the difference between the average fitness of population at generation $t$  and its average fitness at generation $t-1$. \\
\begin{equation}
R(t)=\overline{f}(t)-\overline{f}(t-1)
\end{equation}

where $\overline{f}(t)$ is the average fitness of the population at generation $t$.
	\item \textbf{Selection Differential} $S(t)$: Conversely, this measures the difference between the average fitness of the parent set $\overline{f}_{p}(t)$ (i.e. individuals selected for reproduction) at generation $t$ and the average fitness of the entire population $\overline{f}(t)$ also at generation $t$.\\
\begin{equation}
S(t)=\overline{f}_{p}(t)-\overline{f}(t)
\end{equation}

where $\overline{f}(t)$ and $\overline{f}_{p}(t)$ are the average fitness of the entire population and that of the parent population respectively.
	\item \textbf{Selection Intensity} $I(t)$: This is a dimensionless quantity derived by taking the ratio of the selection differential $S(t)$ with the population standard deviation also at generation $t$ such that:
\begin{equation}
I(t)=\frac{S(t)}{\sigma(t)}=\frac{\overline{f}_{p}(t)-\overline{f}(t-1)}{\sigma(t)}
\end{equation}
where $\sigma(t)$ is the population's standard deviation at generation $t$.\\
Although this definition restricts the use of selection intensity to analyse the effect of selection to a specific generation in the evolution, it can be deployed to investigate the effect of selection for the entire evolution run if the distribution of the population vector is normal and remains constant (which is often true \cite{ReevesRow2004GABookPrincplnPers}). In this case, $I(t)=I$. Goldberg et al. \cite{Goldberg1991paper93} argued that when the population vector has a standard normal fitness distribution, the selection intensity is basically the expected average fitness of the population under the influence of the selection algorithm. 
	\item \textbf{Selection Pressure} $P$: Notice that although selection intensity $I(t)$ can be useful for comparison purposes, it is application is limited to generational GAs where there is a clear set of parent population, i.e. $I(t)$ cannot be used to analyse elitist or steady state GAs. Thus, selection pressure is another measure that estimates the expected number of offspring a best fit individual will have after selection. It is highly related to the convergence rate and can be defined in various ways as noted by \cite{Hancock1994paper97}. A straight forward definition for $P$ as given in \cite{ReevesRow2004GABookPrincplnPers} is in terms of the ratio of probabilities:\\
\begin{equation}
\label{SelectionPressEqn}
P=\frac{Pr(\textrm{selecting best string})}{Pr(\textrm{average string})}
\end{equation}
	\item \textbf{Takeover time}: This is a measure that estimates how long it will take the best individual to take over the entire population. For any selection method applied on a population of size $N$ consisting of a single copy of the best individual, takeover time can be determined by evaluating the expected number of generations before getting a population vector that entirely consists of the copies of the best individual from the initial population. Majority of the studies on takeover time \cite{Rudolph2000paper98, Goldberg1991paper93} are conducted on selection-only genetic algorithms or an elitist GA with mutation. Nevertheless, it remains an important measure that furnishes valuable insight into the complexity, growth rate and many other characteristics of selection methods. 
\end{enumerate}

An early work by Goldberg et al. \cite{Goldberg1991paper93} reviewed various selection methods and compare them based on their time complexities, takeover times, growth ratios and selection pressure. They analytically prove that the complexity of the fitness proportional selection methods is at its best $\mathcal{O}(n\log n)$  and the worse case can be $\mathcal{O}(n^{2})$. Ranking methods are also no different. However, tournament selection methods have a polynomial complexity of order $\mathcal{O}(n)$. 

EC algorithms are increasingly popular because many of the evolution processes involved can easily be executed in parallel. However, selection operators are originally designed to work on the entire population and thus, their traditional implementation is sequential and can severely inhibit performance. B\"ack et al. \cite{Back1997paper16} suggest applying a parallel algorithm to conduct local selection within subpopulations or neighbourhoods such as in \textit{migration} or \textit{diffusion} models. Goldberg et al. \cite{Goldberg1991paper93} share the same view with regard to the need for global information by various selection methods, but they argue that tournament selection is an exception. Tournament selection scheme can easily be parallelized since it naturally works by setting up contests on subpopulations rather than the entire population as a whole. 

With regard to selection pressure, while on one hand \cite{Harik1999paper83, Goldberg1989paper99} noted that fitness proportionate methods are slower and have the least selection pressure as compared to ranking and tournament selection methods. Although this may occasionally be beneficial on the quality of convergence (since less pressure is applied to force bad decisions), it simply heightens the number of unnecessary function evaluations. It may also exacerbate the effect of \emph{genetic drift} \cite{LouisRawlins1993paper115} thereby forestalling convergence to promising solutions. On the other hand, steady state genetic algorithms (such as genitor) have the highest selection pressure. \cite{Goldberg1991paper93} stressed that the selection pressure of genitor is twofold (one for always selecting the best and the other for always having to replace the worse individual). The consequence of this is intense susceptibility to premature convergence to suboptimal solutions as a result of rapid loss of diversity. We therefore argue that with tournament selection method, moderate levels of selection pressure can be maintained throughout the evolution process and it can easily be tuned by simply varying the tournament size. Detailed discussion on selection pressure from the perspective of natural evolution can be found in \cite{Atmar1994paper17}. 

Moreover, successful choice of a selection method requires analysing their convergence characteristics. Based on the normal distribution theory, Thierens et al. \cite{Thierens1994paper13} developed elegant convergence models for the fitness proportional, truncation, tournament and elitist recombination selection schemes. Their analysis reveals that selection essentially leads to build-up of covariance among allele, and, the growth rate in fitness proportional selection is directly proportional to the fitness variance but inversely proportional to the mean fitness. This fully agrees with the Darwin’s original idea of \textit{sexual selection} \cite{Atmar1994paper17} which he described as a milder form of natural selection and is proportional to population variance. Thus explaining why selection progress in FPS methods drops as the average fitness of the population grows. 

Rudolph et al. \cite{Rudolph2000paper98} modelled the takeover times of various tournament methods as Markov chains and cautioned that stochastic tournament methods just like FPS are prone to sampling errors and should be used with care. They also admit that in spite of the significance of takeover time, it is not sufficient to critically decide on the choice of selection scheme. 

The effect of noise on various selection schemes was investigated by Hancock \cite{Hancock1994paper97}. Gaussian noise was added to the evaluation functions and the growth rates in the presence of mutation for various selection methods were observed. Hancock argued that contrary to the traditional judgement that genetic algorithms are immune to noisy evaluation functions, some empirical results have shown that noise effectively increases sampling error due to selection and consequently lowers selection pressure. 

Elsewhere, \cite{Bassett2005paper4} investigates the placement of selection operation in a typical run of evolution algorithms from two different perspectives. In a classic GA setup, the evolution process begins with selection, crossover and then mutation. This setup is called \textit{parent selection} because it applies the selection operation on the parents. However, a common practice in ES but also used in GA is to put the selection operation at the tail of the evolution, such that the process begins with crossover, mutation and then selection. Because selection is applied on the offspring population to generate the next population, this setup is called \textit{survival selection}. While these two techniques are intuitively similar and may yield same end results, they differ in the way they generate their intermediate populations because, while survival selection ends with a selection operation, parent selection only begins with it.

In the area of hybrid evolutionary computation further investigation of selection techniques yields some novel selection methods. For instance, \cite{Hongfeng2008paper40} hybridizes a genetic algorithm by embedding a Nelder-Meid simplex algorithm into different niches of the genetic algorithm. The simplex algorithm serves as a crossover operator and was used together with simplex multi direction search to improve the selection process in GA. In a similar fashion, \cite{Shi2009paper45} utilizes a simulated annealing (SA) algorithm to undertake selection on the population of candidate solutions such that the SA algorithm is integrated as an operator in the parent genetic algorithm. Some other hybrid algorithms that compare various standard selection algorithms include; \cite{Merkle1996paper58} which compares FPS with tournament selection and \cite{Vallejo2008paper62} which compares ranking and RWS selection methods with the view to categorize their proposed hybrid algorithm. 

Detailed and interesting comparison of various tournament selection methods can be found in \cite{Rudolph2000paper98, Lozano2008paper11}. Further discussions and comparisons of various selection schemes based on the aforementioned parameters can be found in \cite{Goldberg1991paper93, Thierens1998paper96, Thierens1994paper13}. 

To sum up, tournament selection method strikes a \emph{perfect} balance between the FPS methods that tend to be too inclined to the biological evolution (i.e. adaptation) and the steady state GAs that tend to be too inclined to the simulated evolution (i.e. optimization). It maintains moderate level of selection pressure and steady growth rate which are key ingredients that guide the evolution process to converge to optimal solution. These and many other remarkable features of tournament selection are the reasons that we have chosen to adopt this method for the proposed hybrid evolutionary computation algorithm. 

\section{Recombination and Mutation}

Having chosen the parent population via what is basically a biased process (selection) in the first phase, some variation operations are necessary for the evolution to make progress by exploring other areas of the search space. This second phase of evolution is called the reproduction phase and it consists of two main operations, namely recombination (i.e. crossover) and mutation. All the operations in the reproduction phase are taking place in the encoding (genotype) space and that is where the \emph{hidden} power of evolutionary algorithms lies. It is interesting to note that as the only source of introducing new individuals in to the population of candidate solutions, the reproduction phase is completely unaware of the original problem formulation (phenotype) and that is what made evolutionary algorithms problem-independent and robust means of global optimization.

Researchers in evolutionary computation \cite{Boyabaltli2007paper95} categorize parameters in genetic algorithms into structural and numerical parameters (detailed under the parameterization section). The reproduction operators are involved in both category of parameterization and are therefore critical to the successful design of any evolutionary algorithm. Many practitioners of genetic algorithms \cite{Davis1990paper57} are of the view that crossover is the primary reproduction operator that clearly distinguish genetic algorithms from other types of optimization algorithms because mutation can be regarded as a sort of local search or hill climbing operation. 

Besides crossover and mutation, Holland's \cite{Holland1975} original proposal includes another variation operator called inversion, this operator differs slightly from mutation and has not receive wide acceptance in the evolutionary computation community. Thus, inversion will not be covered in the following background as it is not utilized in our proposed hybrid algorithm. 

\subsection{Implementation of Crossover in EAs}
In its simplest form, crossover operator yields two offspring by exchange of genetic materials between two parent strings subject to probabilistic decisions. In a typical genetic algorithm in which individuals are represented as binary strings, if any two parents $P_{1}$ and $P_{2}$ are to participate in crossover operation, then, they will exchange their bits to the right of a randomly chosen locus called a crossover point to yield two new strings (offspring) $O_{1}$ and $O_{2}$. Suppose every individual is encoded with $7$ bits and the random crossover point happens to be at locus $3$, then, parents $P_{1}$ and $P_{2}$ will produce $O_{1}$ and $O_{2}$ after crossover, for example:

\[P_{1}=1010110\quad O_{1}=1011101\]
\[P_{1}=0101101\quad O_{2}=0100110\]

This kind of crossover is called single point (or $1X$ ) and can easily be extended to a number of variants by creating a number of random crossover points after which the parent strings exchange their bits. This extended version of one point crossover is therefore called multi-point ($m-$point) crossover, $m>1$. 

As agreed by many \cite{Grefenstette1986paper94, Boyabaltli2007paper95}, crossover plays two key roles in evolution. First it provide a chance for further examination of the already available hyperplane, like offspring $O_{1}$ simply continues with the exploitation of the hyperplane \mbox{$101\star\star\star\star$}\footnote{The $\star$ fields stand for any possible allele value, e.g. 1 or 0.}. Second, it allow exploration of new area of the search space like the hyperplane \mbox{$\star1001\star\star$} in offspring $O_{2}$. Hence, every evaluation of a string of length $l$ guides the search process by adding knowledge of $2^{l}$ hyperplanes. These two complementary roles of crossover are critical for a successful evolutionary search. Hence, the success of GA in many applications has been attributed \cite{Hong2002paper2} to the adopted method of reproduction. 

An argument reported in \cite{EshelmanCaruana1989} reveals that two bias sources (positional and distributional) are central to the implementation of crossover operators in genetic algorithms. The single point crossover operator is said to suffer positional bias because it tends to favour substrings of contiguous bits. It however lacks distributional bias since the crossover point is randomly chosen from a uniform distribution. Yet, \cite{ReevesRow2004GABookPrincplnPers} argued that none of these biases can be said to be working for the betterment or otherwise of the search process. 

Out of the several proposals to tackle the bias in crossover, a popular alternative is the so-called \textit{uniform} crossover. Uniform crossover seeks to eliminate the bias by making the operation completely random. The process requires representing the crossover operator as a binary mask obtained from a typical Bernoulli distribution. Analyses by De Jong \cite{DeJong1975} reveal that the disruption caused by uniform crossover can be tuned by varying the Bernoulli parameter $p$ associated with the distribution. A value of $p=0.5$ results in a completely random generation of the crossover mask and can eliminate any bias in genetic algorithm due to the crossover. 

\subsubsection*{Other Implementation issues}
Not every optimization problem can be adequately represented in binary, even if it is possible to do so, utilizing any of the aforementioned crossover operators on some optimization problems would simply lead to creation of infeasible solutions. This necessitates the need for specially designed crossover operators for scheduling problems, permutation problems like the TSP and their like. Specially designed crossover operators such as partially matched crossover (PMX) \cite{Wang2008paper54} and order crossover are designed for TSP and other permutation problems. A generalized $n-$point crossover (GNX) \cite{Fogarty1994paper52} is designed to be representation-independent and suitable for all problem domains. Thus, it sets up a general framework for recombination and seeks to alleviate the need for several \emph{ad hoc} crossover implementations.

Nevertheless, many problem dependent crossover operators are still proposed \cite{Jeurissen2008paper66, Hongfeng2008paper40, Wan1999paper49} and applied on various problem domains. Man et al. \cite{Man1999paper59} argued that conventional crossover operators do not perform well on complex optimization problems because they lack problem-specific knowledge in their encoding. Thus, elsewhere, an improved genetic algorithm (IGA) featuring a new crossover operator was proposed by \cite{Wen2010paper36}. The new operator was designed to balance the exploitation-exploration trade off by producing two offspring biased for exploitation and another two biased for exploitation. 

As mentioned earlier, crossover operator is commonly applied across a population of individuals based on some probabilistic decisions. A survey of various theoretical analyses \cite{Back1997paper16} have shown that crossover probabilities $P_{c}=\left[0.6,\,1.0\right]$  are considered optimal for most global optimization problems. Elsewhere, Grefenstette \cite{Grefenstette1986paper94} uses a multi-level genetic algorithm where the parameters of an outer GA are optimized by another GA running internally. Experimental results from \cite{Grefenstette1986paper94} have shown that while smaller population sizes requires high crossover rate, larger sizes do not. In other words, crossover rate can be safely reduced as population size increases. The results suggest that $P_{c}=0.30\textrm{ or }0.88$ are respectively suitable for population sizes of $N=80\textrm{ or }30$. 

Although these proposals have proven to be quite effective on wide range of global optimization problems, several recombination operators that automatically adapt their rates \cite{Xingbo2008paper32} have been proposed. In this regard, a statistics-based adaptive non-uniform crossover was proposed by \cite{Herrera2000paper10}. They argue that there is an implicit convergence trend that leads to the build-up of building blocks by alleles of each string which the conventional crossover operators fail to exploit. Hence, they suggest using some statistical information of alleles at each locus to adaptively calculate its crossover probability. Some experimental results on deceptive test functions have shown that the adaptive crossover performs better than both the conventional multi-point and uniform crossover operators. Punctuated crossover \cite{Back1997paper16} is another typical example of $m-$point crossover where both the positions and number of crossover points can be adapted. 

\subsection{Implementation of Mutation in EAs}
Commonly, evolutionary algorithms rely on two variation operators for the reproduction of new individuals. Besides crossover, mutation is the second variation operator that also works in the genotype space and is capable of producing new individual(s) from a single parent string. Unlike the crossover operator that require large population sizes to effectively exploit the necessary information from parent strings, mutation can effectively guide the exploration of the search space when applied to a small population for large number of generations. It involves simple bit flip operations for binary represented strings or a random addition of Gaussian noise to individuals represented in real values. 

Consider the following binary represented parent string $P$ of length $l=8$ in a hyperplane $101\star\star\star\star\star$\footnote{The $\star$ fields stand for any possible allele value, e.g. 1 or 0.}. Mutating bits $2$ and $6$ yields an offspring $O$  located at an entirely different hyperplane $111\star\star\star\star\star$. Thus, as accorded by many \cite{Bassett2003paper3, Herrera2000paper10} mutation principally enables the search process to escape from sub-optimal regions by jumping across hyperplanes thereby preserving the required diversity in a population.

\[P=10101101\quad \Rightarrow\footnote{The parent $P$ yields the offspring $O$.}\quad O=11101001 \]

Notice that applying mutation to a $q-$ary (where $q>2$) represented population will require a more careful treatment because each mutated bit can take up to $q-1$ possible allelic values. This may somewhat introduce some complications in the mutation operation. Thus, decisions must be made as to whether to allow stochastic replacement of the mutated bit by any possible allele or bias the choice to alleles with values nearer to the current bit. 

In addition to the aforementioned structural parameterization issues, the numerical aspect of the mutation operator involves its application subject to a probabilistic decision. For a simple genetic algorithm having individuals represented as binary strings of length $l$, the rate of application of mutation is usually defined as the probability of flipping a bit and is denoted by the parameter $P_{m}$. Experimental results by \cite{Grefenstette1986paper94} have shown that a range of mutation probabilities of $P_{m}=\left[0.01,\,0.05\right]$ is sufficient for wide range of global optimization problems. A universal setting \cite{DeJong1975} for adapting the probability of mutation is to set $P_{m}=\nicefrac{1}{l}$. This has received wider acceptance \cite{Back1997paper16} in the genetic algorithms community. However these settings are not without some criticism from some researchers. Salomon \cite{Salomon1996paper109} argued that small mutation rates are mostly suitable for problems of unimodal or (pseudo-unimodal)\footnote{pseudo-unimodal problems are multimodal problems having a general convex shape, example is the well known Griewank \cite{Whitley1996paper108} benchmark function.} type, but multimodal problems will require larger rates. Also, B\"ack \cite{Back1993paper111} criticised the universal rate of $P_{m}=\nicefrac{1}{l}$ as being too independent and unaware of the fitness landscape. 

There have been several proposals \cite{Herrera2000paper10} for some kind of adaptive mutation with the aim of reducing the burden of parameterization in GAs. Adapting the probability of mutation $P_{m}$ is mainly aimed at eliminating the tuning problem associated with this parameter in order to make genetic algorithms more parameter-independent. Queiroz et al. \cite{Queiroz2009paper43} proposed an excellent adaptive mutation strategy that assesses the current population diversity prior to setting the mutation rate. Elsewhere, a new mutation operator for permutation problems was proposed by \cite{Hongze2009paper8} that allows every individual and gene to mutate at different mutation rates. In a similar trend, Bassett et al. \cite{Bassett2004paper5} applied an adaptive Gaussian mutation operator to a real valued optimization problem. They argue that the adaptive mutation is more disruptive, but it can contribute more to the search process. B\"ack \cite{Back1993paper111} suggests a new measure called \emph{effective fitness} which helps to categorize the fitness landscape prior to adapting the mutation rate.

Furthermore, in the field of hybrid optimization algorithms, proposals by \cite{Zhufang2009paper24, Chang2001paper42} suggest annealing the rate of mutation as the population nears convergence. In a hybrid setup, a simulated annealing algorithm is integrated into a GA to serve as the mutation operator. Elsewhere, Hong et al. \cite{Hong2000paper1} use various mutation operators simultaneously and argue that dynamically choosing an appropriate mutation operator for a given problem at a given stage of optimization can significantly enhance the robustness of the search process. Thus, although many other strategies are possible, it is evident that many authors use varying mutation rates as a diversity maintenance policy.  

As can be noticed from the foregoing, in spite of the wide acceptance of the standard reproduction operators, there have been quite a large number of adaptive, dynamic and problem-dependent operators that have been tried in the literature. A somewhat strange proposal by \cite{Man1999paper59} in a set up of a hybrid genetic algorithm features a number of specially designed pairs of reproduction operators. The design entails restarting the evolution process with a new pair of operators whenever progress stalls by a given pair of operators. And the process continues until all operator pairs are exhausted. 

To sum up, without loss of generality, we argue that crossover and mutation are the main reproduction means in genetic algorithm. Thus, if appropriately set, they can facilitate the exploration and reachability of the entire search space even in problems having rough and complicated landscapes. Possible consequences of trying to adapt these settings may lead to increased complexity and in the worse case, it risks losing the problem independence characteristics of evolutionary algorithms as a whole. Yet, reproduction operators play a major role to the success of evolutionary search. And as emphasized by \cite{Bassett2004paper5}, it is not the average individuals that drive the evolution forward, but the occasionally exceptional individuals created by crossover and mutation that keep the population improving over generations.

\section{Replacement Strategies}
The field of evolutionary computation as highlighted earlier has three major sub-fields: genetic algorithms, evolutionary strategies and evolutionary programming. The standard replacement strategy in genetic algorithms recommended by Holland \cite{Holland1975} is the \emph{generational} replacement scheme. This scheme mimics the natural evolution in such a way that subsequent to reproduction phase, an offspring population completely replaces the parent population. In the ES and EP communities, this strategy is popularly referred to as $(\lambda,\,\mu)$ strategy. In addition to this, a $(\lambda+\mu)$ strategy combines both parent and offspring population and then select the best fit individuals in the combined set such that the size $N$ of the resulting new population is equal to that of the original population. The gap between these fields of evolutionary computing has been blurred and every replacement strategy has now been tried everywhere else. 

It is quite easy to realize the risk involved in the generational replacement method. It genuinely fits the natural evolution but it is unsuitable for actual optimization purpose where the goal is to explore and keep the best solution found. Realization of the danger of losing the optimum solution \cite{Greenhalgh2000paper12, DeJong1975} over generations following disruptions caused by crossover and mutation has led to the proposals of elitism and its variants details of which will be considered in the next section. 

\subsection{Elitism in Evolutionary Computations}
\label{ElitismSec}
The notion of elitism was originally coined by \cite{DeJong1975, Grefenstette1986paper94} after critical analysis of the behavioural trend in the evolution dynamics based on several empirical experiments. The concept is basically aimed at preserving the candidate solution having the best fitness value as the population evolves over successive generations. Hence, the best fit individual can only be lost if the new population contains another candidate solution of higher fitness value. Thus, while the idea seems contrary to the Holland's original generational replacement, it conforms better to the overall goal of optimization which is to determine the \emph{best} solution rather than merely improving the average fitness of a solution set via an evolutionary technique.

Thereafter, many other variants of the elitist strategy quickly surface. A generalized categorization of replacement schemes by \cite{Back1997paper16} is:

\begin{enumerate}[i.]
	\item \textbf{Generational}: Also called non-elitist simple GA (sGA), it corresponds to the ES $(\lambda,\,\mu)$ strategy.
	\item \textbf{Elitist}: This follows the standard elitism and mildly corresponds to the ES $(\lambda+\mu)$  strategy.
	\item \textbf{Overlapping}: This generalizes other variants of the elitist strategy and can simply be represented as $(\lambda,\, k,\,\mu)$ strategy where $k$ is an aging parameter $1\leq k\leq\infty$ signifying an individual's life span. i.e. the maximum number of generation an individual can survive.
\end{enumerate}

\subsection{Overlapping Populations and Steady State GAs}
As highlighted earlier, overlapping population refers to the replacement strategy in which the parent and offspring populations compete such that some percentage of the parent population survive across generations. In the original elitist strategy, only a single (best) individual from the parent population is carried along to the next generation. The overall influence of which tends to diminish as the population size grows. Thus, overlapping populations are meant to address this lack of scalability associated with the original elitist strategy. The degree of overlap of populations between two successive generations (also called \textit{generation gap}) can vary significantly depending on the ultimate goal of the optimization set up. For instance, \cite{Queiroz2009paper43} replace only $8$ worst individuals in the parent population with the $8$ best new individuals from the offspring population to generate new population. 

Also a product of overlapping population, the steady state GA (ssGA) is a special case that permits replacing only a single or two individual(s) from the parent population by an individual with best fitness value from the offspring population. It is also popularly known as incremental replacement scheme and is what the Whitley's \emph{genitor} \cite{Whitley1989GABookref301, Hancock1994paper97} is uniquely known for. Studies \cite{Lozano2008paper11, Whitley1989GABookref301} have shown that ssGAs have in many cases outperformed the standard genetic algorithms for reasons attributed to their ability to effectively exploit promising regions of the search space while retaining the necessary diversity. 

Worth noting is that the idea of always replacing the worst by the best individual can severely heighten selection pressure. Thus, most of ssGA implementations require some radical measures to evade premature convergence due to potential loss of diversity. One of the typical ways used to stabilize the selection pressure in ssGAs is to deploy excessively higher mutation rates. Kaur et al. \cite{Kaur2008paper68} use an overlapping population with $50\%$ overlap for optimization of a TSP problem but adopt a mutation rate of $P_{m}=0.85$ and utilize RWS selection method. Recall that fitness proportionate selection methods like RWS inherently have lower selection pressure that continue to fall over generations. Therefore, utilizing RWS with overlapping populations like ssGA is also a common practice \cite{Zhufang2009paper24, Fogarty1994paper52, Wen2004paper53, Vallejo2008paper62, Zhao2008paper69, Wu2009paper70, Huo2000paper74} that aids balance their disproportionate selection pressure. 

In agreement with the above issue, Il-Seok et al. \cite{IlSeok2004paper56} recommend setting the crossover probability as $P_{c}=1$ for the elitists, overlapping replacement schemes and all their variants like ssGA. They however argued that a much lower value like $P_{c}=0.6$ is necessary for a generational replacement scheme in order to ensure the highly fit individuals survive across generations. 

Also, in an attempt to preserve population diversity in ssGAs, a recent proposal \cite{Lozano2008paper11} advocates replacing only the worst individual that has the least contribution to diversity in the parent population. They defined the contribution to diversity as a measure of the similarity between an individual and its nearest neighbour in the population. Elsewhere, a replacement strategy called \emph{elitist recombination} was proposed by \cite{Thierens1994paper13} in which the replacement and recombination operations are interleaved. Thierens \cite{Thierens1998paper96} justified the effectiveness of the method and derived the expressions for its selection intensity and selection pressure. 

Other strategies include the use of \emph{niching} and \emph{crowding}. A niche is a subset of individuals in a given population that share some similarity. The idea is to allow a new individual to replace only those individual in its own niche, rather than potentially any individual in the population. Several other replacement schemes include restricted tournament selection (RTS), worst among most similar replacement policy (WAMS), family competition replacement schemes \cite{Lozano2008paper11} such as correlative family-based selection to mention but a few. 

To sum up, it is worth mentioning that the main purpose of elitist replacement scheme and its variants is not to guard against sampling error due to selection algorithms \cite{Hancock1994paper97}, but as a safeguard to possible disruptions caused by crossover and mutation operators. Moreover, as compared to original elitist strategy, the overlapping populations prove to perform better \cite{Nagata2004paper104, Whitley1996paper108} for both GA and ES evolutionary algorithms. Finally, even though elitism has proven effective and widely accepted, Whitley et al. \cite{Whitley1996paper108} cautioned that the strategy might be a poor choice when dealing with problems having dynamically changing landscape.

\section[Parameterization in Evolutionary Computations]{Parameterization in Evolutionary\\ Computations}
It can be noticed from the foregoing that unlike the gradient-based optimization techniques, evolutionary computation algorithms are involved with a huge number of parameters upon which their overall success relies. Worth noting is that because the performance of an evolutionary algorithm heavily relies on the correct settings of these parameters, tuning the parameters of a simple GA to their optimal values is by itself a complex optimization problem. Thus, \cite{Grefenstette1986paper94} who wishes to derive the optimal parameters of a genetic algorithm for some specified problems put forward a meta-level GA. The design consists of an internal user parameterized genetic algorithm that is used to tune the parameters of a main genetic algorithm. 

The parameters of a typical genetic algorithm can be classified into two main categories

\begin{enumerate}[i.]
\item \textbf{Structural Parameters}: These include choice of data type (i.e. representation scheme) and the types of genetic operators used.
	\item \textbf{Numerical Parameters}: These include but not limited to the population size, probabilities of mutation and crossover, maximum number of generations and so on.
\end{enumerate}

\subsection{The Standard Parameter Settings}
Early works in genetic algorithms by Holland \cite{Holland1975} and later by De Jong \cite{DeJong1975} have led to the development of the most widely used \emph{standard parameter} sets. Although mainly obtained from empirical experiments, many other theoretical studies \cite{Back1993paper111, Goldberg1991paper93, Boyabaltli2007paper95} have reinforced the validity of these standard parameters. 

The space of a simple genetic algorithm comprises of at least six basic parameters depicted in table \ref{GAParametersTab}. The table shows the ranges and commonly used types of the standard parameters for small to medium sized global optimization problems.

\begin{table}
\begin{center}
\caption[Standard Parameters of a Typical GA]{Standard Parameters of a Typical Genetic Algorithm}
\label{GAParametersTab}
\begin{tabular}{lcp{178.54pt}}
\toprule
Parameter Name          & Symbol  & Typical Values/Types\\
\midrule
Population Size         & $N$     & $[50-200]$\\
Selection Scheme        & --      &  FPS, Ranking, Tournament, etc.\\
Crossover Probability   & $P_{c}$ & $[0.6-1.0]$ \\
Mutation Probability    & $P_{m}$ & $[0.01-0.05]$ or $\nicefrac{1}{l}$ where $l=$ string length \\
Replacement Scheme      & --      & Generational, Elitist, Overlap, etc. \\
Termination Criteria    & --      & Maximum limits on: Runtime, Generations/Evaluations or any heuristic convergence measure. \\
\bottomrule
\end{tabular}
\end{center}
\end{table}




\subsection{Adaptive and Dynamic Parameters in EC}
In order to alleviate the need for tuning and fine tuning of GA parameters as the nature of problems or optimization goals vary, several proposals of adaptive, self-adaptive or dynamic parameters are made. A proposal by \cite{Wu2009paper70} suggest adapting the probabilities of crossover and mutation based on the population entropy at any stage of the evolution. They described the population entropy as the distribution of individuals in the population. The greater is the difference among the individuals in a population, the higher the entropy and vice versa. The idea is to avoid loss of diversity by adaptively raising the probabilities of crossover and mutation as the population entropy drops. In a similar trend, \cite{Fogarty1989} use varying mutation rates and argue that mutation rates that decrease exponentially as the population average fitness grows over generations have superior performance over their non-adaptive counterparts. 

In a different respect, Niehaus et al. \cite{Niehaus2001paper112} adopt three new methods of dynamically adjusting the probabilities of reproduction operators in GA. The aim was to establish an adaptive system that can not only perform better than randomly chosen settings, but compete with the empirically proven standard settings. The adaptation methods used were: population-level dynamic probabilities, individual-level dynamic probabilities and fitness-based dynamic probabilities. An earlier work by \cite{Davis1989} exhibited similar argument. Elsewhere, Spears \cite{Spears1995paper113} studies the role of crossover in GA, EP and other fields of evolutionary computation and noted that despite the wealth of theoretical analysis, it is sometimes difficult to decide a priori which type of crossover to use and at what rate. Spears then suggests an adaptive mechanism for controlling the use of crossover operator in evolutionary algorithms and argue that the method can also be used to enhance the performance of non-adaptive EAs. Detailed background on various evolutionary algorithms with adaptive and dynamic operators can be found in \cite{Hong2002paper2, Spears1995paper113}.

\subsection{A Parameter-less GA}
From a somewhat extreme angle of parameter adaptation in evolutionary algorithms, Harik et al. \cite{Harik1999paper83} noted that majority of the target end users of EAs barely have sufficient understanding of the EA dynamics. Thus, users may lack the knowhow on manual tuning of its parameters for optimal performance. They argued that having an evolutionary algorithm showcased as a black-box is necessary to relieve its user from having to care about the required population size, crossover or mutation probabilities or the type of selection scheme. They proposed self adaption of the population size by starting multiple instances of a recombinative (i.e. crossover-only) GA running with different and increasing population sizes. The schema theory of building-blocks was used to adapt the crossover probabilities. 

Although it is quite a remarkable idea, the proposed technique can severely increase the computational cost as a result of naively running many instance of the GA at all times. Moreover, the idea used to adapt the crossover probability is based on the Holland's building-block theory the soundness of which cannot be justified by many EA theorists \cite{ReevesRow2004GABookPrincplnPers}. Finally, the adaptation is limited to crossover operator without addressing the mutation aspect. 

To sum of, some researchers have a contrary view to the original conception of GAs as been problem-independent. For instance, Boyabaltli et al. \cite{Boyabaltli2007paper95} argue that if viewed from the parameterization perspective, GAs can be very problem specific algorithms. We would like to emphasize here that this is not the view of many EA practitioners. Results from several empirical and analytical experiments \cite{Grefenstette1986paper94} have shown that the parameters of a GA need not to be overly tuned to suit a particular problem, the algorithm can perform fairly optimal for much wider range of problems so long as its parameters are set to within the range of the standard parameter settings. 

\section{The Proposed Adaptive Elitism}
\label{ProposedAdaptElitism}

Considering the fact that GAs using the elitist strategy always perform better than most of the generational GAs. However, as described previously, with basic elitism, the single elite individual may lack influence on the fitness growth in the population (i.e., scalability problem). Also, increasing the size of the elite population may lead to rapid loss of diversity in the evolution. Therefore, we propose a new variant of elitist strategy, an adaptive overlapping population having a variable generation gap that is some percentage ($G\%$) of the total population size $N$. We will henceforth refer to this parameter as \textit{adaptive overlap size}. The technique ensures that up to $G\%$ of the best individuals in the parent population always scale to the next generation so long as the following two conditions are concurrently met. A maximization problem is assumed.

\begin{enumerate}[i.]
	\item If the population average fitness value $\overline{f}(t)$ in the current generation $t$ is no better than that at the previous generation $(t-1)$, and;
	\item If the population fitness variance $Varf(t)$ in the current generation $t$ is no less than that of the previous generation $(t-1)$.
\end{enumerate}
The procedure in algorithm \ref{Algorithm2} elaborates the proposed scheme. 

\begin{algorithm}
\caption{The Proposed Adaptive Elitism}
\label{Algorithm2}
\begin{algorithmic}[1] 
\STATE $\textbf{begin}$\\
\STATE $\qquad$ $t\leftarrow 0$\\
\STATE $\qquad$ $N\leftarrow$ size of $P(t)$\\
\STATE $\qquad$ $N_{Elite}\leftarrow G\%\textrm{ of }N$\\
\STATE $\qquad$ initialize $P(t)$\\
\STATE $\qquad$ $f_{P}(t)\leftarrow$ evaluate and rank $P(t)$\\
\STATE $\qquad$ $P_{Elite}(t)\leftarrow$ Top $N_{Elite}$ of $P(t)$\\
\STATE $\qquad$ $\overline{f}_{p}(t)\leftarrow$ average of $f_{P}(t)$\\
\STATE $\qquad$ $Varf_{P}(t)\leftarrow$ variance of $f_{P}(t)$\\
\STATE 	$\qquad$ $\textbf{while}$ $\textit{not converged}$ $\textbf{do}$\\
		\STATE $\qquad \qquad$ $Q(t)\leftarrow$ evolve $P(t)$\\
		\STATE $\qquad \qquad$ $f_{Q}(t)\leftarrow$ evaluate and rank $Q(t)$\\
		\STATE $\qquad \qquad$ $\overline{f}_{Q}(t)\leftarrow$ average of $f_{Q}(t)$\\
		\STATE $\qquad \qquad$ $Varf_{Q}(t)\leftarrow$ variance of $f_{Q}(t)$\\
			\STATE 	$\qquad \qquad$ $\textbf{if}$ {$\overline{f}_{Q}(t)>\overline{f}_{P}(t) \quad \textbf{and} \quad  Varf_{Q}(t)> Varf_{P}(t)$} 			$\textbf{then}$
			\STATE $\qquad \qquad \qquad$ $N_{Elite}\leftarrow G\%\textrm{ of }N_{Elite}$\\
			\STATE $\qquad \qquad \qquad$ $P_{Elite}(t)\leftarrow$ Top $N_{Elite}$ of $P(t)$\\
			\STATE $\qquad \qquad \qquad$ $P(t+1)\leftarrow Q(t)\cup P_{Elite}(t)$\\		
		\STATE $\qquad \qquad$ $\textbf{else}$
			\STATE $\qquad \qquad \qquad$ $P(t+1)\leftarrow Q(t)\cup P_{Elite}(t)$\\
		\STATE $\qquad \qquad$ $\textbf{end if}$
		\STATE $\qquad \qquad$ $t\leftarrow t+1$\\
\STATE	$\qquad$ $\textbf{end while}$
\STATE$\textbf{end}$\\
\end{algorithmic}
\end{algorithm}


Notice from algorithm \ref{Algorithm2} that the size of the elite population is initially set to some percentage (i.e. $G\%$) of the original population size (line $4$). This however gets continuously halved at the end of every generation so long as the population's average fitness grows and its fitness variance does not shrink (line $15-21$). After the initialization stages (lines: $2-5$), the initial population is evaluated and ranked based on the fitness of the candidate solutions (line $6$). The elite population constitute the top best individuals in the parent population (line $7$). Lines ($8-9$) compute and store the average fitness and fitness variance of the current population. Thereafter, the evolution process (line $11$) which includes the selection, crossover and mutation continues iteratively until some convergence measure is satisfied (lines $10-23$).

The proposed idea of the adaptive elitism has two main goals. First, it safeguards individuals with high fitness values from been lost due to potential catastrophic effect of genetic operators at early stages of the evolution. Second, it ensures maintenance of useful diversity by adaptively shrinking the size of the elite population (i.e. the generation gap) as the quality of the population improves over generations. A generation gap of $5\%$  was chosen following empirical experiments on some global optimization test problems with various population sizes. Because a strict binary tournament selection method is adopted, the successive halving of the elite population size will not stop the best individual from surviving across generations. Ultimately, optimum level of selection pressure can be maintained and the risk of premature convergence to sub-optimal regions of the search space will be avoided.

\section{Contributions}

In addition to the wide survey and analysis on small and major aspects of evolutionary computation algorithms, the chapter gives an in-depth treatment on parameterization issues upon which development of successful global optimization methods relied. Thereafter, a new replacement technique (adaptive elitism) aimed at enhancing evolutionary algorithms to efficiently exploit high quality areas of the search space without compromising exploration of other potentially viable regions was developed. Ultimately, the insight derived from this investigation could pave a way for further improvements in the design of not only efficient, but robust EAs.

\section{Remarks}
This chapter commenced with an overview of the historical foundation and investigates the development and simulation of evolutionary computations. Key initialization aspects such as problem \emph{representation}, creation and sizing of the \emph{initial population} are reviewed. Major parameterization issues such as the choice of \emph{selection scheme} and \emph{replacement strategy}, the type of crossover and mutation operators and their probabilities are investigated. Finally, the chapter presented a novel adaptive elitism technique designed to enhance the ability of evolutionary algorithms to fully exploit the promising regions of their search space. In the next chapter, various stopping criteria of genetic algorithms will be investigated and the convergence analysis of these algorithms will be examined.

\singlespacing
\bibliographystyle{ieeetr}





\chapter[Convergence Analysis in EC Algorithms]{Convergence Measurement in EC using Price's Equation}
\label{ConvergenceAnalysisChap}

This chapter will focus on investigating the convergence of evolutionary algorithms. After a brief introduction, some commonly used stopping and convergence measures will be reviewed. Price's theorem will be examined and the use of the extended Price's equation to view the effect of genetic operators on evolution progress will be investigated. Finally, a novel convergence measure built on the extended Price's equation will be presented.

\section{The Need for Convergence Measure in EC Algorithms}

Subsequent to the earlier investigation of the dynamics of evolutionary computations, it is imperative to carefully assess the long-term behaviour of these algorithms. Evolutionary computations are naturally inspired stochastic algorithms that by design are capable of running perpetually. However, for the purpose of solving optimization problems, it is always necessary to prescribe a set of definitive stopping criteria that if satisfied, the process could be brought to a halt. Typically, at some stage of evolution, the search progress slows and the quality of the candidate solutions barely improve, this is a signal for convergence and majority of the convergence measures are designed to detect such conditions. After the search process is halted, the best candidate solution obtained so far is returned as the supposedly \textit{optimal} solution.

\section{Conventional Measures of EC Convergence}

Beside terminating evolution based on some user prescribed limits on evaluations, generations or execution time for a run, more sophisticated convergence measures that are mainly based on population diversity are used to automatically terminate the evolution. In this respect, population diversity could relate to the genotype, phenotype or even the average fitness of the population directly. A common practice is to measure the diversity by assessing the similarity among the candidate solutions in the genotype space using a distance measure \cite{LouisRawlins1993paper115} such as the Hamming distance. This agrees well with the notion that GA converges when the candidate solutions in the population become identical.

While the convergence characteristics of all evolutionary algorithms rely on several factors, the most important factors are the types of selection mechanism, the reproduction operators and the size of the population of candidate solutions. The selection operation being an exploitation process, it mainly favours the fitter individuals at the expense of the weaker ones and is seen as the major derive to convergence in EC. Thierens et al. \cite{Thierens1994paper13} use the normal fitness distribution method to model the convergence characteristics of various selection methods. Simple yet elegant ordinary differential equations (ode) models that estimate the true convergence behaviours of the fitness proportional selection (FPS), tournament selection, truncation selection and elitist recombination were developed and the convergence speed was related to the takeover time\footnote{Takeover time is the number of generations needed for the best individual in the population to take over the entire population.} of the associated selection method. They argue that FPS selection schemes have the longest convergence time and can further slow down as the search approaches optimal solution.

Regarding the manner in which the reproduction operators influence GA convergence, it is important to realize that besides the types of the operators, the frequency of their application also play a major role. A new diversity measure that estimates the average Hamming distance in the population was proposed by \cite{LouisRawlins1993paper115}. They noted that while mutation and its probability of application can severely influence the convergence rate, vast majority of the traditional crossover operators such as $n-$point, uniform and punctuated crossover have little effect on GA convergence. They analytically prove that the new diversity measure (i.e., average Hamming distance) is not affected by crossover operators but mostly influenced by the selection mechanism. The measure was then used to predict the average Hamming distance at convergence and thus define an upper bound limit on the evolution run time.

Similarly, Sharapov et al. \cite{Sharapov2006paper14} derive the mean convergence rate of genetic algorithms due to various reproduction operators. Although no numerical experimentation was conducted, probabilistic models were used to theoretically analyse the convergence characteristics due to crossover, mutation and inversion operators.

Elsewhere, Greenhalgh et al. \cite{Greenhalgh2000paper12} investigate convergence on the ground of population's fitness. They consider the maximum iteration limit, on-line and off-line performances as the three basic stopping criteria and convergence measures for genetic algorithms. The on-line performance reflects the average of all fitness functions evaluated up to and including the current trial, while the off-line performance is the running average of the best fitness value to a particular time. They use Markov chains to determine an upper bound on the maximum iterations limit that will guarantee convergence to global optimum with certain level of confidence based on a predefined probability.  


Hybridization approaches also play vital role to improving GA convergence rate. As reported by Miura et al. \cite{Miura2000paper41}, for most nonlinear optimization problems, the time required by a genetic algorithm to converge to an optimal solution can be reduced by incorporating some information about the gradients of the problem's variables. Thus, they propose hybridizing GA with a steepest descent algorithm and argue that this can positively influence the overall convergence rate.  Similar proposals were made by \cite{Isaacs2007paper35, Dawei2010paper51, Wang2008paper54, IlSeok2004paper56, Hernandez2008paper71} where genetic algorithms are combined (in various ways) with various local search methods to speed up convergence. Nearly $200\%$ improvement in the convergence rate was reported by Kaur et al. \cite{Kaur2008paper68} as a result of hybridizing a genetic algorithm with a nearest neighbour search in solving a TSP combinatorial optimization problem.

Worth mentioning is that all the aforementioned approaches for estimating GA convergence share their merits and demerits. For instance, in a typical binary coded GA, monitoring the similarity among chromosomes using a distance measure such as Hamming distance can allow effective analysis of the diversity profile in a given population. This is backed by the assertion that whenever evolutionary search nears convergence, its search space draws near a Hamming landscape. However, we observe that this similarity measure works in the genotype space and lacks any knowledge of the dynamics in the solution space. Thus, depending on the mapping technique, individuals that seem very similar in the genotype space may differ significantly in the phenotype space. In other words, a binary represented chromosome \mbox{$a_{1}=10110111$} (on hyperplane \mbox{$10\star\star\star\star\star\star$}) looks more similar to chromosome \mbox{$a_{2}=01110111$} (which is on an entirely different hyperplane \mbox{$01\star\star\star\star\star\star$}) than chromosome \mbox{$a_{3}=10\star\star\star\star00$} that shares the same hyperplane.

We further observe that estimating GA convergence by marrying the two diversity measures that monitor similarity in both the encoding and solution spaces might not totally eliminate dissension between the methods which could trigger false alarms for convergence. Moreover, the combination can severely increase the computational cost of the convergence detection process. 

We therefore argue that since progress in evolutionary search is internally governed by the actions of genetic operators (selection and reproduction operators like crossover and mutation), building a convergence measure that assesses the activities of these operators can be a promising alternative to tackling this dilemma. Thus, a proposal for utilizing the extended Price's equation to investigate the individual role of genetic operators to the progress of evolution will be presented in the next section. A new convergence measure will then be proposed.

\section{Monitoring the Effect of EC Operators with Price's Equation}
\label{PricesEqnSec}
In order to examine the individual effect of evolution operators while in interaction, George Price \cite{Price1972} formulated a theorem that permits decomposition of the evolutionary process to separate the genetic effect (or contribution) of the selection operator from that of other reproduction operators (i.e. crossover and mutation).  Although Price's work was mainly in the field of evolutionary genetics, the proposed equation \eqref{PricesEqn} provides greater insight into general selection processes beyond the originally intended genetical selection.

Price's equation states that:
\begin{equation}
\label{PricesEqn}
\Delta Q=\frac{\texttt{Cov}(\textbf{z},\textbf{q})}{\bar{z}}+\frac{\sum_{i=1}^{N}z_{i}\Delta q_{i}}{N\bar{z}}
\end{equation}
where $\Delta Q=Q_{2}-Q_{1}$ is the change in the measured characteristics (such as fitness), $N$ is the number of individuals in the parent population (i.e., population size), $z_{i}$ is the number of offspring of parent $i$, and $\bar{z}=\sum_{i}z_{i}/N$ is the average number of the offspring produced. Also, $\Delta q_{i}=q'_{i}-q_{i}$ where  $q_{i}$ is the fitness of parent $i$ and $q'_{i}$ is the average value of the measured $q_{i}$ in the offspring of parent $i$. And finally, $\textbf{z}$ and $\textbf{q}$ are vectors of $z_{i}$ and $q_{i}$ respectively.

The two terms in the Price's equation \eqref{PricesEqn} represent the contribution of different operators as the mean of the characteristic being measured. Since we deploy the Price's equation to analyse convergence in EC, the measured characteristic $Q$ in this case will be fitness. The first term represents the contribution of selection operator while the second term gives an aggregated contribution of the reproduction operators involved. Notice that the effect due to selection is modelled in terms of the covariance between the individuals $\textbf{z}$ and their fitness $\textbf{q}$. This conforms to the Fisher's fundamental theory of natural selection that relates the change in the mean fitness in a population to the population's fitness variance. It also agrees with the argument presented by \cite{Frank1997} that sees the covariance between the phenotypic values of individuals and their fitness as the cause of differential productivity that leads to the change in phenotype.

The following lemmas 1 and 2 will elaborate how Price's equation decomposes the fitness progress into separate contributions from the selection operator and other reproduction operators.
\begin{description}
\item [{Lemma 1:}]Supposing the fitness $q_{i}\in\mathbb{R}$ of each member of the parent population in equation \eqref{PricesEqn} is represented by a vector $\textbf{q}$ such that:
\begin{equation}
\textbf{q}=\left[q_{1},q_{2},...,q_{N}\right]^{T}
\end{equation}
Suppose also the number of offspring $z_{i}\in\mathbb{Z}$ produced by each one of the $N$ parent is represented by a vector $\textbf{z}$ such that:
\begin{equation}
\textbf{z}=\left[z_{1},z_{2},...,z_{N}\right]^{T}
\end{equation}
Then, the following expansion of the two terms in equation \eqref{PricesEqn} will demonstrate that it is only the first term that represents the contribution of the selection operator.

\textit{First term}:
\begin{equation}
\label{1stTermPriceEqn}
\frac{\texttt{Cov}(z,q)}{\bar{z}}=\texttt{mean}\left[\begin{array}{cc}
\left(z_{1}-\mu_{z_{1}}\right) \left(q_{1}-\mu_{q_{1}}\right)\\
\left(z_{2}-\mu_{z_{2}}\right) \left(q_{2}-\mu_{q_{2}}\right)\\
\vdots\\
\left(z_{N}-\mu_{z_{N}}\right) \left(q_{N}-\mu_{q_{N}}\right)\\
\end{array}\right]/\bar{z}
\end{equation}

where $\mu_{z_{i}}$ is the mean of the number of offspring of parent $i$ and $\mu_{q_{i}}$ is the mean fitness of the parents.

\textit{Second term}:
\begin{equation}
\label{2ndTermPriceEqn}
\frac{\sum_{i=1}^{N} z_{i}\Delta q_{i}}{N\bar{z}}=\frac{z_{1}(q'_{1}-q_{1})+z_{2}(q'_{2}-q_{2})+\cdots+z_{N}(q'_{N}-q_{N})}{N\bar{z}}
\end{equation}
Recall that following any typical evolutionary selection process, the resulting offspring have the same fitness as their parents (i.e. selection process adds no new solutions to the population), therefore, 
\begin{equation}
\Delta q_{i}=q'_{i}-q_{i}=0.
\end{equation}
Consequently, the value of equation \eqref{2ndTermPriceEqn} (i.e., the second term of Price's equation) sums to zero. Therefore, the contribution of selection operator is only in the first term of the Price's equation.
\begin{flushright}
$\Box$
\end{flushright}

\item [{Lemma 2:}]Unlike in the above case for selection process, the following examination of the Price's equation \eqref{PricesEqn} will show that the second term of the Price's equation uniquely represents the contribution of the reproduction operators.

Assuming any traditional $n-$point crossover operator is employed; crossing any two parents yields two offspring. Therefore, if all parents undergo the crossover operation (i.e. $P_{c}=1$) and the population size $N$ is even, then, the number of offspring produced will always be equal to the number of their parents, i.e., $\mu_{z_{i}}=z_{i}$. Hence, the first term of the Price's equation, i.e. equation \eqref{1stTermPriceEqn} is equal to zero. This is also true for any bit-flip mutation operation where a parent chromosome yields a single offspring after mutation.

A worth noting exception here is that if the population size $N$ is odd or in the case where not all chromosomes undergo the crossover operation (i.e., when $P_{c}<1$), then $\mu_{z_{i}}\neq z_{i}$ and thus $\frac{\texttt{Cov}(z,q)}{\bar{z}}\neq0$. This is contrary to the traditional notion of decomposing Price's equation in the literature \cite{Bassett2005paper4} where it is often assumed that the contribution of the reproduction operators in the first term of Price's equation is \textit{always} zero. This investigation reveals the contrary if taking into account the special cases mentioned above.

Now, expanding the second term as in equation \eqref{2ndTermPriceEqn} reveals that the summation is non-zero since the fitness $q_{i}$ of any parent $i$ is different from that of all its offspring\footnote{Although it is possible for any or all the offspring resulting from crossover to have equal fitness as their parent, this is rare and usually only occur when the population diversity collapses. Thus it signals convergence.}, and hence, $q'_{i}\neq q_{i}$ (at least in most cases). Hence the contribution of crossover and mutation operators is mainly from the second term of Price's equation though it may include the first term in the exceptional situations noted above.
\begin{flushright}
$\Box$
\end{flushright}
\end{description}

\subsection{Extension of Price's Equation}

Bassett et al. \cite{Bassett2004paper5, Bassett2003paper3} extend the second term \eqref{2ndTermPriceEqn} of Price's equation to allow monitoring the contribution of individual reproduction operators so as to ascertain which among them is more effective at various stages of the evolution process. It was argued that utilizing fitness averages alone will not guarantee accurate conclusions on the convergence of the evolution. This is because the average individuals are not the major driving force for the evolution. Thus, it become essential to dynamically analyse the fitness variance throughout the population paying attention to the exceptional (\textit{best} and \textit{worst}) individuals mainly created by the reproduction operators.

The extended Price's equation can be defined as:
\begin{equation}
\label{ExtPricesEqn}
\Delta Q=\frac{\texttt{Cov}(\textbf{z},\textbf{q})}{\bar{z}}+\sum_{j=1}^{k}\frac{\sum_{i=1}^{N}z_{i}\Delta q_{ij}}{N\bar{z}}; k=1,2,...
\end{equation}
where $k$ is the number of genetic operators, $q_{ij}$ is the average value of the fitness of all the offspring of parent $i$ after the application of operator $j$, and $\Delta q_{ij}=q'_{ij}-q_{i(j-1)}$ is the difference between the average value $q$ (i.e., fitness) of the offspring of parent $i$ measured before and after the application of operator $j$. 

For the proposed hybrid evolution algorithm, we deploy only the mutation and crossover operators, thus, $k=2$. Therefore, the following proposed extension to the Price's equation \eqref{ProposedExtPricesEqn} contains only three terms; a term for the selection, crossover and mutation respectively.
\begin{equation}
\label{ProposedExtPricesEqn}
\Delta Q=\frac{\texttt{Cov}(\textbf{z},\textbf{q})}{\bar{z}}+\frac{\sum_{i=1}^{N}z_{i}\Delta q_{iCrossover}}{N\bar{z}}+\frac{\sum_{i=1}^{N}z_{i}\Delta q_{iMutation}}{N\bar{z}}
\end{equation}
Hence, each of the terms in this equation estimates the changes in the mean of the population's fitness due to one of the three genetic operators.

\section{Analysing Evolution Progress with Extended Price's Equation}
In order to analyse the effect of the proposed extension of Price's equation \eqref{ProposedExtPricesEqn}, we conduct a number of experiments with a genetic algorithm having some standard parameters previously highlighted in table \ref{GAParametersTab} and some newly proposed parameters previously described in section \ref{ProposedAdaptElitism}. Table \ref{GAParamForPricesExpTab} shows the detailed parameter settings for these experiments. Various multidimensional global optimization problems are used. However, the results that will be presented here{\footnote{Similar results are obtained with many other global optimization benchmark functions, and therefore omitted.}} are those obtained on simulating the solution process of the well known Schwefel function previously described in \eqref{SchwefelFunction}.

\begin{table}[htb]
\begin{center}
\caption[GA Parameter Settings for the proposed Price's Experiment]{GA Parameter Settings for Investigating the Proposed Extended Price's Equation}
\label{GAParamForPricesExpTab}
\begin{tabular}{lcp{222.54pt}}
\toprule
\textbf{Parameter Name} & \textbf{Symbol} & \textbf{Typical Values/Types}\\
\midrule
Population Size         & $N$     & $50 - 100$\\
Representation          & --      & Binary encoding\\
Selection Scheme        & --      & Roulette wheel and Binary Tournament selection without \mbox{replacement}\\
Crossover Probability   & $P_{c}$ & $1.0$ or $100\%$ \\
Mutation Probability    & $P_{m}$ & $\nicefrac{1}{l}$ where $l=$string length \\
Replacement Scheme      & --      & The Proposed Adaptive Elitist, see section \ref{ProposedAdaptElitism} \\
Adaptive overlap size   & $G$     & $0.05$ i.e. $5\%$ of population size $N$\\
Termination Criteria    & MaxGen  & Maximum generations limit $100$ \\
\bottomrule
\end{tabular}
\end{center}
\end{table}

The objectives for this experiment are:
\begin{itemize}
\item  To observe and analyse the growth in the fitness of the population of candidate solutions as the evolution progresses; and
\item •	To investigate the change in fitness that can be attributed to the selection operator, and changes due to the crossover and the mutation operator respectively.
\end{itemize}
Ultimately, the outcome of this experiment is aimed at providing better insight through observation of the manner in which the selection, crossover and mutation operators collectively move the evolutionary search forward. It will also give insight on their individual effect on ensuring sustainable balance for the exploitation and exploration of problem search space, making GAs effective and promising global search mechanisms.

\begin{description}
\item[{Experiment 1:}] Visualizing Evolution with Roulette Wheel Selection (RWS)\\
In this experiment, RWS is used to investigate the individual effects of the genetic operators over the entire period of evolution. Table \ref{PricesExperiment1and2ParamTab} shows the various parameter sets for this experiment.
\item[{Experiment 2:}] Visualizing Evolution with Binary Tournament Selection (BTS)\\
Here a binary tournament selection without replacement is used to investigate the effects of the genetic operators during the evolution process. Similar parameter sets are used as in table \ref{PricesExperiment1and2ParamTab}.
\end{description}

Notice from table \ref{PricesExperiment1and2ParamTab} that there are six parameter sets for testing each of the two selection methods. The first half will be tested with a population size of $50$ while for the other half, a population size of $100$ will be used. Note that with the parameter sets in tables \ref{GAParamForPricesExpTab} and \ref{PricesExperiment1and2ParamTab}, a population size of $50$ is \textit{fairly} sufficient for the proposed GA to converge to the optimum of the 2-dimensional Schwefel problem \eqref{SchwefelFunction} used in the test. However, the motive here is to investigate whether changes in population size has any significant effect on the behaviour\footnote{The contribuion of the genetic operators on the fitness growth defined in the extended Price's equation \eqref{ProposedExtPricesEqn}.} of the genetic operators during evolution. Finally, in these experiments all the six simulation runs were carried out with a maximum generation limit of $100$ and all results are averaged over $100$ runs for enhanced statistical significance.

\begin{table}[htb]
\begin{center}
\caption[Parameter Settings for Experiments 1 \& 2 with RWS and BTS]{Parameter Settings for Experiments 1 \& 2 with RWS and BTS. $P_{c}$ and $P_{m}$ are crossover and  mutation probabilities respectively, MaxGen is the maximum generation limit and $l$ is the string length for the binary chromosome.}
\label{PricesExperiment1and2ParamTab}
\begin{tabular}{ccccc}
\toprule
Population Size $N$ & $P_{c}$ & $P_{m}$ & MaxGen  & Total Runs\\
\midrule

\multirow{3}{*}{50} & $1.0$  &  $1/l$   &  $100$  &  $100$\\
                    & $0.6$  &  $0.01$  &  $100$  &  $100$\\
                    & $0.7$  &  $0.05$  &  $100$  &  $100$\\ \midrule 

\multirow{3}{*}{100}& $1.0$  &  $1/l$   &  $100$  &  $100$\\
                    & $0.6$  &  $0.01$  &  $100$  &  $100$\\
                    & $0.7$  &  $0.05$  &  $100$  &  $100$\\
\bottomrule
\end{tabular}
\end{center}
\end{table}

\subsection{Results: Comparing RWS and BTS methods}
\label{ResultCompTournRWS}

The results obtained for the above experiments 1 and 2 are summarized in table \ref{PricesExperiment1and2ResultsTab}. The table compares RWS and BTS methods from two perspectives. First, on their performance under two different population sizes $N$ $(50$ and $100)$. Second, on their performance against three different settings for the probabilities of crossover $P_{c}$ and mutation $P_{m}$.

\begin{table}[htb]
\begin{center}
\caption[Comparison of RWS and BTS methods]{Results for Experiments 1 \& 2: Comparing RWS and BTS methods for three sets of crossover and  mutation probabilities $(P_{c}, P_{m})$ under increasing population size $N$. Maximum generation limit is $100$ and all results are obtained after averaging $100$ runs.}
\label{PricesExperiment1and2ResultsTab}
\begin{tabular}{cccccc}
\toprule
& \multicolumn{3}{c}{Test Setup}  & \multicolumn{2}{c}{Best Fitness values for:} \\ \midrule
Population Size $N$ & Test & $P_{c}$ & $P_{m}$ & RWS  & BTS\\
\midrule

\multirow{3}{*}{50}& (i)  & $1.0$  &  $1/l$   &  $794.76$  &  $\underline{834.14}$\\
                   & (ii) & $0.6$  &  $0.01$  &  $791.52$  &  $828.75$\\
                   & (iii)& $0.7$  &  $0.05$  &  $808.15$  &  $\underline{835.87}$\\ 
                    \midrule

\multirow{3}{*}{100}&(i)  &  $1.0$  &  $1/l$   &  $826.74$  &  $\textbf{837.93}$\\
                    &(ii) & $0.6$  &  $0.01$  &  $825.08$  &  $836.89$\\
                    &(iii)& $0.7$  &  $0.05$  &  $830.18$  &  $\textbf{837.93}$\\
\bottomrule
\end{tabular}
\end{center}
\end{table}

Notice from table \ref{PricesExperiment1and2ResultsTab} that for the two different population sizes, the results obtained with the BTS clearly outperform those when RWS selection method is used. Also, when BTS method is utilized, the results obtained with a population size of $50$ are fairly close to when the population size is doubled to $100$. Of key interest here is the fact that although setup (iii) generally appears to yield better results\footnote{It benefits from the high mutation rate of $0.05$, which increase the chances for further exploration of the problem space. This however may increase wasteful function evaluations.}, a careful comparison of the two bold face and two underlined values confirm that setup (i) is as good (especially when the population size $N=100$). 

Recall that the overall goal is to explore which among the setting will lead to highest fitness and at the same time ensure convergence detection with ease. Therefore, with the results presented in table \ref{PricesExperiment1and2ResultsTab} and the analysis of the following figures \ref{Expriment1and2FitnessAndPricesFigure} and \ref{Expriment1and2PricesSTDFigure} for the Price's plots, we infer that setup (i) (i.e., $P_{c}=1.0, P_{m}=1/l$ with BTS) has the following two key peculiarities: 
\begin{description}
\item[First:] With setup (i), convergence to a good approximation of the optimum solution is possible even with low population sizes. Thus, it is computationally cheap since the required function evaluations can be minimized.
\item[Second:] As will be seen from figures \ref{Expriment1and2FitnessAndPricesFigure} and \ref{Expriment1and2PricesSTDFigure}, setup (i) is more suitable for convergence detection via monitoring the effect of genetic operators.
\end{description}
Therefore, setup (i) is the best candidate for achieving the proposed objective.

\begin{figure}[hbtp]
	\centering

	\includegraphics[scale=0.75]{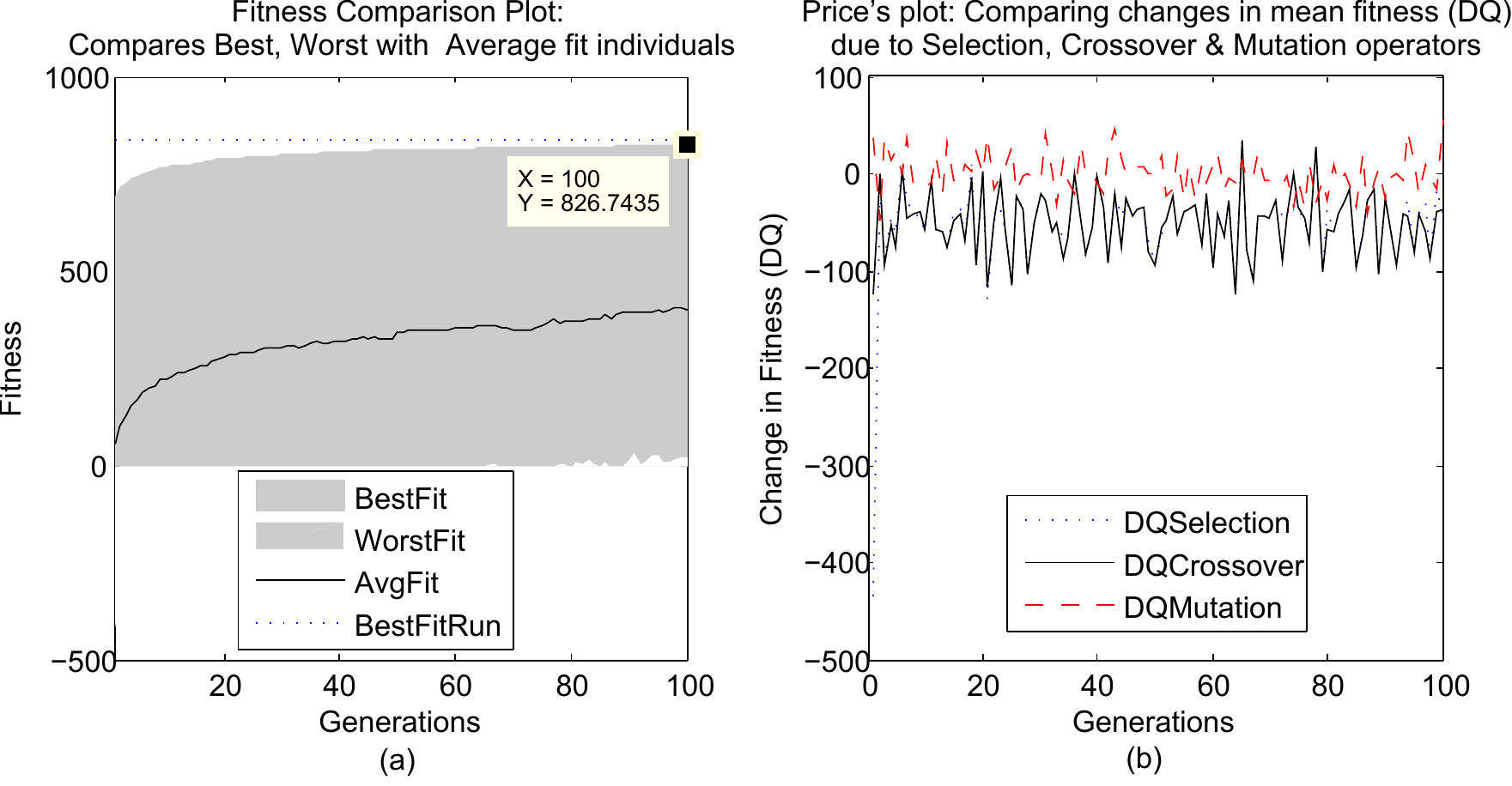}
	\includegraphics[scale=0.75]{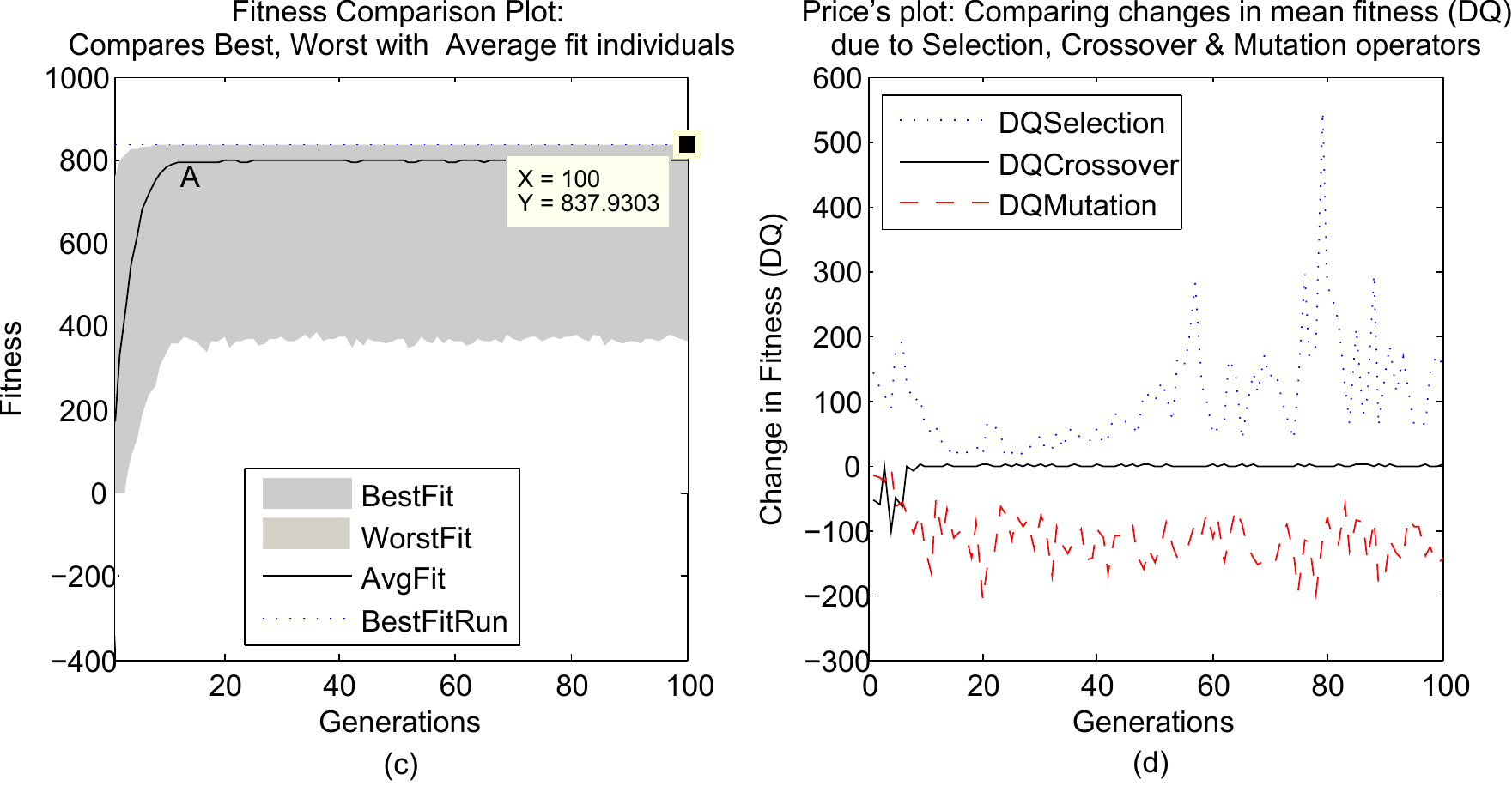}	
	
	\caption[Fitness comparison plots and the effect of genetic operators]{Simulation results: Experiment 1 (RWS): Plot (a) fitness comparison plot for the \textit{best}, \textit{worst} and \textit{average} fitness achieved over a maximum generation limit of $100$. Plot (b) is a comparison plot for the effect of genetic operators using extended Price's equation \eqref{ProposedExtPricesEqn} on maximization of a 2 dimensional Schwefel function with global optimum at $838$. Plots (c and d) show the corresponding results for Experiment 2 (BTS). For all plots $P_{c}=1.0, P_{m}=1/l$ and $N=100$ and results averaged over $100$ runs.}
	\label{Expriment1and2FitnessAndPricesFigure}
\end{figure}

The plots shown in figure \ref{Expriment1and2FitnessAndPricesFigure} are for setup (i) (see table \ref{PricesExperiment1and2ResultsTab}) where $P_{m}=1/l$ and $P_{c}=1.0$, the remaining setups are relatively similar and therefore figures omitted. The fitness comparison plots in figures \ref{Expriment1and2FitnessAndPricesFigure}(a and c) compare the \textit{best}, \textit{worst} and the \textit{average} fitness. The Price's plots (figures \ref{Expriment1and2FitnessAndPricesFigure}(b and d)) on the other hand depict the changes in the averages of the population's fitness due to each of the three genetic operators according to the definitions in equation \eqref{ProposedExtPricesEqn}, i.e., it shows the average contributions  of the selection, crossover and mutation operators to the fitness progress over the entire generations. All results are obtained after averaging 100 runs of the evolution.

\subsubsection*{Discussion of Results for Experiments 1: RWS selection}

A quick glimpse at the curves in the Price's plot in figure \ref{Expriment1and2FitnessAndPricesFigure}(b) for the contribution of operators barely allows any meaningful conclusion. One might naively infer that the idea of decomposing the fitness progress to investigate the individual effect of the genetic operators provides no additional information on the evolution dynamics. However, we would like to make the following observations:

\begin{enumerate}[i.]
\item It can be noticed from figure \ref{Expriment1and2FitnessAndPricesFigure}(b) that the curves for all the three operators fluctuate above and below zero on the fitness axis in a somewhat random manner. Thus, the plot provides no vital information as to which among the three operators ensures continues growth in fitness (i.e. exploitation) and which is responsible for maintenance of diversity in the population by exploring other areas of the search space.
\item A careful look at the initial shape of the curve for the selection operator in figure \ref{Expriment1and2FitnessAndPricesFigure}(b) reveals that there is a huge selection pressure at the early generations. This corresponds to the steep gradient in the \textit{best} and \textit{average} fitness curves (figure \ref{Expriment1and2FitnessAndPricesFigure}(a)) during the early generations. However, the selection pressure suddenly drops after around the first 10 generations and thereafter, it continues to fluctuate within a somewhat uniform range for the rest of the generations.  Consequently, the roles for the crossover and selection operator throughout this period remain quite undifferentiated. It is therefore hard to clearly identify which among the two operators is responsible for exploitation or exploration and at what stage of the evolution does it contribute most.
\item Another crucial observation is that because all the curves for the three operators in figure \ref{Expriment1and2FitnessAndPricesFigure}(b) continuously effect changes in the population's fitness, the population tends to slowly evolve thereby growing the \textit{average} fitness (see figure \ref{Expriment1and2FitnessAndPricesFigure}(a)) continuously (although at a slower rate) till the evolution is terminated. This unwanted phenomenon thwarts any possibility of detecting the precise moment at which the evolution converges.
\end{enumerate}
Nevertheless, the above behaviour of the selection operator agrees with the fact that RWS is a fitness proportionate selection method and as previously discussed in section \ref{FPSselectionSection}, it suffers inherent stochastic sampling errors\footnote{Some selection methods like Fitness proportionate and tournament (with replacement) due to being stochastic may lose the best member of the population by simply not picking it for any contest.}. These sampling errors increase as the \textit{average} fitness of the population grows. Thus, it is this phenomenon that led to the sudden collapse in the selection pressure just after the very early generations. 

Moreover, as can be seen from the Price's plot in figure \ref{Expriment1and2FitnessAndPricesFigure}(b), the contribution of the selection operator to the fitness progress becomes quite similar to (or no better than) that of the crossover and mutation operators. In other words, all the three operators play the role of exploration and exploitation of the search space. This effect may not be suitable for optimization purpose where the ultimate aim is to explore and as much as possible converge to the optimum solution point.

\subsubsection*{Discussion of Results for Experiments 2: BTS}

The plots shown in figure \ref{Expriment1and2FitnessAndPricesFigure}(c and d) convey similar information to those in experiment 1 above but now BTS is employed. Moreover, contrary to the results obtained in experiment 1, an examination of the actions of the individual operators, and comparing their interactions against the changes in fitness over generations in these plots led to the following deductions:

\begin{enumerate}[i.]
\item A simple observation of the fitness curves in figure \ref{Expriment1and2FitnessAndPricesFigure}(c) reveals that both the \textit{average} and the \textit{best} fit individuals in the population rapidly grow with a steep gradient to their peak values within the very early generations of the evolution. It is interesting to note how at the same time useful diversity is maintained in the population by retaining a reasonable amount of the \textit{worst} fit individuals throughout the evolution. This is a clear manifestation of achieving proper separation of roles among the three genetic operators as will be analysed shortly. Hence, while the BTS operator is undertaking its role of exploitation, mutation is ensuring effective exploration and the crossover operator serves as a regulator by performing both roles.
\item Notice from figure \ref{Expriment1and2FitnessAndPricesFigure}(d) that the curve for the selection operator is always above zero on the fitness axis. Thus, the selection operator can now be seen as a biased process that primarily guides the search towards the fitter individuals seen so far. It effectively drives the evolution towards converging to the promising regions of the search space thereby increasing the population's \textit{average} fitness. We therefore acknowledged that this behaviour conforms to the characteristics of a typical tournament selection method previously described in section \ref{TournamentSelectionSection}, and thus, it is as expected.
\item Notice also from the same figure that the curve for the crossover operator swing above and below zero on the fitness axis. This is indicative of the fact that the crossover operator has both exploration and exploitation effects. In other words, while crossover improves the population's average fitness via exploitation, it also lowers it during exploration. A worth noting observation is that the effect of crossover lessens over generations and eventually, neither its exploitation nor its exploration effect tend to influence the fitness growth in the population as convergence sets in.
\item Contrary to the previous operators, notice that the curve for the mutation operator always lie beneath zero on the fitness axis (see figure \ref{Expriment1and2FitnessAndPricesFigure}(d)). This is because mutation operator improves population's diversity by exploring other parts of the search space. But in so doing, it tends to drastically lower the fitness growth in the population. Moreover, the manner in which mutation affects the fitness growth is unaffected by convergence. Therefore, the effect of mutation on the overall fitness progress remains fairly constant over the entire run.
\end{enumerate}
Following the above observations, it is thought that among all the three genetic operators, it is the behaviour of the crossover operator with regard to the fitness growth in the population that is most crucial to effective detection of convergence of the evolution process. This is because both the mutation and selection operators continue to evolve throughout the evolution. Moreover, as highlighted in (ii) above, the selection operator is a mere biased process that favours highly fit solutions. It lacks the ability to add any new solution points into the population. It is therefore not suitable for convergence detection. 

Finally, for the same set of parameters (test setup (i)), it is interesting to notice how the \textit{best} and \textit{average} fitness curves in the fitness comparison plot in figure \ref{Expriment1and2FitnessAndPricesFigure}(c) (where BTS is used) clearly outperform those in figure \ref{Expriment1and2FitnessAndPricesFigure}(a) (where RWS is used) by converging rapidly towards the global optimum solution of $838$.

In the following section an investigation on how the crossover can be used to detect convergence will be presented.

\section[Using Extended Price's Equation to Measure Convergence]{Using Extended Price's Equation to Measure\\ Convergence}
\label{AutoConvergDetcSec}

Having decided, during the discussions of the previous section \ref{ResultCompTournRWS}, that the crossover operator can be utilized as a means for assessing convergence in the evolution, this section will focus on formulating and setting up the parameter for the convergence measurement. A graphical means for evaluating this parameter will also be developed.

Recall from equation \eqref{ProposedExtPricesEqn} that $\Delta Q$ is only the collective change in the average fitness due to the three genetic operators. As argued by \cite{Bassett2004paper5} observing changes in averages alone does not convey sufficient information on the true effect of an operator. This is evident from the plot of the crossover operator in figure \ref{Expriment1and2FitnessAndPricesFigure}(d) where it is hard to appreciate its effect on the fitness even during the early generations of the evolution. Therefore monitoring the spread in fitness in the population by exploring the best and worst individuals produced by an operator is a viable alternative that could yield better insight. The spread can be measured by evaluating the change in fitness variance in the population caused by application of the genetic operator. 

From equation \eqref{ProposedExtPricesEqn}, let the change in fitness due to a genetic operator $j$ (crossover in this case) be:
\begin{equation}
\label{CrossoverTermPricesEqn}
\Delta Q_{ij} = \frac{\sum_{i=1}^{N}z_{i}\Delta q_{ij}}{N\bar{z}}
\end{equation}
Then, since $\Delta q_{ij}=q'_{ij}-q_{ij}$ is the random variable of interest, the expectation and variance of the crossover term can be obtained by respectively taking the first and second moment of \eqref{CrossoverTermPricesEqn} with respect to the $\Delta q_{ij}$ values, i.e., the changes in fitness in the individuals of the population due to the crossover operator, such that:
\begin{equation}
E\left[\Delta q_{ij}\right] = \frac{\sum_{i=1}^{N}z_{i}\frac{\sum_{z_{i}}\Delta q_{ij}}{z_{i}}}{N\bar{z}}=\frac{\sum_{i=1}^{N}\sum_{z_{i}}\Delta q_{ij}}{N\bar{z}}
\end{equation}
where $\sum_{z_{i}}$ is the sum over all the children of parent $i$. Typically, there will be two children for each parent when a  crossover probability of $P_{c}=1.0$ is used. All other parameters are as previously defined in section \ref{PricesEqnSec}. The variance, $Var\left[\Delta q_{ij}\right]$ is:
\begin{equation}
Var\left[\Delta q_{ij}\right]=E\left[\Delta q_{ij}^{2}\right] -\left[E(\Delta q_{ij})\right]^{2}.
\end{equation}
Hence, the standard deviation is:
\begin{equation}
\sigma\left[\Delta q_{ij}\right]=\sqrt{Var\left[\Delta q_{ij}\right]}.
\end{equation}

\begin{figure}[hbtp]
	\centering

	\includegraphics[scale=0.75]{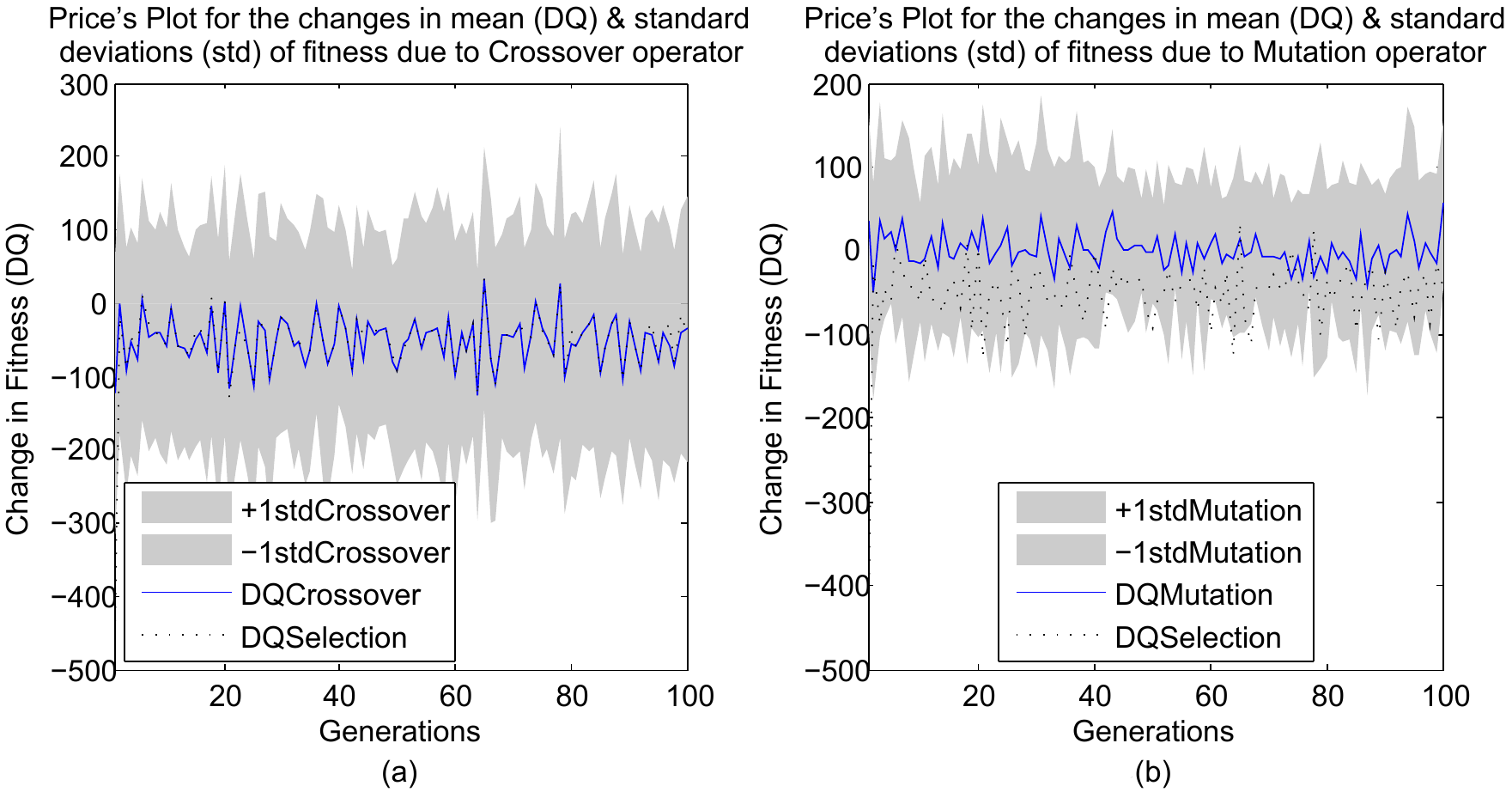}
	\includegraphics[scale=0.75]{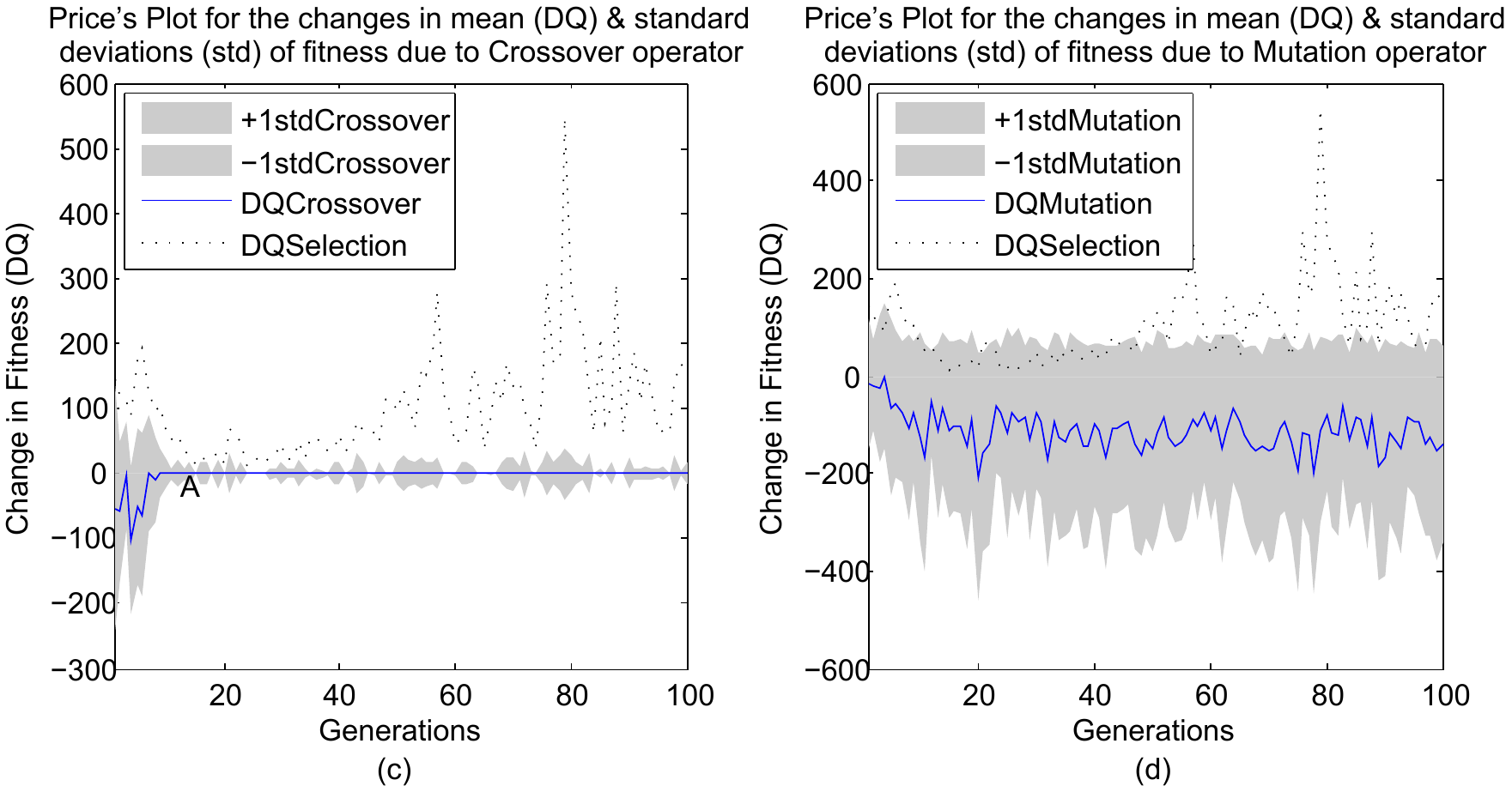}
	
	\caption[Visualizing $\pm \sigma$ (standard deviations) for the effect of genetic operators]{Visualizing one standard deviations for the effect of crossover and mutation operators: Experiment 1 (RWS): Plot (a) shows the \textit{average} and the $\pm \sigma$ (shaded areas) for the crossover operator. Plot (b) shows the \textit{average} and the $\pm \sigma$ (shaded areas) for the mutation operator. Plots (c and d) show the corresponding results for Experiment 2 (BTS). For all plots $P_{c}=1.0, P_{m}=1/l$ and $N=100$ and results averaged over $100$ runs.}
	\label{Expriment1and2PricesSTDFigure}
\end{figure}

Now, overlaying the curves for the one standard deviation interval ($\pm\sigma$) on the plot of the $\Delta Q$ changes in the \textit{average} fitness due to the crossover operator yield the envelope of the shaded areas shown in plots (a and c) of figure \ref{Expriment1and2PricesSTDFigure}. The plots give a better impression of the distribution of the effect of this operator on changes in fitness in the population over generations. A similar technique was used to generate the plot for the effect of mutation operator shown in plots (b and d) of the same figure.

Plots (a and b) in figure \ref{Expriment1and2PricesSTDFigure} show the results for experiment 1 where RWS is used. The shaded area above the curve for the change in the \textit{average} fitness due to crossover in figure \ref{Expriment1and2PricesSTDFigure}(a) reveal that crossover operator does indeed contribute to the fitness growth, but at the same time, the shaded area underneath it which lie under zero on the fitness axis indicates how much the operator contributed to producing low fit individuals. Notice how the $\pm\sigma$ (standard deviation) envelopes for both the crossover and mutation opeators (plots \ref{Expriment1and2PricesSTDFigure}(a and b)) remain uniformly constant throughout the period of the evolution. Hence, when RWS is utilized, the effects of both crossover and mutation on fitness growth continue till evolution is terminated.

Similarly, plots (c and d) of figure \ref{Expriment1and2PricesSTDFigure} show the results for experiment 2 where BTS is utilized. From the crossover plot (c) it is interesting to note how the envelope for the $\pm\sigma$ shrinks towards zero as the curve for the change in the \textit{average} fitness settles around zero on the fitness axis. This is indicative of the fact that beyond this stage, the crossover operator barely contribute to the fitness progress. Hence, it shows how the use of BTS method facilitates monitoring the effect of crossover operator to effectively detect loss of diversity in the population. Therefore, this justifies the previous supposition that crossover operator can be used to sufficiently detect convergence in EC.

On the other hand, an observation of plot \ref{Expriment1and2PricesSTDFigure}(d) reveals that the curve for the change in the \textit{average} fitness due to mutation operator lie and remain beneath zero on the fitness axis throughout the evolution. Also, larger portion of the shaded area for the $\pm\sigma$ envelope is beneath zero\footnote{It means mutation operator particularly ensures maintenance of diversity in the population by producing lower fit individuals when BTS is used.}. This is contrary to its behaviour when RWS is used (see figure \ref{Expriment1and2PricesSTDFigure}(b)) where mutation lacks precise role. Nevertheless, the standard deviation envelopes for the mutation operator in both plots (b and d) of figure \ref{Expriment1and2PricesSTDFigure} remain virtually uniform (without shrinking) throughout the evolution. This means the effect of mutation on the fitness of the population continues for the entire evolution run. Therefore, it cannot be deployed to conduct any convergence detection.

\begin{figure}[ph]
	\centering

\includegraphics[scale=0.75]{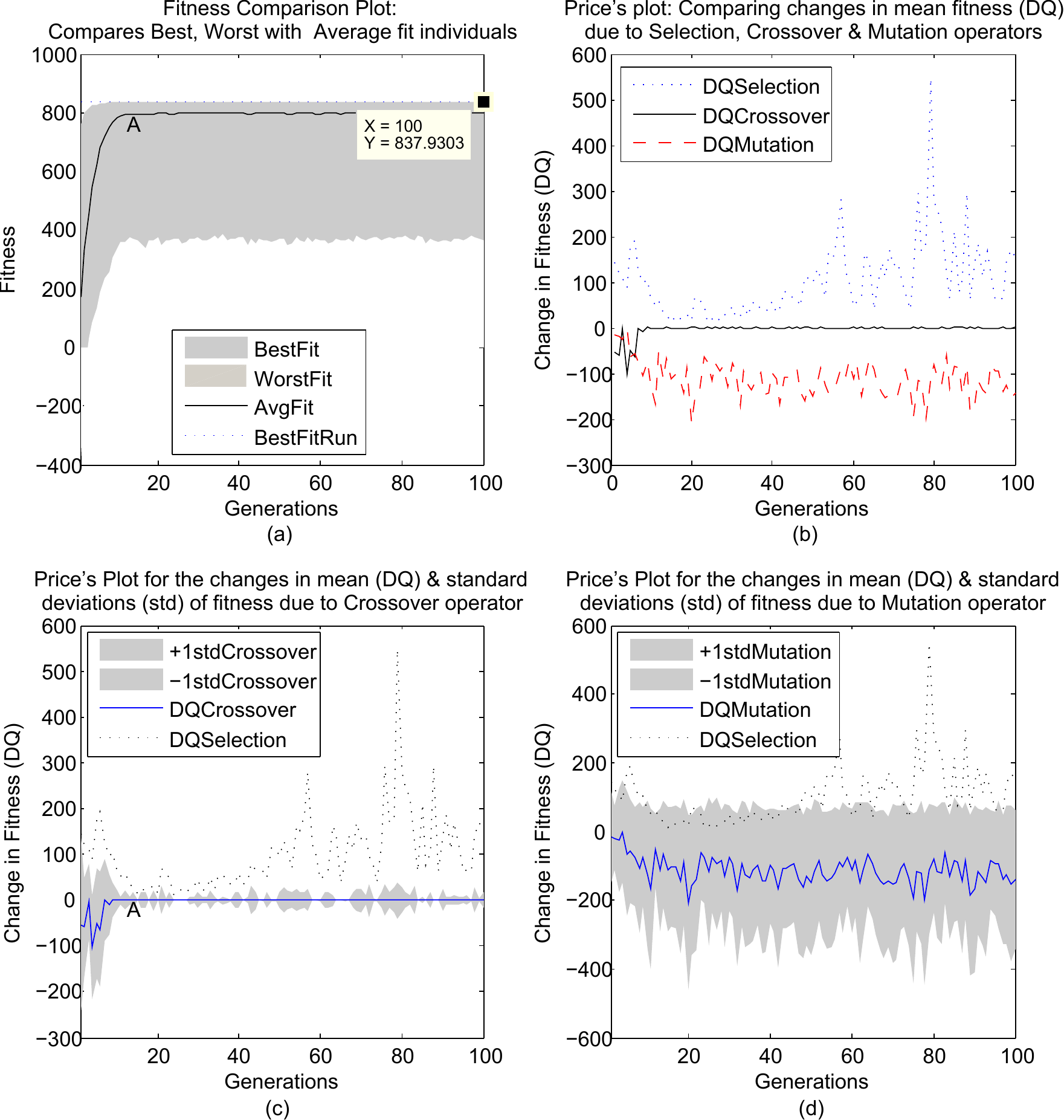}
	
	\caption[Convergence detection by monitoring the contribution  of crossover operator]{Convergence detection by monitoring the $\pm \sigma$ (standard deviation) interval of the contribution  of crossover operator with BTS: Plot (a) is a fitness comparison plot for the \textit{best}, \textit{worst} and \textit{average} fitness. Plot (b) compares the effect of genetic operators using extended Price's equation \eqref{ProposedExtPricesEqn}. Plots (c and d) show the \textit{average} and the $\pm \sigma$ (shaded areas) for the crossover and mutation operators respectively. For all plots $P_{c}=1.0, P_{m}=1/l$ and $N=100$ and results are averaged over $100$ runs. The label A on plot (c) marks the generation at which the evolution converges and it corresponds to label A on plot (a) where fitness progress stalls.}
	\label{ConvergenceDetectionWithTournamentSelFigure}
\end{figure}

\subsection{The proposed convergence threshold parameter}
\label{ProposedConvergenceThresholdSec}

Having used one standard deviation interval to analyse the effect of genetic operators, then, the width of the $\pm\sigma$ envelope for the effect of an operator $j$ (crossover in this case) on the fitness growth at every $k$th generation lies within the interval:
\begin{equation}
\left[\Delta Q_{jk}-\sigma_{jk}, \; \Delta Q_{jk}+\sigma_{jk}\right]
\end{equation}
where $\Delta Q_{jk}$ is the change in the \textit{average} fitness of the population at iteration $k$ due to operator $j$ and $\sigma_{jk}$ is the corresponding standard deviation. Let the width be represented by $\sigma q_{jk}$, then it can be determined as follows:

\begin{equation}
\label{ConvergenceThresholdParamEqn}
\sigma q_{jk} = \left(\Delta Q_{jk}+\sigma_{jk}\right)-\left(\Delta Q_{jk}-\sigma_{jk}\right)=2\sigma_{jk}.
\end{equation}

Consider the complete set of plots for experiment 1 (for BTS) shown on figure \ref{ConvergenceDetectionWithTournamentSelFigure}, the point labelled A on plot (a) directly corresponds to the point labelled A on plot (c) when read from the x-axis (i.e., the generations axis). It is fairly easy to notice that the more the width of the $\pm\sigma$ envelope for the crossover operator shrinks to zero (plot (c)), the flatter the gradient of the \textit{best} and \textit{average} fitness curves on plot (a). In other words, the width $\sigma q_{jk}$ in equation \eqref{ConvergenceThresholdParamEqn} indirectly represent the available diversity in the population. Hence, the generation at which $\sigma q_{jk}$ tends to zero signifies the beginning of convergence in the overall search process.

Therefore, the proposed convergence measure is to prescribe a threshold value for the parameter $\sigma q_{jk}$ such that whenever $\sigma q_{jk}$ falls below this threshold, the evolutionary search process is automatically terminated. We must remark here that this threshold parameter is user defined and its appropriate value is determined empirically, preliminary investigations reveal that a value of $\sigma q_{jk}\le 0.01$ is suitable for crossover and mutation probabilities of $P_{c}=1.0$ and $P_{m}=1/l$ respectively. Ultimately, since determining optimum value for this threshold parameter relies on some other EC parameters, as part of our further investigations, a thorough sensitivity analysis will be carried out.

\section{Contribution}
This chapter provides insight on the relevance of convergence detection in EC and sheds light on some conventional convergence measures. Most importantly, it provides insight on the interactions among the genetic operators during the evolutionary search process. A visual means for investigating the individual roles of genetic operators on fitness progress is developed using the extended Price's equation.

The experimental results obtained after comparing the plots for the contribution of the genetic operators using RWS and BTS reveal two interesting findings. 

First, while using the RWS method, the proposed technique for visualizing the effect of genetic operators using extended Price's equation have shown that all the three operators play random roles of exploiting and exploring the search space throughout the evolution period. This has made it infeasible to deploy their individual effect as a means to measure convergence when RWS is utilized. 

Second, substituting the RWS with a BTS method reveals the contrary. The extended Price's plots in this case demonstrate a clear separation of roles among the three genetic operators. The BTS operator takes the lead in exploiting the highly fit areas of the search space; the mutation operator handles the exploration aspect while the crossover operator serves as a moderator between the two. This has made it possible to monitor and use the crossover's effect as an effective means of detecting convergence in the evolution. 

Consequently, the insight gathered above led to the development of a novel convergence measure (section \ref{AutoConvergDetcSec}) that can allow automatic convergence detection in EC. 

\section{Remarks}
While the earlier chapters have introduced and analysed the initialization and developmental aspects of evolutionary computation algorithms, this chapter augments the previous work by analysing the process that could lead to successful termination of the evolutionary search, which is a prerequisite to achieving one of the objectives for the proposed hybrid optimization algorithm. Therefore, in the following chapter an investigation of the principles of gradient based methods for local optimization will be presented. This will then be utilized in the later chapters for the proposed hybrid optimization algorithm.

\singlespacing
\bibliographystyle{ieeetr}




\chapter[Local Search Algorithms for Optimization]{Local Search Algorithms for Optimization}
\label{LocalSearchAlgoChap}

In the previous chapters, detailed investigation and analysis on evolutionary computations was given. Such algorithms are global search \textit{approximate} techniques where success relies upon some stochastic heuristics. The design of the proposed hybrid algorithm requires combining features form both the \textit{approximate} and \textit{exact} methods. Thus, before plunging into the hybrid algorithms, this chapter will present the general idea behind the state-of-the-art gradient-based local optimization algorithms. Details on the design of the Newton-based local optimization techniques (i.e. interior point method (IPM) and the sequential quadratic programming (SQP) algorithm) that will be used for the hybridization will be given. An algorithmic approach for effective evaluation of derivatives will then be presented.

\section{An Overview of Local Optimization Algorithms}

Consider a general expression of a continuous optimization problem shown in equation \eqref{NonlinearObjFuncEqn}, the objective function to be optimized\footnote{Minimization or Maximization.} $f:\mathbb{R}^{n}\rightarrow \mathbb{R}$ is defined in terms of a vector of the design variables $x$ of length $n$ that is in the set of real numbers $\mathbb{R}$.
\begin{equation}
\label{NonlinearObjFuncEqn}
\min f(x): f\in \mathbb{R}^{n}; n\geq 1
\end{equation}

Commonly, gradient based optimization algorithms sequentially generate at every $k^{\textrm{th}}$ iteration, a vector of solution points $x_{k}$ that is expected to terminate at $x^{*}$ (i.e the optimal solution) when either no more progress can be made or when the optimal solution has been attained with sufficient accuracy. Beginning with an arbitrarily chosen initial solution point $x_{0}$, the iterative progression from point $x_{k}$ to $x_{k+1}$ depends on the ability of the algorithm to decide on the nature of the gradient of the objective function $f$ at any point $x_{k}$, and possibly, some additional information about the previous points $x_{k-1},x_{k-2},...,x_{0}$.

Typically, every new solution point $x_{k+1}$ is expected to yield a lower function value than its predecessor $x_{k}$. In fact, a critical distinction between the various local optimization algorithms is on the nature of their successive iterations. A class of algorithms that insists on reduction in the function value at every iteration enforces $f(x_{k}) < f(x_{k-1})$  and constitutes the so-called \textit{greedy} algorithms \cite{Bhatti2000Book}. The Newton and quasi-Newton algorithms are part of this class. The other class that do not insist on minimizing the value of the objective function at the end of \textit{every} iteration usually enforces $f(x_{k})<f(x_{k-m});$ where $m>1$ is the maximum acceptable iterations without a decrease in the objective function value. This is the class of \textit{nonmonotone} algorithms which are non-greedy descent algorithms.

Another major categorization of the gradient based algorithms is on their approach for stepping from one iteration point $x_{k}$ to the next $x_{k+1}$. The approach for taking a step is always either a \textit{line search} or a \textit{trust region} based. The general expression for deriving the next iteration point $x_{k+1}$ is dependent on two key parameters; the evaluated search direction $d_{k}\in \mathbb{R}^{n}$ and an estimated value for the step length parameter $\alpha_{k}:\left\{\alpha_{k}\in \mathbb{R}:0<\alpha_{k}\leq 1\right\}$ along the obtained direction such that:
\begin{equation}
\label{NextIterPointEqn}
x_{k+1}=x_{k}+\alpha_{k}d_{k}.
\end{equation}

The line search based algorithms evaluate the search direction and then decide on how long to search along that direction by estimating a suitable value for the step length parameter $\alpha_{k}$. On the contrary, trust region based algorithms start by defining a region around the current iteration point $x_{k}$ within which they believe a model derived from the second order Taylor approximation of the objective function \eqref{2ndOrderTaylorExpEqn} is a good approximation of the actual objective function. Based on the size of the defined trust region, these algorithms choose the search direction and the step length simultaneously.

One might have noticed that for the line search based methods, if the step length parameter does not lead to a decrease in the value of the objective function, the algorithm can easily try to re-evaluate a feasible one. However, in trust region methods, both the search direction and the step length must be discarded, the size of the trust region must be contracted and the procedure repeated. This and many other reasons made the line search based optimization algorithms computationally cheaper than their trust region counterparts. Nevertheless, we must remark here that some researchers \cite{NocedalWright2006Book} have the view that trust region methods can be more reliable compared to the line search based methods especially when the initial starting point is significantly away from the actual minimizer. Yet, since the local search algorithm (SQP) to be proposed in the later sections will be a line search based method, this investigation will be limited to only line search based local optimization techniques.

\section{Line Search Based Local Optimization Methods}

As highlighted earlier, the iterative progress in line search based methods relied on the computed search direction $d_{k}$ and the evaluation of the possible distance along the direction to which the search can progress, i.e. step length $\alpha_{k}$. The manner in which the search directions are computed is where the gradient based methods principally differ. 

For any smooth continuously differentiable function $f:\mathbb{R}^{n}\rightarrow \mathbb{R}$ in the neighbourhood of iteration point $x$, and an assumed vector of search direction $d\in \mathbb{R}^{n}$, the necessary and sufficient conditions for optimality requires $\nabla f(x^{*})=0$, and $\nabla^{2} f(x^{*})\in \mathbb{R}^{n\times n}$ to be a symmetric positive definite matrix for $x^{*}$ to be a local minimizer of the function $f$. The Taylor expansion of $f$ yields:
\begin{equation}
\label{TaylorExpEqn}
f(x+d)=f(x)+\nabla f(x)^{T}d+\frac{1}{2}d^{T}\nabla^{2}f(x)^{T}d+ \cdots
\end{equation}

Gradient based methods assume that the objective function $f$ is differentiable and it is approximately quadratic (i.e. convex) in the vicinity of the stationary point $x$. Therefore, the second-order Taylor expansion of equation \eqref{TaylorExpEqn} is:
\begin{equation}
\label{2ndOrderTaylorExpEqn}
f(x+d)\approx f(x)+\nabla f(x)^{T}d.
\end{equation}
Thus, setting the gradient of equation \eqref{2ndOrderTaylorExpEqn} to zero at stationary point and solving for $d$ yields:
\begin{equation}
\label{SearchDirectionEqn}
d=-\frac{\nabla f(x)}{\nabla^{2}f(x)^{T}} = -\left[H(x_{k})\right]^{-1}\nabla f(x)
\end{equation}
where $H(x)=\nabla^{2}f(x)$ is the Hessian of the function $f$ and $d$ is the search direction which is required to be a descent direction.

\section{Search Directions and Step Length in Gradient Based Algorithms}

As described previously, any step of the gradient based algorithms involves evaluation of the search direction $d_{k}$ and the step length parameter $\alpha_{k}$ such that the next iteration is defined as in equation \eqref{NextIterPointEqn} above. In the following, the nature of the search directions and step length parameters for various gradient based algorithms will be presented.

\subsection{Methods of Evaluating Descent Search Directions}

Commonly, gradient based algorithms require the search direction $d_{k}$ at any given iteration $k$ to be a descent direction such that the directional derivative:
\begin{equation}
\label{DirectionalDerivativeEqn}
d_{k}^{T}\nabla f(x)<0
\end{equation}
will guarantee a reduction in the value of the objective function $f$ along this direction. From equation \eqref{SearchDirectionEqn}, a general form for the search directions $d_{k}$ is:
\begin{equation}
\label{GeneralizedSearchDirectionEqn}
d_{k}=-B_{k}^{-1}\nabla f(x),
\end{equation}
where $B_{k}\in \mathbb{R}^{n\times n}$ is a symmetric non-singular matrix. The most commonly used types of search directions for local optimization algorithms are as follows:

\begin{enumerate}[i.]
	\item \textbf{Steepest Descent Direction:} The search direction $d_{k}=-\nabla f(x)$ is called the \textit{steepest descent direction} and among all the directions via which the search could move from point $x_{k}$ to $x_{k+1}$, this is the direction along which $f$ decreases most rapidly. Therefore, \textit{steepest descent algorithms} are line search methods that move along the steepest direction at every iteration. Notice that the Hessian matrix in equation \eqref{GeneralizedSearchDirectionEqn} is set to an identity matrix (i.e., $B_{k}=I$) for this category of algorithms. Thus, the primary advantage of these algorithms is that they require only the computation of the gradient of the objective function, and hence, they have low computational cost per iteration.
	\item \textbf{Newton Direction:} Another important search direction is the \textit{Newton direction}.  Derived from the second-order Taylor expansion of $f$, the value of $B_{k}=\nabla^{2} f(x)$ is the true Hessian of the objective function. The Newton direction is quite a reliable descent direction if the Hessian $\nabla^{2} f(x)$ is sufficiently smooth so that the quadratic approximation of the objective function in \eqref{2ndOrderTaylorExpEqn} is sufficiently accurate. Thus, \textit{Newton methods} are the algorithms that use Newton directions at every iteration with the condition that, the Hessian $\nabla^{2} f(x)$ is symmetric and positive definite\footnote{A positive definite matrix must have a positive determinant, i.e., it is always non-singular. Thus, a real and symmetric matrix is positive definite iff all its eigenvalues are positive. It is positive semidefinite, negative semidefinite or negative definite iff all of its eigenvalues are non-negative, non-positive or negative respectively.}. These methods typically have fastest convergence rate compared to all other line search based local optimization techniques. Moreover, when the search point $x_{k}$ is within the neighbourhood of the minimizer, quite a few steps/iterations are required to converge to the solution point with high accuracy.\\
The main drawback of Newton methods is the need for the evaluation of the true Hessian $\nabla^{2} f(x)$ which is a matrix of second derivatives and can be quite cumbersome, error prone and expensive especially when the dimension of $x$ is large. Furthermore, whenever $\nabla^{2} f(x)$ is not positive definite, the Newton direction may not be defined since the inverse of $\nabla^{2} f(x)$ may be ill-conditioned and consequently yields a search direction $d_{k}$ that violates the descent property in equation \eqref{DirectionalDerivativeEqn}.
	\item \textbf{Quasi-Newton Direction:} Yet another search direction is the \textit{quasi-Newton direction}. In this case, instead of evaluating the true Hessian $\nabla^{2} f(x_{k})$, the value of $B_{k}$ \eqref{GeneralizedSearchDirectionEqn} is only a mere approximation of it. The initial approximation of the Hessian $(B_{k})$ is usually an identity matrix $I$, which is then updated iteratively to take into account the additional information derived from subsequent iterations. In essence, the Hessian update methods mainly rely on the fact that delta changes in the gradient of the objective function $f$ from one iteration point to another provide vital information about the nature of its second derivative along the search direction. The algorithms that rely on this principle of Hessian update are called \textit{quasi-Newton algorithms} and can achieve high rate of convergence (superlinear) without the expensive explicit evaluation of the Hessian matrix.\\
A common drawback of quasi-Newton methods is that their rate of convergence is slower than that of Newton methods as they require running through several iterations. Also, after certain number of steps, the Hessian approximation often tends to yield an ill-conditioned matrix that may cause the entire search process to diverge. This necessitates having the approximate Hessian $B_{k}$ been reset to its initial value which consequently slows progress even further. Because for the majority of the popular Hessian update techniques such as BFGS, DFP and SR1 \cite{Powell1986, Bhatti2000Book}, the initial approximation of $B_{k}$ is an identity matrix $I$, the entire search process is always reduced to a simple steepest descent after every reset of the Hessian matrix.
	\item \textbf{Conjugate Gradient Direction:} This category of search directions are typically more effective than steepest descent directions and are fairly computationally equivalent. Unlike in the Newton and quasi-Newton directions, this search direction requires no storage of large Hessian matrices. The methods that use this search direction are called \textit{conjugate gradient methods}. These methods however are only first order and therefore do not achieve fast convergence rate as Newton and quasi-Newton methods. As argued by Powell \cite{PowellGriffiths1984ref207, Powell1977ref203}, conjugate gradient methods may fail to converge on non-convex problems and therefore need restart.
\end{enumerate}

\subsection{Methods of Evaluating the Step Length Parameter}
\label{EvaluatingStepLengthParameterSec}

The step length parameter $\alpha \in \mathbb{R}:(0<\alpha \leq 1)$ is a positive scalar that is expected to give a substantial reduction in the function value. It is normally the minimizer of the \textit{merit function}\footnote{Originally used in regression, a merit function is a function that measures the agreement between data and the fitting model for a particular choice of the parameters \cite{Weisstein2011}. Parameters are adjusted based on the value of the merit function until a smallest value is obtained, the resulting parameters are known as the best-fit parameters.} defined by:
\begin{equation}
\label{MeritFuncEqn}
\varphi (\alpha)=f(x_{k}+\alpha_{k}d_{k}); \alpha >0.
\end{equation}

Generally, the ideal value for the step length parameter $\alpha$ is the global minimizer of the merit function \eqref{MeritFuncEqn} itself. Determination of this global minimizer requires several evaluations of the objective function $f$ and its gradient $\nabla f$. Therefore, since evaluation of even a local minimizer of equation \eqref{MeritFuncEqn} is expensive, a tradeoff is necessary. Thus, at every $k$th iteration, inexact line search algorithms are used to try out a sequence of candidate values for $\alpha_{k}$. Based on some pre-defined termination conditions, a suitable value for $\alpha_{k}$ is accepted. A simple condition that ensures $\alpha_{k}$ provides a meaningful reduction in $f$ entails:
\begin{equation}
f(x_{k}+\alpha_{k}d_{k})<f(x_{k}).
\end{equation}
Worth noting is that the step length $\alpha$ need not to lie near the minimizer of the merit function $\varphi (\alpha)$ for it to effectively yield a sufficient reduction in the objective function.

Several inexact line search algorithms for estimating a suitable value for the step length parameter exist, and they are usually named after the termination condition used. Thus, they include the method based on Armijo’s rule, Goldstein condition and the popular Wolf conditions \cite{Bhatti2000Book}. All these algorithms typically possess two key stages. The first stage is a backtracking procedure that finds an interval containing the desirable step lengths, while the second is an interpolation phase where a good value for $\alpha$ is computed within the obtained interval.

The most commonly used termination condition is based on the two Wolf conditions and it has been analytically proven that \cite{BAALI1985ref3}, to every smooth continuous function there exist a value for the step length parameter that satisfies the following two Wolf conditions.
\begin{description}
	\item[First:] The step length $\alpha$ must provide sufficient decrease in $f$ such that:
\begin{equation}
\label{Wolfconditions1Eqn}
f(x_{k}+\alpha_{k}d_{k})\leq f(x_{k})+c_{1}\alpha_{k}\nabla f_{k}^{T}d_{k},
\end{equation}
where $c_{1}\in [0,1]$ is a positive constant which is quite small in practice, typically $c_{1}=10^{-4}$. Thus, the reduction in $f$ should be proportional to both the value of $\alpha_{k}$ and the directional derivative $\nabla f_{k}^{T}d_{k}$.

	\item[Second:] This is called the curvature condition, it ensures that the derivative of the merit function $\varphi (\alpha_{k})$ (see equation \eqref{MeritFuncEqn}) is greater than a constant $c_{2}$ times the derivative of $\varphi (0)$, i.e.:
\begin{equation}
\label{Wolfconditions2Eqn}
\varphi'(\alpha_{k})\geq c_{2}\varphi'(0)
\end{equation}
such that the new directional derivative satisfies: 
\begin{equation}
\nabla f(x_{k}+\alpha_{k}d_{k})^{T}d_{k}\geq c_{2}\nabla f_{k}^{T}d_{k}
\end{equation}
where $c_{2}\in [c_{1},1]$, its typical value for Newton and quasi-Newton methods is $0.9$ and $0.1$ for nonlinear conjugate gradient methods. For more details on these and other techniques for evaluating the step length parameter, see \cite{BAALI1985ref3, HassanXiming2011IJCISIM, Fletcher1987Book, Hertog1994, Betts2001}.
\end{description}

\section{Convergence Analysis of Gradient Based Algorithms}

Convergence assessment is critical to evaluating the performance of any optimization algorithm. In this section, a brief investigation on the \textit{rate of convergence} and \textit{global convergence} of the gradient based algorithms will be presented and the local search algorithms will be compared on this basis. 

While a number of measures exist for evaluating the rate of convergence of gradient based algorithms, the commonly used one \cite{Bhatti2000Book} is the $Q-$convergence ($Q$ stands for quotient) measure defined in terms of the quotient of successive errors. For a sequence of solution points \mbox{$\{x_{k}\}:x\in \mathbb{R}$} that iteratively converges to a minimizer $x^{*}$, the rate of convergence of gradient based algorithms is classified as follows.

\begin{description}
	\item[$Q-$Linear:] The convergence is $Q-$linear if there exist a constant $r\in [0,1]$ such that:
\begin{equation}
\frac{\vert\vert x_{k+1}-x^{*} \vert\vert}{\vert\vert x_{k}-x^{*} \vert\vert}\leq r; \;\forall\: k \; \textrm{sufficiently large.}
\end{equation}
	\item[$Q-$Superlinear:] An algorithm is said to converge $Q-$superlinearly if as the number of iterations $k$ tends to infinity, the error between two successive iteration points dies out.
\begin{equation}
\lim_{k\to \infty}\frac{\vert\vert x_{k+1}-x^{*} \vert\vert}{\vert\vert x_{k}-x^{*} \vert\vert}\rightarrow 0
\end{equation}
	\item[$Q-$Quadratic:] These are algorithms which for any scalar $M\in \mathbb{R}:M>0$, their rate of convergence satisfies:
\begin{equation}
\frac{\vert\vert x_{k+1}-x^{*} \vert\vert}{\vert\vert x_{k}-x^{*} \vert\vert^{2}}\leq M; \; \forall\: k \; \textrm{sufficiently large.}
\end{equation}
\end{description}
It is easy to notice that higher order rate of convergence  are also possible and the trend is that a $Q-$quadratic algorithm will always converge faster than a $Q-$superlinear or $Q-$linear algorithm. While all Newton algorithms converge $Q-$quadratically, quasi-Newton methods converge $Q-$superlinearly and at the other extreme, all steepest descent algorithms have $Q-$linear rate of convergence.

Conversely, the global convergence of gradient based algorithms requires not only a suitable estimate of the step length parameter, but also a carefully chosen search direction. In order to have good convergence characteristics, the evaluated search direction must not be orthogonal to the gradient $\nabla f$, i.e. at least steepest descent steps must be taken regularly.

From the foregoing, we would like to remark that one undesirable behaviour associated with general gradient based algorithms is that while on one hand steepest descent algorithms have guaranteed global convergence, their rate of convergence is quite slow (i.e., converge only Q-linearly). On the other hand, Newton algorithms can converge most rapidly (i.e., Q-quadratically) to a minimizer, but global convergence is not guaranteed when the initial search point is not in the vicinity of the minimizer. This is because, when the search point is away from the minimizer, Newton methods tend to produce search directions that cannot lead to any decrease in $f$ as they are nearly orthogonal to the gradient of the objective function $\nabla f$. Thus, so long as the starting point is not guaranteed to be near the minimizer, a \textit{tradeoff} is necessary to ensure global convergence and at the same time achieve rapid rate of convergence with gradient based algorithms.

A summary of the key features for the aforementioned major categories of gradient based algorithms is given in table \ref{ComparisonGradAlgoTab}. It provides a simple comparison of these methods from which it can be deduced that Newton based optimization algorithm is the most suitable choice for the proposed hybrid algorithm. This is because the initial starting point that will be fed to the local algorithm will almost always lie in the vicinity of a minimizer since it is derived after sufficient convergence of the evolutionary algorithm. Therefore, global convergence of the Newton based algorithm and ultimately that of the proposed SQP algorithm is assured.
\begin{table}[htb]
\begin{center}
\caption[Comparison of the three major gradient based algorithms]{Comparison of the three major gradient based algorithms: The nature of their search directions, step length parameter and rate of convergence \qquad \qquad \qquad \qquad \qquad \qquad}
\label{ComparisonGradAlgoTab}
\begin{tabular}{p{2.0cm}p{2.1cm}p{2.4cm}p{1.9cm}p{1.8cm}p{2.0cm}}
\toprule
Algorithms & Search \mbox{Direction} & Hessian \mbox{Requirement} & Step Length Parameter & Degree of \mbox{Computation} & Rate of \mbox{Convergence}\\
\midrule

Steepest \mbox{Descent} &  \multirow{6}{*}{$-B_{k}^{-1}\nabla f(x_{k})$} & Not needed: \mbox{$B_{k}=I$}  &  \multirow{2}{*}{$0<\alpha<1$} & \multirow{2}{*}{1st Order} & \multirow{2}{*}{$Q-$Linear}\\ \cline{3-6}

\multirow{2}{*}{Quasi-Newton} &  & Approximation: $B_{k}\approx \nabla^{2}f(x_{k})$  &  \multirow{2}{*}{$0<\alpha<1$} & \multirow{2}{*}{2nd Order} &  \multirow{2}{*}{$Q-$Superlinear}\\ \cline{3-6}
                 
Newton method & & Exact Hessian: $B_{k}=\nabla^{2}f(x_{k})$  &  \multirow{2}{*}{$0<\alpha\leq 1$}   &  \multirow{2}{*}{2nd Order} & \multirow{2}{*}{$Q-$Quadratic}\\                  
\bottomrule
\end{tabular}
\end{center}
\end{table}

\section{Sequential Quadratic Programming Algorithm}

The standard sequential quadratic programming (SQP) algorithm is a Newton based nonlinear optimization algorithm that can be implemented either in the line search or trust region framework and can successfully handle constrained nonlinear problems.  One major feature that is common to all SQP formulations is that the algorithms are divided into an outer \textit{linearization} and an inner \textit{optimization} loop. The linearization loop is responsible for approximating the nonlinear objective function $f(x)$ in the original optimization problem \eqref{NonlinearObjFuncEqn} with a quadratic model, and if it exist, the nonlinear constraints with their approximate linear expressions at the current point $x_{k}$. The result of this linearization allows representation of the original nonlinear problem \eqref{NonlinearObjFuncEqn} as a sequence of quadratic programming (QP) subproblems \eqref{QPProblemEqn} which can then be solved using any QP solver.
\begin{equation}
\label{QPProblemEqn}
\underset{x}{\min}\:f(x)^{T}x+\frac{1}{2}x^{T}Hx;\;H\in \mathbb{R}^{n\times n}
\end{equation}

The framework of SQP is essentially based on Newton method, and since Newton methods solve nonlinear problems via a sequence of Newton steps, this method is called sequential quadratic programming.  Extensive treatment on SQP algorithm can be found in the books of Fletcher and Nocedal \cite{Fletcher1987Book, NocedalWright2006Book}.

While the standard SQP algorithm uses \textit{active set} strategy that is based on the null-space and range-space methods to solve its QP subproblems, the method adopted here is based on the earlier work in \cite{HassanXiming2011IJCISIM, XimingHassanAppyingIPMCSSE} where interior point method (see section \ref{IPMSec}) will be deployed to minimize the QP problem.

\section[Interior Point Method for Solving Quadratic Problems]{Interior Point Method for Solving Quadratic\\ Problems}
\label{IPMSec}

Interior point methods (IPM) have their origin from linear programming (LP) and have become important tools in mathematical programming, operations research and in many other areas of science. The main idea in IPM algorithms is to approach the optimal solution of the LP problem through the interior of the feasible region. This is the opposite strategy of the well known simplex algorithm proposed by Dantzig \cite{Dantzig1961, Dantzig1963} which moves along the boundary of the feasible region. IPM approach was also proposed by Hoppe \cite{HoppePetrova2002}. 

Over the last two decades, a lot of research works have been reported on IPM algorithm and its variants many of which are surveyed and referenced by Mizuno et al. \cite{MizunoKojima1995} and Wright \cite{Wright1997}. Interior point methods can also be extended to handle general convex problems such as quadratic programming problems. Details can be found in the book of Nesterov and Nemirovskii \cite{NesterovNemirovski1994} and in Hertog \cite{Hertog1994}. 

In this work, IPM algorithm will be employed to minimize the QP subproblems produced as a result of linearization of the original nonlinear optimization problem by the main SQP algorithm. Thus, IPM will form the inner (optimization) loop of the proposed SQP algorithm. 
\\
Let a general quadratic programming problem be defined as:
\begin{equation}
\label{GeneralizedQPProblemEqn}
\underset{x}{\min}\:q(x)=c^{T}x+\frac{1}{2}x^{T}Qx;\;x\in \mathbb{R}^{n}, \; c\in \mathbb{R}^{n}, \; Q\in \mathbb{R}^{n\times n}
\end{equation}
then, the function $q(x)$ \eqref{GeneralizedQPProblemEqn} is convex if at least the matrix of quadratic terms $Q$ is symmetric positive semidefinite.
\\
If the function to be optimized is subject to some constraints $g$, say
\begin{equation}
\label{NonlinearConstraintsEqn}
g_{i}(x)\geq 0: x\in \mathbb{R}^{n}, \; i=1,2,...,m, \; \nabla g \neq 0
\end{equation}
then, IPM algorithms use the method of Lagrange multipliers to redefine the optimization problem into a composite function $\mathcal{L}$ called Lagrange function (named after its inventor Joseph Louis Lagrange \cite{ArfkenWeber2005}). The Lagrange function combines the objective function \eqref{GeneralizedQPProblemEqn} and all the constraints \eqref{NonlinearConstraintsEqn} such that:
\begin{equation}
\label{LagrangeEqn}
\mathcal{L}(x,\lambda)=c^{T}x+\frac{1}{2}x^{T}Qx+\sum_{i=1}^{m}\lambda^{T}g_{i}(x)
\end{equation}
where $\lambda\geq 0$ is a vector of Lagrange multipliers. 

At every iteration of the SQP algorithm, the matrix of quadratic terms $Q$ is derived from the Hessian of the original nonlinear problem. $Q$ is then used to construct the QP subproblem \eqref{GeneralizedQPProblemEqn} to be solved in the inner loop of the SQP algorithm. The minimum of \eqref{LagrangeEqn} can then be obtained by taking its gradient with respect to all its variables (i.e., $\nabla_{x,\lambda}\mathcal{L}(x,\lambda)$) and solving for the unknowns. This is done without explicitly inverting $g$, which according to \cite{ArfkenWeber2005} is the reason why the method of Lagrange multipliers can be quite handy. Hence, minimizing the Lagrange function yields the minimum of the originally constrained function. Details on constrained minimization can be found in \cite{Bhatti2000Book, Fletcher1987Book}. 

Of particular interest at this juncture is the fact that standard implementations of SQP algorithm only utilize approximate Hessians for the nonlinear problems. The Hessian is initially approximated by an identity matrix $I$ and then updated iteratively via the well known Hessian update procedure proposed by Broyden-Fletcher-Goldfarb-Shanno (BFGS) \cite{NocedalWright2006Book, Powell1986, Betts2001}. 

The proposal here is to develop an automatic differentiation package that can algorithmically determine the \textit{exact} values for the gradient and Hessian of any differentiable function at any given search point $x_{k}$. The cost of evaluating the derivatives will be more or less equivalent to that of evaluating the function values themselves. Ultimately, the computational burden of evaluating derivatives via the conventional symbolic or finite difference methods will be alleviated. This is important for the proposed SQP algorithm in two key ways: 
\begin{enumerate}[i.]
\item Having exact Hessians at its disposal, the proposed SQP algorithm has upgraded the standard SQP algorithm from a mere quasi-Newton algorithm that takes steepest descent steps to a Newton algorithm taking full\footnote{Full step means having the value of the step length parameter $\alpha=1$. Quasi-Newton methods barely accept $\alpha=1$ and even when they did, they still run at a superlinear rate \cite{Powell1986}.} Newton steps at every iteration. Hence, the convergence rate is enhanced from superlinear to quadratic. 
\item Since exact Hessian matrices will be used, the \textit{frequent} reset of the search direction due to ill-conditioned approximate Hessians could be avoided or at least minimized\footnote{Resetting the Hessian to an identity matrix $I$ is common in BFGS and other update procedures. If exact Hessians are available and the starting point is in the neighbourhood of the minimizer, then the problem of resetting search direction is eliminated.}. This means both global convergence and quadratic rate of convergence are assured. Therefore, the proposed method is hoped to converge to the minimizer in remarkably few steps.
\end{enumerate}
Algorithm \ref{SQPipmAlgo} shows a detailed framework for the proposed SQP local optimization algorithm. The while loop at line $5$ exits only if one of the three stopping conditions is satisfied. The first two are tolerance parameters $\textrm{Tol}_{1}$ and $\textrm{Tol}_{2}$ which are small positive numbers used to check whether the delta changes in the gradient and directional derivative of the problem have sufficiently approach zero. Their typical value is $10^{-6}$ to $10^{-3}$. The third parameter MaxIter is the maximum iteration limit which is also user defined. Finally, instead of using approximate Hessians (see line $7$), an exact value of the matrix $\nabla^{2}f(x_{k})$ will be evaluated with the aid of an automatic differentiation algorithm to be presented in section \ref{ADAlgorithmSec}.

\begin{algorithm}[htp]
\caption{The SQP local optimization algorithm}
\label{SQPipmAlgo}
\begin{algorithmic}[1]
\STATE $\textbf{begin}$\\
\STATE	$\qquad$ $k\leftarrow0;$\\
\STATE	$\qquad$ $x_{k}\leftarrow x_{0};$ \qquad \qquad \% $x_{0}$ is the starting point returned by the EC algorithm\\
	
\STATE	$\qquad$ $d_{k}\leftarrow d_{0};$ \qquad \qquad \%  Initial search direction $d_{0}$ is a vector of all ones\\

\STATE	$\qquad$ $\textbf{while}$ $\nabla f(x_{x})> \textrm{Tol}_{1}$ \; $\textbf{and}$ \; $\vert\vert d_{k}\vert\vert > \textrm{Tol}_{2}$ \; $\textbf{and}$ \; $k<\textrm{MaxIter}$ \; $\textbf{do}$\\
\STATE	$\qquad \qquad$  Linearize \eqref{NonlinearObjFuncEqn} into a QP subproblem \eqref{GeneralizedQPProblemEqn}\\
\STATE	$\qquad \qquad$  $H(x_{k})\leftarrow \nabla^{2}f(x_{k})$ \qquad \qquad \% computed via Automatic Differentiation\\
\STATE	$\qquad \qquad$  Evaluate $d_{k}$ by minimizing QP subproblem \eqref{GeneralizedQPProblemEqn} \; \% IPM algorithm will be used\\
	
\STATE	$\qquad \qquad$  $\textbf{if}$ \;$\alpha=1$ satisfies Wolf conditions \eqref{Wolfconditions1Eqn} and \eqref{Wolfconditions2Eqn} \; $\textbf{then}$\\
\STATE		$\qquad \qquad \qquad$ $\alpha_{k}\leftarrow 1$\\
\STATE	$\qquad \qquad$ $\textbf{else}$\\
\STATE		$\qquad \qquad \qquad$ Evaluate $\alpha_{k}:\alpha_{k}>0$ that satisfies \eqref{Wolfconditions1Eqn} and \eqref{Wolfconditions2Eqn}\\
\STATE	$\qquad \qquad$ $\textbf{end if}$\\
	
\STATE	$\qquad \qquad$ $x_{k+1} \leftarrow x_{k}+\alpha_{k}d_{k};$\\
\STATE	$\qquad \qquad$ $k \leftarrow k+1;$\\
\STATE	$\qquad$ $\textbf{end while}$\\
\STATE $\textbf{end}$\\	
\end{algorithmic}
\end{algorithm}

In the following, details of the exact Hessian evaluation using automatic differentiation principle will be given.

\section[Automatic Differentiation for Exact Derivatives Evaluation]{Automatic Differentiation for Exact Derivatives\\ Evaluation}
\label{ADAlgorithmSec}

Also called algorithmic differentiation, \textit{automatic differentiation} (AD) is built on the notion that the value of any function, simple or complex, is evaluated via a sequence of basic elementary operations involving unary or binary operators and their operands. The key motivation behind the need for AD follows the fact that conventional methods like the method of divided differences are prone to round-off errors when the differencing interval is small, and suffers from truncation errors when the interval is large. On the other hand, the computational requirements of symbolic method are typically high especially when the function of interest is of higher dimensions.

Unlike the finite differencing method which has its origin from Taylor’s theorem, AD is based on repeated application of the \textit{chain rule} from elementary calculus. Recall that according to chain rule, if $f:\mathbb{R}^{n}\rightarrow \mathbb{R}$ is a composite function defined in terms of a vector $h\in\mathbb{R}^{m}$ which is in turn a function of $x\in\mathbb{R}^{n}$, then, the derivative of $f$ with respect to $x$ is:
\begin{equation}
\label{ChainRuleEqn}
\nabla_{x}f\left(h(x)\right)=\sum_{i=1}^{m}\frac{\partial f}{\partial h_{i}}\nabla h_{i}(x).
\end{equation}
Hence, AD techniques apply chain rule on a computational representation (tree-like graphs) of a function to generate analytic values for the function and its derivatives.

Historically, the basic ideas of AD have been around for long time \cite{BuckerCorliss2006BookAD}. However, it was the early extensive study by Griewank\footnote{The analysis in this section is adopted from their work.} et al. \cite{GriewankBook2008} that revived the interest in the use of algorithmic methods for evaluation of derivatives. Thereafter, a number of researches have been published\footnote{More about AD can be found in this portal: www.autodiff.org} on AD principles and its applications in mathematics and machine learning by Christianson and colleagues \cite{Christianson1992paper19, Christianson1992paper21, Christianson1999paper22} and recently in \cite{Bartholomew2000paper18, Ghate2007paper23, Bischof2002paper27, Forth2006}.

\subsection[Graphical Representation and Algorithmic Evaluation of Functions]{Graphical Representation and Algorithmic Evaluation of \\Functions}

Consider the two dimensional function shown in equation \eqref{SampleFuncForADEqn}, evaluating the function values at any given point $x_{k}=\left[x_{1},x_{2}\right]$ will entail an orderly execution of the elementary operations that made up the function. 
\begin{equation}
\label{SampleFuncForADEqn}
f(x_{1},x_{2})=\left(x_{1}x_{2}+\sin x_{1} +4\right)\left(3x_{2}^{2}-6\right)
\end{equation}
The elementary operations will lead to defining some intermediate variables that link the input variables to the output. Note that all the input, intermediate and the output variables can be represented as vertices of a tree that graphically represents the entire function. The list in table \ref{ListTraceforADSampleProblemTab} gives the detail trace for the evaluation of all the variables from the input to the output side.

\begin{table}[htb]
\begin{center}
	\caption[A list of variables definitions for a 2-dimensional sample problem]{A list of variables definitions for function: $f(x_{1},x_{2})=\left(x_{1}x_{2}+\sin x_{1} +4\right)\left(3x_{2}^{2}-6\right)$}
	\label{ListTraceforADSampleProblemTab}
\begin{tabular}{p{2.7cm}p{2.5cm}p{2.7cm}}
\toprule
\textbf{Variables} & \textbf{Vertices} & \textbf{Values}\\
\midrule
\multirow{5}{*}{Input} & $v_{-4}$  &  $4$\\
					   & $v_{-3}$  &  $3$\\
 				       & $v_{-2}$  &  $6$\\
 				       & $v_{-1}$  &  $x_{1}$\\
 				       & $v_{0}$  &  $x_{2}$\\
\cline{2-3}
\multirow{7}{*}{Intermediate} & $v_{1}$  &  $v_{-1}v_{0}$\\
                 & $v_{2}$  &  $\sin v_{-1}$\\
                 & $v_{3}$  &  $v_{1}+v_{2}$\\
                 & $v_{4}$  &  $v_{3}+v_{-4}$\\
                 & $v_{5}$  &  $v_{0}^{2}$\\
                 & $v_{6}$  &  $v_{5}v_{-3}$\\
                 & $v_{7}$  &  $v_{6}-v_{-2}$\\
                 
\cline{2-3} 
\multirow{2}{*}{Output} & $v_{8}$  & $v_{4}v_{7}$\\
   					    & $v_{8}$  &  $f$\\
\bottomrule
\end{tabular}
\end{center}
\end{table}

\begin{figure}[th]
	\centering

	\includegraphics[scale=0.60]{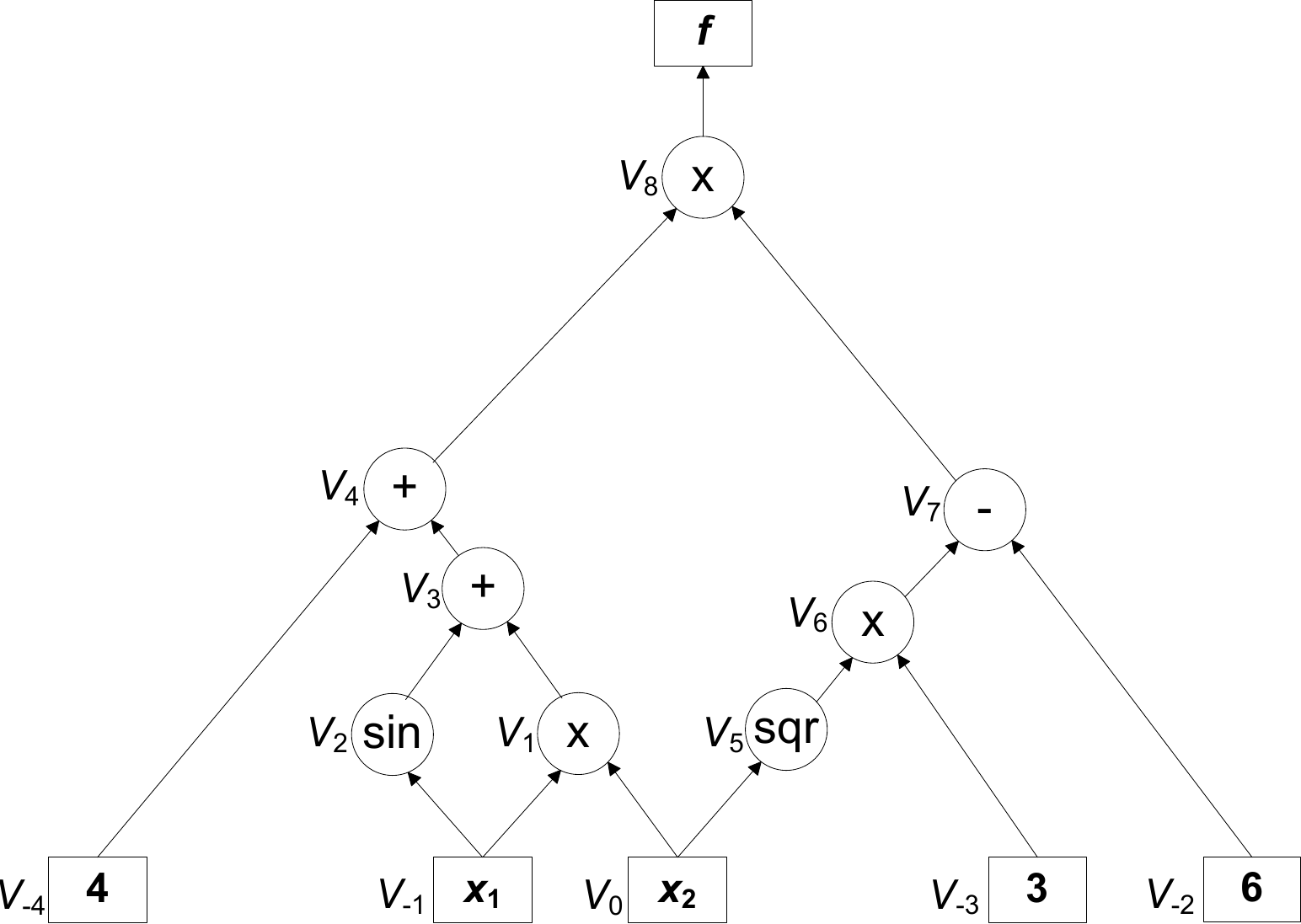}
	\caption[Algorithmic evaluation graph for a 2-dimensional sample problem]{Algorithmic evaluation graph for function: $f(x_{1},x_{2})=\left(x_{1}x_{2}+\sin x_{1} +4\right)\left(3x_{2}^{2}-6\right)$}
	\label{TreeGraphforADSampleProblemFig}
\end{figure}

Consider the tree-like graph shown in figure \ref{TreeGraphforADSampleProblemFig}, it gives a graphical representation of problem \eqref{SampleFuncForADEqn}. The root vertices represent the input variables. The intermediate vertices stand for the intermediate variables and the top vertex is the output variable. According to graph theory, the ordering require that for any vertices $a$ and $b$ linked with an arc from $b$ to $a$, then $a$ is the parent vertex while $b$ is the child. 
\begin{figure}[hbtp]
	\centering

	\includegraphics[scale=0.55]{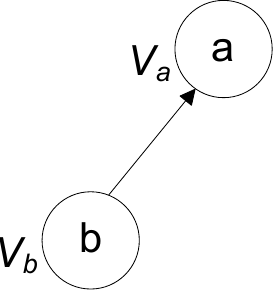}	
\end{figure}
Therefore, the value of the child vertex $b$ must be evaluated before the parent vertex $a$, i.e., the values of all the children vertices must be obtained prior to evaluating that of their parent. Hence, the overall function value can be obtained by evaluating the vertices in the graph from root through the top in an orderly manner. 

It is worth noting that when it comes to implementing the AD tool (see section \ref{ADImplementationTechniquesSec}), the user need not to explicitly break down the function into its elementary components as in the listing of table \ref{ListTraceforADSampleProblemTab}, but the identification of the intermediate variables and the graphical structure is usually done by the compiler in the application itself. Moreover, although the expressions in the listing appear symbolic, in the computer it is always the numerical value that is evaluated and stored. Yet another benefit of AD is that many of the elementary evaluations can be executed in parallel as can be seen from the graph in figure \ref{TreeGraphforADSampleProblemFig}. The level or hierarchy of the evaluations implies parallel execution, e.g. from \{$v_{2}, v_{1}, v_{5}$\}, to \{$v_{3}, v_{6}$\} and then \{$v_{4}, v_{7}$\}, and so on.

\subsection{Modes of Automatic Differentiation and their Complexities}

The two basic modes via which AD can algorithmically yield the values and derivatives of any differentiable function are the \textit{forward} and \textit{reverse} modes. Although forward mode will be adopted here, the reverse mode will also briefly be presented.

\subsection*{Forward mode of AD}

This is also called forward accumulation, it derives its name from the fact that the derivative evaluation sweeps in the same direction as that of evaluating the function value. First, the computer interprets the function (such as problem \eqref{SampleFuncForADEqn}) as a sequence of elementary operations on the work variables\footnote{For the example problem \eqref{SampleFuncForADEqn}, the work variables are: $v_{i}=\{x_{1},x_{2},4,3,6\}$.} mapped as vertices $v_{i}$ such that $i \leq 0$ for the input variables and $i>0$ for the intermediate and output variables. Then, beginning from the root vertices, both the function value and the directional derivative (i.e., the derivative with respect to any given variable augmented by the AD tool) at every vertex are simultaneously evaluated and carried forward to the top vertex. Every forward sweep will then yield the function value and its derivative with respect to the chosen independent (i.e., input) variable. 

Therefore, for a function of $n-$dimensions\footnote{The dimensionality is determined by the number of independent variables, e.g. $x_{1},x_{2},...,x_{n}$.}, evaluating its complete gradient will require $n$ forward sweeps; in our example problem \eqref{SampleFuncForADEqn}, the function has 2 independent variables and will therefore need 2 forward sweeps to get the derivatives with respect to the 2 variables.  It is easy to realize that forward mode will be more suitable for a function having several dependent variables and few independent variables but less suitable otherwise. In other words, forward accumulation is superior to reverse accumulation for functions of the form:
\begin{equation}
\label{FuncSuitableForForwardADEqn}
f:\mathbb{R}\rightarrow \mathbb{R}^{m}; \; m\gg 1
\end{equation}

The listing in table \ref{ListTraceforForwardModeADTab} shows the expressions for the forward derivatives of all the vertices propagated from the input, intermediate to the output vertex. At the beginning of every forward sweep, the derivative of one of the input vertices is seeded to $1$ and all other inputs to zero. In other words, while differentiating function \eqref{SampleFuncForADEqn} with respect to $x_{1}$, we set $\dot{v}_{-1}=1$ and  $\dot{v}_{0}=0$ and vice versa. Thereafter, the derivatives of the rest of the intermediate variables are evaluated by applying the chain rule \eqref{ChainRuleEqn}. Hence at the end of every sweep the \textit{exact} numerical values for the function and its gradient with respect to one of the variables will be accumulated.

\begin{table}[htb]
\begin{center}
	\caption[A list of values and derivatives for the forward mode AD on a sample problem]{A list of values and derivatives for the forward mode AD on problem \eqref{SampleFuncForADEqn}: $f(x_{1},x_{2})=\left(x_{1}x_{2}+\sin x_{1} +4\right)\left(3x_{2}^{2}-6\right)$}
	\label{ListTraceforForwardModeADTab}
\begin{tabular}{p{2.7cm}p{2.3cm}p{2.7cm}p{4.3cm}}
\toprule
\textbf{Variables} & \textbf{Vertices} & \textbf{Values} & \textbf{Forward Derivatives}\\
\midrule
\multirow{5}{*}{Input} & $v_{-4}$ & $4$ & $\dot{v}_{-4}=0$\\
 & $v_{-3}$ & $3$ & $\dot{v}_{-3}=0$\\
 & $v_{-2}$ & $6$ & $\dot{v}_{-2}=0$\\
 & $v_{-1}$ & $x_{1}$ & $\dot{v}_{-1}=\dot{x}_{1}$\\
 & $v_{0}$  &  $x_{2}$& $\dot{v}_{0}=\dot{x}_{2}$\\
\cline{2-4}
\multirow{7}{*}{Intermediate} & $v_{1}$  &  $v_{-1}v_{0}$ & $\dot{v}_{1}=v_{-1}\dot{v}_{0}+ \dot{v}_{-1}v_{0}$\\
 & $v_{2}$  &  $\sin v_{-1}$ & $\dot{v}_{2}=\dot{v}_{-1}\cos v_{-1}$\\
 & $v_{3}$  &  $v_{1}+v_{2}$ & $\dot{v}_{3}=\dot{v}_{1}+\dot{v}_{2}$\\
 & $v_{4}$  &  $v_{3}+v_{-4}$ & $\dot{v}_{4}=\dot{v}_{3}+\dot{v}_{-4}$\\
 & $v_{5}$  &  $v_{0}^{2}$ & $\dot{v}_{5}=2\dot{v}_{0}v_{0}$\\
 & $v_{6}$  &  $v_{5}v_{-3}$ & $\dot{v}_{6}=\dot{v}_{5}v_{-3}+v_{5}\dot{v}_{-3}$\\
 & $v_{7}$  &  $v_{6}-v_{-2}$ & $\dot{v}_{7}=\dot{v}_{6}-\dot{v}_{-2}$\\
                 
\cline{2-4} 
\multirow{2}{*}{Output} & $v_{8}$  & $v_{4}v_{7}$ & $\dot{v}_{8}=\dot{v}_{4}v_{7}+v_{4}\dot{v}_{7}$\\
   & $v_{8}$  &  $f$ &$\dot{v}_{8}=\dot{f}$\\
\bottomrule
\end{tabular}
\end{center}
\end{table}

The above principle explains the \textit{basic} forward accumulation method of evaluating derivatives. Notice from the listing in table \ref{ListTraceforForwardModeADTab} that if the value of $\dot{v}_{-1}$ is seeded to $1$ and $\dot{v}_{0}$ to $0$, then, the value of $\dot{v}_{8}$ will be the derivative of the function with respect to $x_{1}$. Conversely, if $\dot{v}_{0}$ is seeded to $1$ and $\dot{v}_{-1}$ to $0$, then the resulting value of $\dot{v}_{8}$ will yield the derivative of the function with respect to $\dot{x}_{2}$. Thus, it can easily be seen that up to $n-$forward sweeps are required to evaluate the complete gradient of an $n-$dimensional function.

A more efficient and elegant approach that will be implemented in this work, defines the initial seeds for the vertices of the independent variables as vectors rather than scalars, i.e., the vertices are set to $\dot{v}_{-1}=\left[1,0\right]$ and $\dot{v}_{0}=\left[0,1\right]$. Notice that the two vectors $\dot{v}_{-1}$ and $\dot{v}_{0}$ now correspond to the rows of an identity matrix, therefore, for a function of dimension $n$, a collective way to define the initial seeds for all the input variables is to set them to the rows of an  identity matrix $I_{n}\in \mathbb{R}^{n\times n}$. In this way, it is possible to evaluate the value and accumulate the complete gradient of a multivariate function in a single forward sweep. Details will be presented in section \ref{PropsedMatlabADImplementationSec}.

\subsection*{Reverse mode of AD}

Also called reverse accumulation, this mode acquired its name from the fact that instead of evaluating the derivative of every intermediate variable (from the root to top of figure \ref{TreeGraphforADSampleProblemFig}) with respect to a chosen input variable $x_{i}$, here an output variable is chosen and its derivative is obtained with respect to every intermediate variable from top to the root of the graph. Thus, the derivative is obtained via application of chain rule through the original evaluation trace in a backward manner. 

Therefore, the sequence for reverse accumulation is such that the function value is first evaluated and stored during a forward sweep. Then, this is followed by accumulation of the function derivative with respect to each of the intermediate and input variables during the reverse sweep. Hence, at the end of a reverse sweep, the overall gradient for the output variable (i.e. the function $f$ in this case) is derived in a single sweep from the accumulation of its partial derivatives $\nicefrac{\partial f}{\partial x_{i}}$ with respect to the independent variables $x_{i}\in \mathbb{R}^{n}$.
 
The accumulation process is achieved through an associated scalar variable $\bar{v}$ (also called \textit{adjoint}\footnote{The bar notation is used to denote a derivative. Its purpose is to differentiate the derivative term used in the reverse mode from that in the forward mode.} variable \cite{NocedalWright2006Book}) attached to every vertex. The adjoint $\bar{v}_{k}$ for the output vertex $k$ is initially set to $1$, while that of each of the remaining $k-1$ vertices is set to zero. These initial adjoint values for the $k-1$ vertices are then updated going from top (i.e. the output), through the intermediate vertices and down to the input vertices during the reverse sweep such that:
\begin{equation}
\label{ReverseADEqn}
\bar{v}_{k}=\sum_{i>k}\frac{\partial v_{i}(x)}{\partial v_{k}}\cdot\bar{v}_{i}
\end{equation}

Technically, the reverse mode resembles symbolic differentiation in the sense that one starts with the final result in the form of a formula for the function and then successively applies chain rule down the graph until the independent variables are reached. Recall from the listing for the function values in table \ref{ListTraceforADSampleProblemTab} that the value of the output vertex $v_{8}=v_{4}v_{7}$, therefore, if the adjoint of $v_{8}$ is seeded to $1$ such that $\bar{v}_{8}=1$, then, based on the definition in equation \eqref{ReverseADEqn}, the reverse derivative of any vertex, say  $\bar{v}_{7}$, is evaluated in the following manner:
\begin{equation}
\bar{v}_{7}=\sum_{i>7}\frac{\partial v_{i}(x)}{\partial v_{7}}\cdot\bar{v}_{i}=\frac{\partial v_{8}(x)}{\partial v_{7}}\cdot\bar{v}_{8}=\frac{\partial v_{4}v_{7}}{\partial v_{7}}\cdot\bar{v}_{8}=v_{4}\bar{v}_{8}.
\end{equation}
In the same way, the derivative of the input vertex $v_{0}$ is:
\begin{equation}
\bar{v}_{0}=\sum_{i>0}\frac{\partial v_{i}(x)}{\partial v_{0}}\cdot\bar{v}_{i}=\frac{\partial v_{1}(x)}{\partial v_{0}}\cdot\bar{v}_{1}+\frac{\partial v_{2}(x)}{\partial v_{0}}\cdot\bar{v}_{2}+\cdots+\frac{\partial v_{8}(x)}{\partial v_{0}}\cdot\bar{v}_{8}.
\end{equation}
Because all the vertices with the exception of $v_{1}$ and $v_{5}$ are independent of $v_{0}$ (see the listings in table \ref{ListTraceforADSampleProblemTab}), their derivative with respect to $v_{0}$ is zero. Hence, the derivative $\bar{v}_{0}$ reduces to:
\[
\bar{v}_{0}=\frac{\partial v_{1}(x)}{\partial v_{0}}\cdot\bar{v}_{1}+\frac{\partial v_{5}(x)}{\partial v_{0}}\cdot\bar{v}_{5}
\]
\begin{equation}
\bar{v}_{0}=\frac{\partial (v_{-1}v_{0})}{\partial v_{0}}\cdot\bar{v}_{1}+\frac{\partial (v_{0}^{2})}{\partial v_{0}}\cdot\bar{v}_{5}=v_{-1}\bar{v}_{1}+2v_{0}\bar{v}_{5}.
\end{equation}

\begin{table}[htb]
\begin{center}
	\caption[A list of values and derivatives for the reverse mode AD on a sample problem]{A list of values and derivatives definitions for reverse mode AD on problem \eqref{SampleFuncForADEqn}: $f(x_{1},x_{2})=\left(x_{1}x_{2}+\sin x_{1} +4\right)\left(3x_{2}^{2}-6\right)$}
	\label{ListTraceforReverseModeADTab}
\begin{tabular}{p{2.7cm}p{2.3cm}p{2.7cm}p{4.3cm}}
\toprule
\textbf{Variables} & \textbf{Vertices} & \textbf{Values} & \textbf{Reverse Derivatives}\\
\midrule

Output & $v_{8}$  & $v_{4}v_{7}$& $\bar{v}_{8}=\bar{f}=1$ \\

\cline{2-4}
\multirow{7}{*}{Intermediate} & $v_{7}$  &  $v_{6}-v_{-2}$ & $\bar{v}_{7}=v_{4}\bar{v}_{8}$ \\
 & $v_{6}$  &  $v_{5}v_{-3}$ & $\bar{v}_{6}=\bar{v}_{7}$ \\
 & $v_{5}$  &  $v_{0}^{2}$ & $\bar{v}_{5}=v_{-3}\bar{v}_{6}$ \\
 & $v_{4}$  &  $v_{3}+v_{-4}$ & $\bar{v}_{4}=v_{7}\bar{v}_{8}$ \\
 & $v_{3}$  &  $v_{1}+v_{2}$ & $\bar{v}_{3}=\bar{v}_{4}$\\
 & $v_{2}$  &  $\sin v_{-1}$  & $\bar{v}_{2}=\bar{v}_{3}$\\
 & $v_{1}$  &  $v_{-1}v_{0}$ & $\bar{v}_{1}=\bar{v}_{3}$\\
                 
\cline{2-4} 
\multirow{5}{*}{Input} & $v_{0}$  &  $x_{2}$& $\bar{v}_{0}=v_{-1}\bar{v}_{1}+2v_{0}\bar{v}_{5}$\\
 & $v_{-1}$ & $x_{1}$& $\bar{v}_{-1}=v_{0}\bar{v}_{1}+\bar{v}_{2}\cos v_{-1}$ \\
 & $v_{-2}$ & $6$ & $\bar{v}_{-2}=-\bar{v}_{7}$\\
 & $v_{-3}$ & $3$& $\bar{v}_{-3}=v_{5}\bar{v}_{6}$\\
 & $v_{-4}$ & $4$ & $\bar{v}_{-4}=\bar{v}_{4}$\\
\bottomrule
\end{tabular}
\end{center}
\end{table}
The reverse derivatives of all the remaining vertices are accumulated in the above manner and the complete listing is shown in table \ref{ListTraceforReverseModeADTab}. Notice how the sequence is reversed, i.e., while the values were initially obtained in a forward sweep, the derivatives here are evaluated beginning at the output and ending at the input vertices.

Notice that seeding the adjoint of the output vertex to $1$, (i.e. setting $\bar{v}_{8}=1$) and evaluating downward sets  the adjoints of the input variables $(\bar{v}_{-1}$ and $\bar{v}_{0})$ to be the derivatives of the function with respect to $x_{1}$ and $x_{2}$ respectively. This is where reverse accumulation method tends to supersede\footnote{Not without its set back, an initial forward sweep is required to evaluate and store the function value.} the previously described \textit{basic} forward accumulation method especially when dealing with $n-$dimensional scalar functions like:
\begin{equation}
\label{FuncSuitableForRerverseADEqn}
f:\mathbb{R}^{n}\rightarrow \mathbb{R}.
\end{equation}

One special case of the reverse accumulation principle is the \textit{backpropagation} of errors in multilayer perceptron, a commonly encountered problem in artificial intelligence and neural networks in particular.

\subsubsection*{Complexity of AD modes}

Unlike in the forward mode, an obvious drawback in the use of the reverse mode is the need to store the evaluated computational graph to be used during the reverse sweep for accumulating the gradient. A \textit{naive} implementation of the reverse mode AD may lead to a hike in the storage that is proportional to the number of operations required to evaluate the function value. A simple workaround to this storage issue is via exploitation of the sparsity in the data structure. A more advanced but rather complicated method called \textit{checkpointing} \cite{Bischof2008paper25} favours more computations. It requires partial evaluations and partial storage at the same time. In other words, checkpointing involves re-evaluating the values of the graph vertices rather than storing the entire graph structure.

When dealing with functions of type \eqref{FuncSuitableForRerverseADEqn}, theoretical analysis by Bischof et al. \cite{Bischof2008paper25} revealed that if $N$ operations are needed to evaluate the value of function $f$, then,  the total computational requirement associated with the reverse accumulation of its gradient is not more than $3N+21$ operations\footnote{This is independent of the function's dimensionality, i.e., size of $n$.}. But, the \textit{basic} forward accumulation AD will require up to $12N+10$ operations to obtain the gradient. Conversely, if the function is of type \eqref{FuncSuitableForForwardADEqn}, then forward accumulation wins. Similar argument can be found in \cite{NocedalWright2006Book}.

Besides, if the multivariate problem under consideration is a vector function of length $m$ such that:
\begin{equation}
f:\mathbb{R}^{n}\rightarrow \mathbb{R}^{m}
\end{equation}
then, the relative cost of using forward or reverse accumulation becomes similar. In fact, in such cases, it is often hard to determine the appropriate balance between the forward and reverse computations that will minimize the cost of evaluating the derivatives. This problem, also called \textit{optimal Jacobian accumulation}, is described to be NP-complete \cite{Naumann2007}.

Ultimately, a comparison of AD techniques with the method of divided differences in \cite{Bischof2002paper27} reveals that the divided differences method can only be as fast as the AD methods when the problem size is small. This is however not the case as problem size grows. Furthermore, there is no guarantee that the derivatives obtained via the divided differences approach are in anyway \textit{accurate}, whereas AD delivers \textit{exact} derivatives up to machine precision.

\subsection[Implementation Techniques of Automatic Differentiation]{Implementation Techniques of Automatic Differentiation}
\label{ADImplementationTechniquesSec}

As mentioned earlier, the key to successful implementation of AD tool is the simultaneous evaluation of the function value and its derivative at every vertex of the computational graph. There are two approaches for implementing AD tool: \textit{source transformation} and \textit{operator overloading}.  The source transformation (also called source code transformation or precompiler) method uses a precompiler that accepts and translates the given source code into a subroutine that compute both the function values and its gradients. This approach requires more tedious programming but it is said \cite{GriewankBook2008} to yield a very efficient AD implementation as it allows compile time optimization.

The operator overloading approach which is the method adopted in this work requires a new user defined data type (AD object) that combines the function value; it's gradient and Hessian in a single object. The function is decomposed into its elementary components, the value and gradient of which are evaluated by calling a series of subroutines. The operator overloading facilities\footnote{Not all languages or programming tools support operator overloading.} available in languages such as C++, C\# or high level tools like Matlab is then used to extend the capability of the built-in operators and functions to handle the user defined AD object. As can be noticed, operator overloading is significantly easier to implement than the precompiler method. It however require overloading all the arithmetic floating point unary and binary operators, and the trigonometric, logarithmic and exponential, etc. functions in order to operate on all differentiable functions.

Presently, a number of available AD implementations include the ADOL-C, ADIFOR, Tapenade, ADIC mostly in C, and the ADiMaT and MAD which are Matlab based \cite{Bischof2002paper27} implementations built on source code transformation and overloaded operators respectively.

\subsection{The Proposed Matlab Implementation of the AD Algorithm}
\label{PropsedMatlabADImplementationSec}

Matlab is a computational tool that has both the object oriented programming capability and operator overloading facility. Therefore, it is a suitable platform for the proposed AD implementation. The method adopted here is based on the vector-mode approach of forward accumulation where for a multivariate function of the form:
\begin{equation}
Y=F(x)X:\; Y\in \mathbb{R}^{m},\; X\in \mathbb{R}^{n\times n}
\end{equation}
the scalar augmented derivative term\footnote{The framework of this mode of differentiation has originated from the algebra of dual numbers \cite{Berland2006}.} $\dot{x}$ is redefined to a matrix $\dot{X}$ such that:
\begin{equation}
\dot{Y}=F'(x)\dot{X}:\; \dot{X}=I_{n\times n\times m}; \; m>1
\end{equation}
where for the input variables, the gradient $\dot{X}$ with respect to the input variables is what is initialized as $m-$array of identity matrices $I_{n\times n}$. Because the problems to be addressed here are of the form in equation \eqref{FuncSuitableForRerverseADEqn}, i.e. $m=1$, the identity matrix above is simply of dimension $n$, hence,
\begin{equation}
\dot{Y}=F'(x)\dot{X}:\: \dot{X}=I_{n}\in \mathbb{R}^{n\times n}.
\end{equation}
The exact gradient value is obtained by updating this initial gradient $(\dot{X})$ as the evaluation propagates from the input variables to the output variable.

Similarly, the expression for the second derivative (i.e., the Hessian) is as follows:
\begin{equation}
\ddot{Y}=F''(x)\ddot{X}:\: \ddot{X}=0_{n}\in \mathbb{R}^{n\times n\times n}
\end{equation}
where the Hessian of the input variables is what is initialized as $0_{n}$ which is an $n-$dimensional array of zeros. This is also updated as the evaluation propagates from the input variables, through the intermediate and up to the output variable thereby accumulating the exact numerical value for the Hessian.

Therefore, defining the AD objects in this way guarantees that evaluation of both the function value and its derivatives is completed in a single forward sweep. The three fundamental steps involved in the development of the AD algorithm via overloading method are:

\begin{enumerate}[i.]
	\item Defining a new \textit{data type} (Class) the \textit{instances} (objects) of which possess separate \textit{fields} (properties) for the values and the derivatives and can execute the user-defined overloaded \textit{functions} (methods).
	\item Defining a \textit{constructor} function that can create an instance of the above class and automatically initialize its properties.
	\item Creating overloaded functions for all the arithmetic operators and all functions (trigonometric, logarithmic, exponential etc.) upon which the AD object get its properties evaluated.
\end{enumerate}
Thus, for the proposed design, each instance of the AD class will have the following three property fields:

\begin{enumerate}[(i)]
	\item A function value field: $\mathtt{funcValue}$
	\item A function derivative field: $\mathtt{funcDerivative}$ , and
	\item A function Hessian field: $\mathtt{funcHessian}$.
\end{enumerate}

To realize a vector-form implementation for the forward accumulation, when initializing any given function of dimension $n$ at an initial point $x_{0}\in \mathbb{R}^{n}$, the constructor class for the AD objects will always initialize the object's properties as follows:

\begin{enumerate}[i)]
	\item 	$\mathtt{funcValue}$ is initialized to the initial point $f(x_{0})$.
	\item 	$\mathtt{funcDerivative}$ is initialized to an identity matrix of size $n$, i.e., $I_{n}\in \mathbb{R}^{n\times n}$.
	\item 	$\mathtt{funcHessian}$ is initialized to an array of zeros i.e., $0_{n}\in \mathbb{R}^{n\times n\times n}$.
\end{enumerate}

Thus, the AD object $F(X)$ will then be a data type that holds the function value, the derivative and the Hessian right at a spot. This model conforms to the classic extended differentiation arithmetic\footnote{The well-known concept of differentiation arithmetic \cite{Rall1986} that suggest grouping function value and derivative as an ordered pair is one of the early breakthroughs in automatic differentiation.} model proposed by Rall \cite{Rall1981, Rall1986}. Hence, 
\begin{equation}
\label{ADObjCompleteEqn}
F(X)=\left(f(x), \nabla f(x), Hf(x)\right)\equiv \left(f(x), \dot{x}f'(x), \ddot{x}f''(x)\right)
\end{equation}
where according to the extended differentiation arithmetic, any independent variable $X$ is defined as a triplet having a component for its own value $x$, one for its derivative $\nicefrac{dx}{dx}=1$ and another one for its second derivative or Hessian $\nicefrac{d^{2}x}{dx^{2}}=0$, such that:
\[
X=\left(x,1,0\right).
\]

In the same way, any constant $C$ is a triplet with value  $c$, derivative, $\nicefrac{dc}{dx}=0$ and second derivative $\nicefrac{d^{2}c}{dx^{2}}=0$, such that:
\[
C=\left(c,0,0\right).
\]
Hence, the corresponding values for the intermediate and dependent variables are obtained by operating on the above independent variables (initialized as AD objects) based on the chain rule \eqref{ChainRuleEqn}.

An algorithmic template for the AD class structure is shown in \ref{ConstructorClassForAD}. It is made up of a method that defines the three property fields for the AD objects (lines $2-6$) and a constructor function (lines $8-13$) that construct and initialize the AD objects based on the dimensionality $n$ of the problem (derived at line $9$) and other initialization settings described above. This template is based on a typical Matlab syntax.

\begin{algorithm}
\caption[A constructor class template for an automatic differentiation object]{A constructor class template for an automatic differentiation object}
\label{ConstructorClassForAD}
\begin{algorithmic}[1]
\STATE $\textbf{classdef}$ ADClass\\

	$\qquad$ \% Defining the AD object's properties - Comment\\
\STATE 	$\qquad$ $\textbf{properties}$ \\
\STATE 		$\qquad \qquad$  $\mathtt{funcValues}$\\
\STATE 		$\qquad \qquad$  $\mathtt{funcDerivatives}$\\	
\STATE 		$\qquad \qquad$  $\mathtt{funcHessian}$\\	
\STATE 	$\qquad$ $\textbf{end}$ \\

	$\qquad$ \% The Class constructor function - Comment\\
\STATE 	$\qquad$ $\textbf{methods}$ \\
\STATE 		$\qquad \qquad$  $\textbf{function}\; \textrm{ADObj}= \textrm{ADClass}(x)$\\	
\STATE 			$\qquad \qquad \qquad$ $n=\textrm{length}(x)$\\
			$\qquad \qquad \qquad$ \% Initialization of the object's  properties - Comment\\	
\STATE 			$\qquad \qquad \qquad \; \textrm{ADObj}.\mathtt{funcValues}=x$;\\
\STATE 			$\qquad \qquad \qquad \; \textrm{ADObj}.\mathtt{funcDerivatives}=\textrm{eye}(n)$;\\			
\STATE 			$\qquad \qquad \qquad \; \textrm{ADObj}.\mathtt{funcHessian}=\textrm{zeros}(n,n,n)$;\\						
\STATE 		$\qquad \qquad$  $\textbf{end}$\\
\STATE 	$\qquad$  $\textbf{end}$\\
\STATE $\textbf{end}$\\	
\end{algorithmic}
\end{algorithm}

In the following section, some overloaded arithmetic operators and functions will be presented. Using a simple example (section \ref{ExampleADSec}), a demonstration of how this automatic differentiation method can be used to evaluate the value, first and second derivatives of a given function in a single run will also be presented.

\subsection[Overloaded Operators and Functions for AD Objects]{Some Overloaded Operators and Functions for AD Objects}

Having constructed the AD objects with their property fields initialized, it is possible to execute all arithmetic operations on them so long as the built-in operators and functions (i.e., the standard real arithmetic operators and other mathematical functions) are properly overloaded to handle objects of their type. To achieve this, we overloaded several Matlab built-in operators and functions some of which are as presented in the following.

\begin{itemize}
	\item Arithmetic operators--Both the binary and unary versions
	\item Logarithmic and exponential operators
	\item Trigonometric functions
	\item Norm (ABS), etc.
\end{itemize}

The following collection presents the process of overloading some basic operators and functions. Remember that for all the arithmetic operators, basic chain rule of differentiation \eqref{ChainRuleEqn} will be applied, but for any arbitrary differentiable functions such as trigonometric, logarithmic, exponential etc., a more general formulation \eqref{ADObjCompleteEqn} will be used.

\subsubsection*{Binary Addition/Subtraction:}

Suppose the function of interest is a $2-$dimensional function $f$, such that:
\begin{equation}
\label{AddSubADExampleEqn}
f(x_{1},x_{2})=x_{1}\pm x_{2}
\end{equation}
\begin{figure}[hbtp]
	\centering

	\includegraphics[scale=0.55]{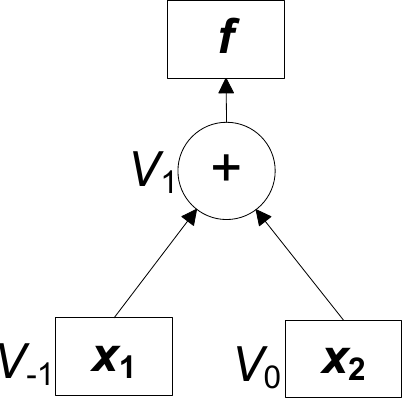}
\end{figure}
Then, the simple graph with root vertices corresponding to each of the two independent (input) variables and a top vertex corresponding to the dependent (output) variable represents the evaluation process graphically.

Let the vertices be: $v_{-1}, v_{0}$ and $v_{1}$, then, based on equation \eqref{ADObjCompleteEqn}, we have:
\[
v_{-1}=\left(x_{1}, \frac{dx_{1}}{dx_{1}}, \frac{d^{2}x_{1}}{dx_{1}^{2}}\right)=\left(x_{1}, \dot{x}_{1}, \ddot{x}_{1}\right) 
\]
and
\[
v_{0}=\left(x_{2}, \frac{dx_{2}}{dx_{2}}, \frac{d^{2}x_{2}}{dx_{2}^{2}}\right)=\left(x_{2}, \dot{x}_{2}, \ddot{x}_{2}\right) 
\]

However, since the proposed implementation seek to vectorize the definition for each of these vertices\footnote{This is the way to ensure evaluation of the complete derivatives in a single sweep of the forward accumulation.}, we initialize their fields based on the dimensionality $n$ of the function under consideration, such that:
\begin{equation}
\label{ADvectorizeGradEqn}
\dot{x}_{1}=
 \begin{bmatrix}
\frac{\partial x_{1}}{\partial x_{1}} & \frac{\partial x_{1}}{\partial x_{2}}\\
\end{bmatrix}=
\begin{bmatrix}
\dot{x}_{1} & 0
\end{bmatrix}
\end{equation}
and
\begin{equation}
\ddot{x}_{1}=
 \begin{bmatrix}
\frac{\partial x_{1}}{\partial x_{1}\partial x_{1}} & \frac{\partial x_{1}}{\partial x_{1}\partial x_{2}}\\[0.3em]
\frac{\partial x_{1}}{\partial x_{2}\partial x_{1}} & \frac{\partial x_{1}}{\partial x_{2}\partial x_{2}}\\
\end{bmatrix}=
\begin{bmatrix}
\ddot{x}_{1} & 0\\
0  &  0\\
\end{bmatrix}
\end{equation}
Similarly,
\begin{equation}
\dot{x}_{2}=
 \begin{bmatrix}
\frac{\partial x_{2}}{\partial x_{1}} & \frac{\partial x_{2}}{\partial x_{2}}\\
\end{bmatrix}=
\begin{bmatrix}
0 & \dot{x}_{2} 
\end{bmatrix}
\end{equation}
and
\begin{equation}
\label{ADvectorizeHessEqn}
\ddot{x}_{2}=
 \begin{bmatrix}
\frac{\partial x_{2}}{\partial x_{1}\partial x_{1}} & \frac{\partial x_{2}}{\partial x_{1}\partial x_{2}}\\[0.3em]
\frac{\partial x_{2}}{\partial x_{2}\partial x_{1}} & \frac{\partial x_{2}}{\partial x_{2}\partial x_{2}}\\
\end{bmatrix}=
\begin{bmatrix}
0 & 0\\
0  &  \ddot{x}_{2}\\
\end{bmatrix}
\end{equation}
Therefore, the input vertices $(v_{-1}$ and $v_{0})$ can now be redefined as:
\begin{equation}
	\label{ADobj1defnEqn}
	v_{-1}=\left(x_{1}, \begin{bmatrix}
\dot{x}_{1} & 0
\end{bmatrix}, \begin{bmatrix}
\ddot{x}_{1} & 0\\
0  &  0\\
\end{bmatrix}\right)
\end{equation}
\begin{equation}
	\label{ADobj2defnEqn}
	v_{0}=\left(x_{2}, \begin{bmatrix}
0 & \dot{x}_{2}
\end{bmatrix}, \begin{bmatrix}
0 & 0\\
0  &  \ddot{x}_{2}\\
\end{bmatrix}\right)
\end{equation}
Then, the output vertex $v_{1}$ is:
\begin{align*}
v_{1} &=\left(v_{-1}\pm v_{0}\right) \\
 &=\left( \left(x_{1}, \begin{bmatrix}
\dot{x}_{1} & 0
\end{bmatrix}, \begin{bmatrix}
\ddot{x}_{1} & 0\\
0  &  0\\
\end{bmatrix}\right) \pm \left(x_{2}, \begin{bmatrix}
0 & \dot{x}_{2}
\end{bmatrix}, \begin{bmatrix}
0 & 0\\
0  &  \ddot{x}_{2}\\
\end{bmatrix}\right) \right)\\
 &=\left(x_{1}\pm x_{2}, \begin{bmatrix}
\dot{x}_{1} & \pm \dot{x}_{2}
\end{bmatrix},\begin{bmatrix}
\ddot{x}_{1} & 0\\
0  &  \pm \ddot{x}_{2}\\
\end{bmatrix}\right)\\
 &=\left(f, \nabla f, \nabla^{2}f\right)
\end{align*}
Notice how a mere addition/subtraction of the two AD objects (i.e., vertices: $v_{-1}$ and $v_{0}$) leads to evaluation of the value, derivative and Hessian of the function $f$ \eqref{AddSubADExampleEqn} as components of the output AD object $v_{1}$. Such single sweep execution demonstrates the power of the vectorized forward accumulation of derivatives.

\subsubsection*{Multiplication Operator:}

Let the function of interest be defined as:
\begin{equation}
\label{MulADExampleEqn}
f(x_{1},x_{2})=x_{1}\times x_{2}
\end{equation}
\begin{figure}[hbtp]
	\centering

	\includegraphics[scale=0.55]{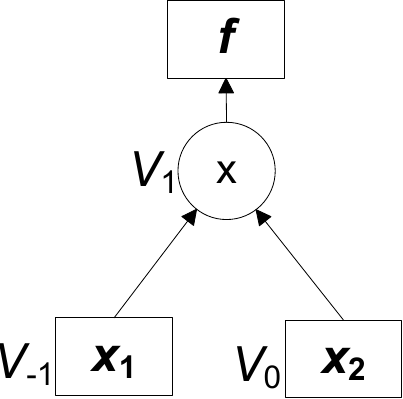}
\end{figure}
Then, the AD objects can also be seen as the vertices of the graph and their components can be defined as in equations \eqref{ADobj1defnEqn} and \eqref{ADobj2defnEqn}. The output vertex $v_{1}$ is therefore:
\[
 v_{1}=\left(v_{-1}\times v_{0}\right)=\left( \left(x_{1}, \begin{bmatrix}
\dot{x}_{1} & 0
\end{bmatrix}, \begin{bmatrix}
\ddot{x}_{1} & 0\\
0  &  0\\
\end{bmatrix}\right) \times \left(x_{2}, \begin{bmatrix}
0 & \dot{x}_{2}
\end{bmatrix}, \begin{bmatrix}
0 & 0\\
0  &  \ddot{x}_{2}\\
\end{bmatrix}\right) \right)
\]
Now, multiplying the two AD objects based on the chain rule \eqref{ChainRuleEqn} gives:
\begin{align*}
v_{1}
 &=\left(x_{1}x_{2}, \begin{bmatrix}
\dot{x}_{1}x_{2} & 0
\end{bmatrix}+\begin{bmatrix}
0 & \dot{x}_{2}x_{1}
\end{bmatrix}, \begin{bmatrix}
\ddot{x}_{1}x_{2} & 0\\
0  &  0\\
\end{bmatrix}+ \begin{bmatrix}
0 & 0\\
0  &  \ddot{x}_{2}x_{1}\\
\end{bmatrix}+\begin{bmatrix}
\dot{x}_{1}\\
0\\
\end{bmatrix}\begin{bmatrix}
0 & \dot{x}_{2}
\end{bmatrix} + \begin{bmatrix}
0\\
\dot{x}_{2}\\
\end{bmatrix}\begin{bmatrix}
\dot{x}_{1} & 0
\end{bmatrix}\right)\\
 &=\left(x_{1}x_{2},\begin{bmatrix}
\dot{x}_{1}x_{2} & \dot{x}_{2}x_{1}
\end{bmatrix}, \begin{bmatrix}
\ddot{x}_{1}x_{2} & \dot{x}_{1}\dot{x}_{2}\\
\dot{x}_{2}\dot{x}_{1}  &  \ddot{x}_{2}x_{1}\\
\end{bmatrix}\right)\\
 &=\left(f, \nabla f, \nabla^{2}f\right)
\end{align*}

\subsubsection*{The Sine Function}

Now consider the following trigonometric function (sine):
\begin{equation}
f(x)=\sin x
\end{equation}
\begin{figure}[hbtp]
	\centering

	\includegraphics[scale=0.55]{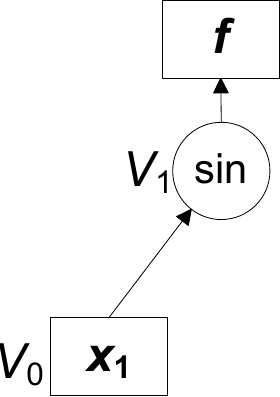}
\end{figure}
\\
The root vertex $v_{0}$ and the top vertex $v_{1}$ as shown in the graph can be defined as:
\[
v_{0}=\left(x, \dot{x}, \ddot{x}\right)
\]
and therefore, the AD variables will be defined as:
\begin{align*}
v_{1} &=\sin v_{0}\\
 &= \left(\sin x, \; \dot{x}\cos x, \; \ddot{x}\cos x - \dot{x}\dot{x}\sin x \right)\\
 &=\left(f, \nabla f, \nabla^{2}f\right)
\end{align*}

All other operators and functions are overloaded based on the principles above. In the following, an example is provided to further illustrate the application of the AD method described above by comparing it with the classical symbolic method of differentiation.

\subsection{Example: Comparing Symbolic and Forward mode AD methods}
\label{ExampleADSec}
In order to compare the computational approach of symbolic differentiation method with forward mode AD technique, consider the $2-$dimensional function \eqref{SampleFuncExampleADEqn},  in the following, the value and derivatives of this function will be evaluated at $x=\left(\pi, \pi/2\right)$ both via traditional symbolic method and AD approach described above.
\begin{equation}
\label{SampleFuncExampleADEqn}
f(x_{1},x_{2})=\left(x_{1}x_{2}+\sin x_{1} +4\right)
\end{equation}
\textbf{Symbolic Differentiation:}\\
This entails direct substitution of the solution point $x_{k}=\left(x_{1},x_{2}\right)=\left(\pi, \pi/2\right)$ after evaluating the formula for the value and the derivatives using chain rule.
\\
\\
\underline{Value:}
For the problem under consideration \eqref{SampleFuncExampleADEqn},
\[
f(x_{1},x_{2})\bigg|_{x_{k}}=\left(\pi. \pi/2 +\sin \pi +4\right)=\frac{\pi^{2}+8}{2}
\]
\underline{Gradient}: The gradient is evaluated in two stages: first, the formula is derived by applying chain rule \eqref{ChainRuleEqn} on the partial derivatives of problem \eqref{SampleFuncExampleADEqn}. Second, the solution point is substituted into the obtained formula to get the gradient.
\[
f'(x_{1},x_{2})=\begin{bmatrix}
\frac{\partial f(x_{1},x_{2})}{\partial x_{1}} &
\frac{\partial f(x_{1},x_{2})}{\partial x_{2}}\\
\end{bmatrix}=\begin{bmatrix}
x_{2}+\cos x_{1} & x_{1}\\
\end{bmatrix}
\]
\[
\therefore f'(x_{1},x_{2})\bigg|_{x_{k}}=\begin{bmatrix}
\frac{\pi}{2}+\cos \pi & \pi
\end{bmatrix}=\begin{bmatrix}
\frac{\pi-2}{2} & \pi
\end{bmatrix}
\]
\underline{Hessian}: In the same way, the Hessian is obtain as follows:
\[
f''(x_{1},x_{2})=\begin{bmatrix}
\frac{\partial^{2} f(x_{1},x_{2})}{\partial x_{1}\partial x_{1}} & \frac{\partial^{2} f(x_{1},x_{2})}{\partial x_{1}\partial x_{2}}\\[0.3em]
\frac{\partial^{2} f(x_{1},x_{2})}{\partial x_{2}\partial x_{1}} & \frac{\partial^{2} f(x_{1},x_{2})}{\partial x_{2}\partial x_{2}}\\
\end{bmatrix}=\begin{bmatrix*}[c]
-\sin x_{1} & 1\\
1  & 0\\
\end{bmatrix*}
\]
\[
\therefore f''(x_{1},x_{2})\bigg|_{x_{k}}=\begin{bmatrix}
-\sin \pi & 1\\
1  & 0
\end{bmatrix}=\begin{bmatrix}
0  & 1\\
1   & 0
\end{bmatrix}
\]
Notice how the above symbolic approach requires the computer to explicitly evaluate and store the formula before substituting the values and solving via the basic real arithmetic. In the following, the AD forward mode will be presented.
\\
\\
\textbf{Forward mode AD:}\\
Here we will redefine the problem variables (both dependent and independent) in terms of AD objects. Then, the differentiation arithmetic described above will be used to concurrently evaluate the function value and derivatives algorithmically. It will be interesting to realize how this can be achieved by simply evaluating problem \eqref{SampleFuncExampleADEqn} with the newly defined AD variables using the previously overloaded operators. Let us break the problem $\left(x_{1}x_{2}+ \sin x_{1}+4 \right)$ into the following three input variables (vertices) such that if
\[
v_{-2}=x_{1}; \quad v_{-1}=x_{2}; \quad v_{0}=4 \; 
\]
then, these vertices can be initialized as AD objects based on the solution point $x_{k}=\left(x_{1},x_{2}\right)=\left(\pi, \pi/2\right)$, such that:
\[
v_{-2}=\left(\pi, \begin{bmatrix}
1 & 0
\end{bmatrix}, \begin{bmatrix}
0  &  0\\ 0  &  0\\
\end{bmatrix}\right); \quad v_{-1}=\left(\frac{\pi}{2}, \begin{bmatrix}
0 & 1
\end{bmatrix}, \begin{bmatrix}
0  &  0\\ 0  &  0\\
\end{bmatrix}\right); \quad v_{0}=\left(4, \begin{bmatrix}
0 & 0
\end{bmatrix}, \begin{bmatrix}
0  &  0\\ 0  &  0\\
\end{bmatrix}\right)
\]
Hence, problem \eqref{SampleFuncExampleADEqn} is now
\begin{equation}
\label{VertexformForExampleADEqn}
	v_{1}=v_{-2}v_{-1}+\sin v_{-1}+v_{0}
\end{equation}
Therefore, the value and derivatives of problem \eqref{SampleFuncExampleADEqn} can now be obtained by evaluating equation \eqref{VertexformForExampleADEqn} via the differentiation arithmetic (i.e. the AD approach).
\begin{align*}
v_{1} & =\left(\pi, \begin{bmatrix}
1 & 0
\end{bmatrix}, \begin{bmatrix}
0  &  0\\ 0  &  0\\
\end{bmatrix}\right) \left(\frac{\pi}{2}, \begin{bmatrix}
0 & 1
\end{bmatrix}, \begin{bmatrix}
0  &  0\\ 0  &  0\\
\end{bmatrix}\right) \\
&\quad + \left(\sin \pi, \begin{bmatrix}
\cos \pi & 0
\end{bmatrix}, \begin{bmatrix}
-\sin \pi  &  0\\ 0  &  0\\
\end{bmatrix}\right) \\
& \quad + \left(4, \begin{bmatrix}
0 & 0
\end{bmatrix}, \begin{bmatrix}
0  &  0\\ 0  &  0\\
\end{bmatrix}\right)
\end{align*}
Therefore,
\begin{align*}
v_{1} & =\left(\frac{\pi^{2}}{2}, \begin{bmatrix}
\frac{\pi}{2} & \pi
\end{bmatrix}, \begin{bmatrix}
0  &  1\\ 1  &  0\\
\end{bmatrix}\right)+ \left(0, \begin{bmatrix}
-1 & 0
\end{bmatrix}, \begin{bmatrix}
0  &  0\\ 0  &  0\\
\end{bmatrix}\right) + \left(4, \begin{bmatrix}
0 & 0
\end{bmatrix}, \begin{bmatrix}
0  &  0\\ 0  &  0\\
\end{bmatrix}\right)\\
&=  \left(\frac{\pi^{2}+8}{2}, \begin{bmatrix}
\frac{\pi^{2}-2}{2} & \pi
\end{bmatrix},  \begin{bmatrix}
0  &  1\\ 1  &  0\\
\end{bmatrix}\right)\\
&= \left(f(x_{k}), f'(x_{k}), f''(x_{k})\right)
\end{align*}
which is similar to the solution obtained via the traditional symbolic method above. Now, from the solution $v_{1}$ which is an AD object (i.e., a structure), one can extract the function value, derivative and Hessian respectively as follows:
\begin{description}
\item Function Value =  $v_{1}.\mathtt{funcValue}=\frac{\pi^{2}+8}{2}$
\item Function Derivative = $
 v_{1}.\mathtt{funcDerivative}= \begin{bmatrix}
\frac{\pi^{2}-2}{2} & \pi
\end{bmatrix} $, and
\item Function Hessian = $v_{1}.\mathtt{funcHessian}=\begin{bmatrix}
0  &  1\\ 1  &  0\\
\end{bmatrix}$ 
\end{description}

The elegance of this approach is in its suitability for algorithmic computation in computer. Notice how the final solution yields the exact results for the function value, gradient and the Hessian. Using such exact Hessians, the proposed SQP local search algorithm could easily derive effective search directions. Hence, the advantage of this vectorized forward AD method is two fold; it is both accurate and computationally inexpensive.

\section{Contribution}

The earlier analysis of the characteristics of various gradient based local search algorithms in this chapter has helped us picked the sequential quadratic programming (SQP) algorithm which is a Newton based method. The choice was made based upon the following two findings:

First, since in the proposed hybrid setup (details in the next chapter), the SQP algorithm will be invoked after \textit{sufficient} convergence of the global algorithm, it is quite certain that the local algorithm will be initialized with a solution that is always in the vicinity of the global optimum point. Thus, whenever initialized in this way, its global convergence and quadratic rate of convergence are both assured. 

Second, to further minimize the number of iterations required by this local search method, an automatic differentiation algorithm based on the proposed vectorized forward accumulation method is used to cheaply evaluate the derivatives. This improves the quality of the evaluated search directions and alleviates the need for reverting to a basic steepest descent method after every few iterations.


This ultimately leads to the realization of the proposed local search algorithm that will ensure rapid convergence to the optimum solution by taking long but few steps.

\section{Remarks}

This chapter begins with an investigation of various types of gradient based local search algorithms. Then, the advantages and disadvantages of the steepest descent, Newton, quasi-Newton and conjugate gradient methods were analyzed. Thereafter, various methods for evaluating search directions and step sizes were investigated. Finally, a brief report on convergence analysis of the gradient based algorithms was given. Because the proposed local search SQP algorithm is designed to utilize exact Hessians during evaluation of its search directions, this chapter concluded with a broad treatment on the automatic differentiation principles for efficient evaluation of exact derivatives. The next chapter will propose and evaluate a new scheme for hybridizing the global EC algorithm presented in chapters \ref{ECAlgoChap} and \ref{ConvergenceAnalysisChap} with the SQP algorithm presented herein.


\singlespacing
\bibliographystyle{ieeetr}




\chapter[Hybridizing Evolutionary Computation Algorithms]{The Proposed Hybrid Evolutionary Algorithm}
\label{ProposedHybridAlgoChap}

Besides a brief treatment of the motivation behind the need for the development of hybrid optimization methods, this chapter will begin by introducing the current trends in hybridizing evolutionary algorithms. Taxonomies of various categories of hybrid algorithms will be presented. Thereafter, we will propose a novel approach for combining evolutionary algorithm (EC) with the SQP algorithm which is a Newton based local search optimization method. Finally, a series of experiments undertaken to evaluate the proposed hybrid system will be presented and analyzed.

\section{Why the need for hybrid algorithms}

In the last few decades, it has become well understood that population based search methods like the evolutionary algorithms (EAs) are effective in exploring search spaces even when faced with problems having high-dimensionality, nonconvexity, multimodality, isolated optima, nonuniformity and/or correlated variables. This can be attributed to the fact that at initialization, these nature-inspired methods generally try to capture a global picture of the search space. Then, during the search process they successively try to focus on the most promising regions of the search space. However, these global search methods are usually not effective in converging to the best solutions in these high quality regions \cite{Raidl2003paper92, Blum2011paper91}. On the other hand, local search methods are generally more effective when it comes to exploiting specific regions of the search space, i.e., they can easily converge to better solutions in the vicinity of any given solution. Therefore, the notion of hybridizing various categories of algorithms with the aim of establishing a robust optimization method is currently receiving wide acceptance from both the system optimization and operations research communities. In the literature, such hybrid algorithms are also referred to as hybrid \textit{metaheuristics} \cite{Blum2008paper88}.

Prior to the development of any hybrid system, it is imperative to address the following issues in order to ascertain whether a hybrid system is needed, and if so, which kind of hybridization approach is suitable for the problem under consideration; the issues are:
\begin{enumerate}[i.]
	\item Understand the type of problem at hand and based on the optimization goal, one can decide whether to use only approximate, exact or a hybrid of the two algorithms. Typically, when dealing with simple convex problems of lower dimensions, local algorithms like the gradient based methods can suffice. Also, when the quality of the final solution and computational time are not very critical, then approximate algorithms like the evolutionary algorithms are usually sufficient for many of the low to medium sized nonconvex problems. Thus, in most cases, it is only when very good solutions are needed which cannot be obtained by an exact or approximate method in a feasible time frame, the development of hybrid algorithms is advised.
	\item Determine what algorithms to combine and which type of combination of these algorithms might work well for the class of problem at hand and why.
	\item Ascertain what role enhancing the capabilities of the individual algorithms can play to the success of the proposed hybrid system.
	\item Determine how to fine tune the hybridized system to optimality for the category of problems under consideration.
\end{enumerate}

Unfortunately, not all of the above questions have direct or simple answers, in fact, the previous chapters in this work have so far concentrated on selection and tuning of the individual algorithms to be used in the hybrid system. The goal of this chapter is to look into the current hybrid methodologies and suggest a new way of combining the selected evolutionary computation algorithm (EC) with the chosen local search algorithm (SQP) presented in the previous chapters. This is hoped to yield an efficient and robust optimization algorithm suitable for medium to large scale continuous global optimization problems.

\section{Taxonomy of Hybrid Evolutionary Algorithms}

As noted by \cite{Blum2008paper88}, it is not possible to exhaustively enumerate the various types of hybrid algorithms in the literature. This is true because the notion of hybridization in itself lacks a precise definition or a specific framework that clearly defines what should constitute the so-called hybrid algorithms. Although this may sound like a drawback, it is actually thought of as the reason behind the breakthroughs achieved with these kinds of systems so far. As pointed by \cite{Blum2010paper87}, lack of precise boundary in the area of hybrid algorithms is what made the research field very rich and versatile. In other words, raising rigid boundaries between related fields of research often impedes creative thinking and exploration of new research directions.

\begin{figure}[hbtp]
	\centering

	\includegraphics[scale=0.75]{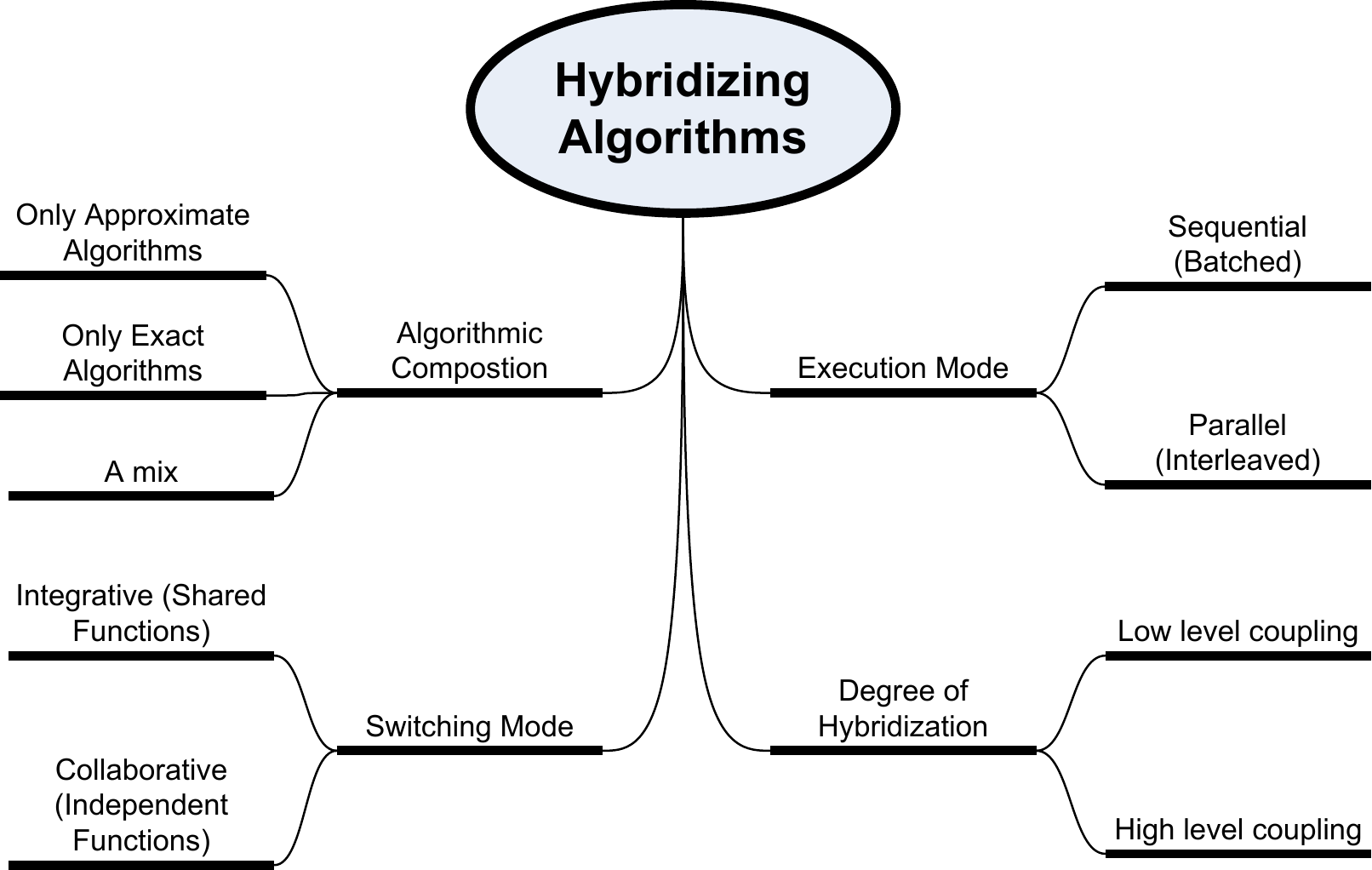}
	\caption[A concise categorization of hybrid algorithms]{A concise categorization of hybrid algorithms - Adapted from \cite{Almeida2006paper86}.}
	\label{HybridAlgosClassificationFig}
\end{figure}

Yet, a noteworthy effort made by Raidl and colleagues \cite{Almeida2006paper86} categorizes the various aspects of hybrid algorithms. They proposed a classification that attempt to unify the general framework of hybridizing algorithms. A concise schematic for the classification is shown in figure \ref{HybridAlgosClassificationFig}. The four major features depicted in this figure constitute the nature/type of the algorithms that compose the hybrid system, the switching/control method, execution mode and the extent or degree to which these algorithms are coupled. These features have in principle, virtually covered every aspect of the hybridization paradigm.

Although as earlier highlighted, in recent years, a lot of research works involving hybrid algorithms are reported in the literature. Another view of classifying hybrid approaches entails the following:
\begin{enumerate}[i.]
	\item \textbf{Hybrid metaheuristics:} Though this can be seen as a superset of the rest of the hybridization classes, it particularly refers to the combination of two or more approximate algorithms. Thus, it involves combining various nature inspired algorithms like EAs and/or non-nature inspired algorithms like tabu search, iterated local search etc. The motive is to maximize the coverage of the entire problem space during the search with the aim of exploring the promising regions. Under this class, a common way of hybridizing EC is through:
	\begin{enumerate}[a.]
		\item The development of problem specific representation \cite{Queiroz2009paper43};
		\item The development of additional genetic operators \cite{Man1999paper59}; and
		\item The incorporation of domain specific knowledge or features of classical algorithms \cite{IlSeok2004paper56}.
	\end{enumerate}
	\item \textbf{Memetic Algorithms:} May be seen as a subset of the above class, but they uniquely consist of combination of approximate algorithms like EAs with exact methods like gradient based local search algorithms. The main focus of this class is to facilitate rapid convergence to the optimum solution (i.e., exploitation of the promising regions) as soon as they are explored. Raidl et al. \cite{Raidl2003paper92} argue that although this class yields very successful hybrid algorithms, not much work exist in this direction.
	\item \textbf{Algorithms portfolio:} This notion is based on the intuition that executing many short runs of one or more algorithms (in a parallel or interleaved manner) over a prescribed solution period can provide improvement in terms of overall performance. Gomes et al. \cite{Gomes1997paper78} formulated the computational cost of a portfolio as a random variable having a probability distribution and evaluated its mean (i.e., the expected evaluation cost) and standard deviation (i.e. the dispersion or the risk involved). They argue that a skilful schedule of multiple copies of a single algorithm can outperform a hybrid made from different algorithms. Conversely, a later work by Matthew et al. \cite{Matthew2009paper80} highlighted that a portfolio essentially comprises of both scheduling and machine learning aspect. They divided the scheduling aspect into either a \textit{restart} or \textit{task-switching} schedules. The former is a schedule for a single randomized heuristic, executed in a restart model. While the latter consist of a set of one or more deterministic heuristics, each executed in a suspend-and-resume model. They argue that the suspend-and-resume model can yield better performance so long as proper scheduling is learned.
\end{enumerate}

In essence, each one of the above three classes have its merits and demerits. The first class favours exploration of problem space at the expense of exploiting high quality regions. Although the second class tried to alleviate this problem, a naive design of a memetic algorithm that is overly focused on exploitation may risk premature convergence to sub-optimal solutions. And finally, the last class have largely left the additional parameter (i.e., the scheduling aspect) to be handled in a sort of a trial and error procedure.

In table \ref{HybridAlgoSurveyTab} we present some recent hybrid algorithms reported in the literature. The table summarizes the methods from different perspectives ranging from the type, class and numbers of the combined algorithms, the adopted control/switching mode, the feature that is optimized, the validation method employed and the intended application area. Note that the summary given here is only an attempt to provide a reflection of the growing number of hybrid algorithms recently reported in the literature. It tries to capture varieties of local search algorithms usually combined with evolutionary algorithms to enhance either the efficiency or accuracy of an optimization method for various application areas or problem domain.

\begin{table}[htb]
\begin{center}
\caption[A survey of hybrid optimization algorithms in various applications domain]{A chronological survey of recent hybrid optimization algorithms in various applications domain}
\label{HybridAlgoSurveyTab}
\begin{tabular}{p{0.3cm}p{1.1cm}p{2.0cm}p{2.3cm}p{2.0cm}p{1.9cm}p{2.1cm}}
\toprule
s/n & Date  & Global \mbox{algorithm} & Local \mbox{algorithm} & Category & Optimization purpose & Validation/ Application\\
\midrule

1 & 1994\cite{Fogarty1994paper52} & GA & Hill climber  & Integrative & Efficiency & TSP \mbox{problems}\\ 
2 & 1997\cite{Wang1997paper77} & GA & Feasible path method & Collaborative & Accuracy & Numerical problems\\ 
3 & 2004\cite{Wen2004paper53} & GA & Back \mbox{propagation} & Integrative & Efficiency & Fuzzy \mbox{control}\\ 
4 & 2004\cite{IlSeok2004paper56} & GA & Search \mbox{algorithm} for feature selection (SFFS) & Integrative & Efficiency & Feature \mbox{selection}\\ 
5 & 2005\cite{Tenne2005paper50} & GA & Derivative free optimizer & Integrative & Accuracy & Numerical Problems\\ 
6 & 2007\cite{Isaacs2007paper35} & GA & Simplex \mbox{algorithm} & Collaborative& Accuracy & Numerical Problems\\ 
7 & 2008\cite{Wang2008paper54} & GA & Tabu search & Collaborative & Accuracy & Fuzzy scheduling problem\\ 
8 & 2008\cite{Kaur2008paper68} & GA & Nearest neighbour search & Collaborative & Accuracy/ Efficiency & TSP \mbox{problems}\\ 
9 & 2008\cite{Hernandez2008paper71} & NSGA II & Steepest \mbox{descent} & Collaborative & Accuracy/ Efficiency & Multiobjective optimization\\ 
10 & 2009\cite{Queiroz2009paper43} & Adaptive GA & Branch-exchange procedure & Integrative & Accuracy & Power \mbox{distribution} network problems\\ 
11 & 2010\cite{Pengfei2010paper6} & GA & Interval search & Collaborative & Accuracy/ Efficiency & Interval \mbox{optimization}\\ 
12 & 2010\cite{Pelikan2010paper79} & Bayesian optimization algorithm & Deterministic hill climber & Integrative & Accuracy & NK \mbox{landscape} problems\\

\bottomrule
\end{tabular}
\end{center}
\end{table}


\section{The Proposed Task-Switching Hybrid Evolutionary Algorithm}
\label{ProposedHybridAlgoSec}

As a first attempt, we propose combining a global evolutionary algorithm (EC) presented in chapter \ref{ECAlgoChap} with the Newton based local search algorithm (SQP) presented in chapter \ref{LocalSearchAlgoChap} in a collaborative manner. In essence, the two algorithms will run sequentially retaining their individual functionalities such that they complement each other by operating independently on the problem via data/information exchange. Initially, as a population based method, the genetic algorithm will be invoked with a randomly created initial population to provide the driving force for intense exploration of the search space. It is when the high quality regions are found and the genetic algorithm has sufficiently converged to these areas (as can be sensed by the automatic convergence detection mechanism presented in chapter \ref{ConvergenceAnalysisChap}), the local algorithm will take over and exploit the highest quality region explored by the evolutionary algorithm.

In addition, realizing that there is yet a slight tendency for the final solution returned by the local search algorithm to be a sub-optimal one\footnote{This could happen when the genetic algorithm has not properly explore the true global optimum region and the local algorithm is therefore not supplied with a good starting solution.}, a \textit{validation} loop will be utilized to kick-start an additional round of a global search by the EC algorithm. While in the validation loop, the initial population will also be randomly created and will be seeded with a copy of the best solution returned by the local algorithm and its inverted version. Thus, if the required population size is $N$, then, $N-2$ individuals will be created randomly while the remaining two will be derived from the previous run. Because the proposed evolutionary algorithm uses a binary tournament selection method with adaptive elitist strategy, the optimum solution returned at the end of the validation run will always be of equal or higher fitness than the provided seed (i.e., the solution obtained during the initial run).

\begin{algorithm}[t]
\caption[The proposed hybrid EC/SQP algorithm]{The proposed hybrid EC/SQP algorithm}
\label{HybridECSQPAlgo}
\begin{algorithmic}[1]
\STATE $\textbf{begin}$\\

	\STATE	$\qquad$ $t\leftarrow0;$\\
	\STATE	$\qquad$ initialize population $P_{1}(t)$: size $=N$\\
	\STATE	$\qquad$ get Global optimum solution: $x_{EC}\leftarrow$ invoke \textbf{EC}$(P_{1}(t))$ \qquad \quad \% see Algorithm \ref{}\\
	\STATE	$\qquad$ get Local optimum solution: $x_{SQP}\leftarrow$ invoke \textbf{SQP}$(x_{EC})$ \qquad \; \% see Algorithm \ref{}\\
	\STATE	$\qquad$ re-initialize population $P_{2}(t)$: size $=N-2$\\
	\STATE	$\qquad$ validate $x_{SQP}$: $x^{*}\leftarrow$ invoke \textbf{EC}$(P_{2}(t) \cup x_{SQP} \cup \bar{x}_{SQP})$ \quad \% $x^{*}$ is the optimum solution\\

\STATE $\textbf{end}$\\
\end{algorithmic}
\end{algorithm}

Algorithm \ref{HybridECSQPAlgo} demonstrates the working of the proposed hybrid switching procedure. The proposal entails designing a memetic algorithm that relies on a task switching procedure to transfer control between the global and local algorithms. It begins with an initial randomly created population $P_{1}(t)$ of size $N$ from which an initial solution $(x_{EC})$ is derived using the EC algorithm (lines: 3-4). After termination of the EC algorithm, the search switches to the SQP algorithm which will be fed with $x_{EC}$ as its initial solution point (line 5). The solution obtained by the SQP algorithm $(x_{SQP})$ is then validated by running another instance of the EC algorithm. However, the size of the new randomly created initial population $P_{2}(t)$ for the validation run is $N-2$ (line 6). This will be complemented with a copy of $x_{SQP}$ and its mutated version $\bar{x}_{SQP}$ to maintain a uniform population size of $N$ individuals. The search process stops and returns $x^{*}$ as the true optimal solution when validation is complete (lines 7).

\section{Experiments}

An initial evaluation of the proposed hybrid algorithm is done by applying it on some benchmark numerical optimization problems. Details of the parameter settings of the hybrid algorithm used for the experiments are presented in table \ref{HybridAlgoParamSetTab}.

\subsection{Characteristics of Global Optimization Test Problems}

Before evaluating any global optimization algorithm, it is important to seek a suite of test problems that satisfy some of the following qualities:
\begin{enumerate}
	\item \textbf{Nonlinearity:} The function should not be linear
	\item \textbf{Scalability:} The dimensionality of the test function must be extendible to medium or large sizes
	\item \textbf{Non-separability:} A function is separable if it has no nonlinear interaction between its variables
	\item \textbf{Multimodality and Nonconvexity:} The function must possess many sub-optimal peaks/valleys
	\item \textbf{Non-symmetricity:} The global optimum should not be equidistant from any oppositely located local optima pair
	\item \textbf{High Dispersion:} Dispersion indicates how far from being convex is the global topology of the function. It was argued that \cite{Lunacek2006paper106}, a highly dispersed function can be more difficultly to global optimization algorithms than just a multimodal function having convex-like global structure.
\end{enumerate}
Detailed analysis and evaluation of the above features and more for test problems can be found in \cite{Whitley1996paper108}.

\begin{table}[htb]
\begin{center}
\caption[Parameter settings of the proposed hybrid algorithm]{Parameter settings of the proposed hybrid algorithm (where $f=$ fitness, $t=$ generations, $l=$ string length, MaxGen = Maximum generations limit)}
\label{HybridAlgoParamSetTab}
\begin{tabular}{p{1.9cm}p{4.3cm}p{7.1cm}}
\toprule
\textbf{Algorithms}  & \textbf{Parameters} & \textbf{Values/settings}\\
\midrule
\multirow{6}{*}{\textbf{Global (EC)}} & Representation & Binary encoding with mapping function\\
            & Population size & 100\\
            & Selection method & Binary tournament selection\\
            & Crossover: Type, Rate & Single point,  $P_{c}=1.0$\\
            & Mutation: Type, Rate & Bit mutation, $P_{m}=\nicefrac{1}{l}$\\
            & Replacement scheme & Adaptive elitist strategy (see section \ref{ProposedAdaptElitism})\\ \cline{2-3}
\multirow{2}{*}{\textbf{Switching}} & Crossover's contribution to fitness $(\sigma q_{Crossover})$ & when: $\sigma q_{Crossover}\le 0.01$ (see section 			 \ref{ProposedConvergenceThresholdSec})\\
\textbf{criteria}	& Stalled generations &  when: $||f_{max}(t)-f_{max}(t-20)||\le 0.001$ \\
	& Maximum generations & when: Number of Generations $\ge$ MaxGen\\ \cline{2-3}
\multirow{4}{*}{\textbf{Local (SQP)}} & Search direction type & Newton directions\\
			& Search direction evaluation & Exact Hessians: $B_{k}=\nabla^{2} f(x)$, (equation \eqref{GeneralizedSearchDirectionEqn})\\
			& Step size & Wolf conditions $0<\alpha \le 1$: (see section \ref{EvaluatingStepLengthParameterSec})\\
			& Stopping criteria & when: $||\nabla f(t)||-||\nabla f(t-1)||\le 0.001$ or $||d_{k}(t)||-||d_{k}(t-1)||\le 0.001$\\

\bottomrule
\end{tabular}
\end{center}
\end{table}

\subsection{Selected Benchmark Test Problems}

To evaluate the performance of the proposed hybrid algorithm, a series of tests have been carried out on the Ackley, Rastrigin and Schwefel benchmark functions for global optimization.

\begin{enumerate}[i.]
	\item \textbf{Ackley function:} The generalized form of Ackley function is defined as:\\
	\begin{equation}
	\label{AckleyFunctionEqn}
f(x)=20+e^{1}-20 \cdot e^{-0.2\cdot \sqrt{\frac{1}{n}\sum_{i=1}^{n}x_{i}^2}}-e^{\frac{1}{n}\cdot \sum_{i=1}^{n}\cos (2\pi x_{i})}
\end{equation}
where $-15\le x_{i}\le 30 : i=1,2,\dots,n$ and global optimum $x^{*}=(0,\dots, 0), f(x^{*})=0.$\\
	Ackley function is a multi dimensional nonlinear global optimization problem. Although the function is multimodal having many peaks/valleys forming several sub-optimal solutions, it is has low dispersion with a unimodal global topology\footnote{i.e. it is pseudo-convex, Ackley function gets more convex as the dimensionality is increased, thus, solving its higher dimensional versions is always easier} and it is symmetric such that the global optimum is centrally surrounded by the local optimum points. However, Ackley function is scalable, nonlinear, multimodal and non-separable. These qualities have made it a difficult problem that can easily deceive global optimization algorithms to get trapped at sub-optimal solutions.
	\item \textbf{Rastrigin function:} The generalized form of Rastrigin function is defined as:\\
	\begin{equation}
	\label{RastriginFunctionEqn}
f(x)=10\cdot n+\sum_{i=1}^{n}\left(x_{i}^{2}-10\cdot \cos (2\pi x_{i})\right)
\end{equation}
where: $-5.12\le x_{i}\le 5.12 : i=1,2,\dots,n$ and global optimum $x^{*}=(0,\dots, 0), f(x^{*})=0.$\\
	Rastrigin function is a scalable, nonlinear and highly multimodal function with many valleys increasing in depth when approaching the global optimum point. Although it is separable and has a global topology (i.e. pseudo-convexity) it is actually flatter and of higher dispersion compared to Ackley function and therefore more difficult.
	\item \textbf{Schwefel function:} The generalized form of the Schwefel function is:\\
	\begin{equation}
	\label{SchwefelFunctionEqn}
f(x)=418.9829\cdot n-\sum_{i=1}^{n}x_{i}\cdot \sin \left(\sqrt{|x_{i}|}\right)
\end{equation}
where: $-500\le x_{i}\le 500 : i=1,2,\dots,n$ and global optimum $x^{*}=(1,\dots, 1), f(x^{*})=0.$\\
	Also nonlinear, separable and multimodal, Schwefel function is highly dispersed and lacks any global topology. Moreover, it is non-symmetric and scalable to higher dimensions.
\end{enumerate}
If we sort the complexity of the above problems based on their degree of dispersion, the Ackley function is the least and the Schwefel function is the most difficult.

\subsection{Results and Discussions}

Two different sets of experiments were carried out. The first experiment investigates how the proposed hybrid algorithm behaves under increasing problem size.  While the second experiment is aimed at evaluating the performance of the hybrid algorithm (EC/SQP) by comparing it with a standard evolutionary computation (EC) algorithm and the well known covariance matrix adaptation algorithm (CMA-ES) on the three benchmark test problems presented above. All experiments are run $100$ times and averaged results are reported for statistical significance.

\subsubsection*{Experiment 1: Scalability Test}

\begin{figure}[htbp]
  \centering
  
\subfloat[Ackley Function (2, 10 and 100-Dimensions)]{\label{fig:gull}\includegraphics[width=0.53\textwidth]{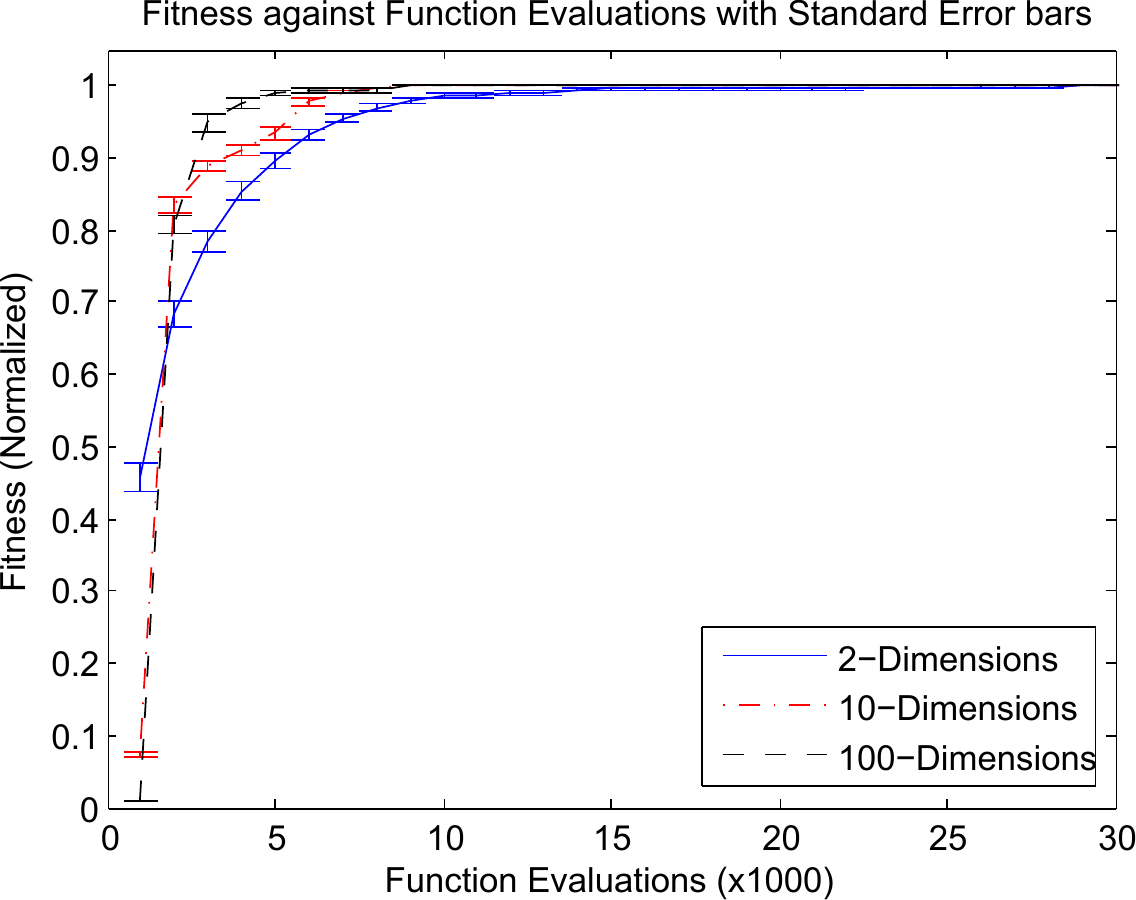}}\\ 
  \vspace*{0.20cm}               
  \subfloat[Rastrigin Function (2, 10 and 100-Dimensions)]{\label{fig:tiger}\includegraphics[width=0.53\textwidth]{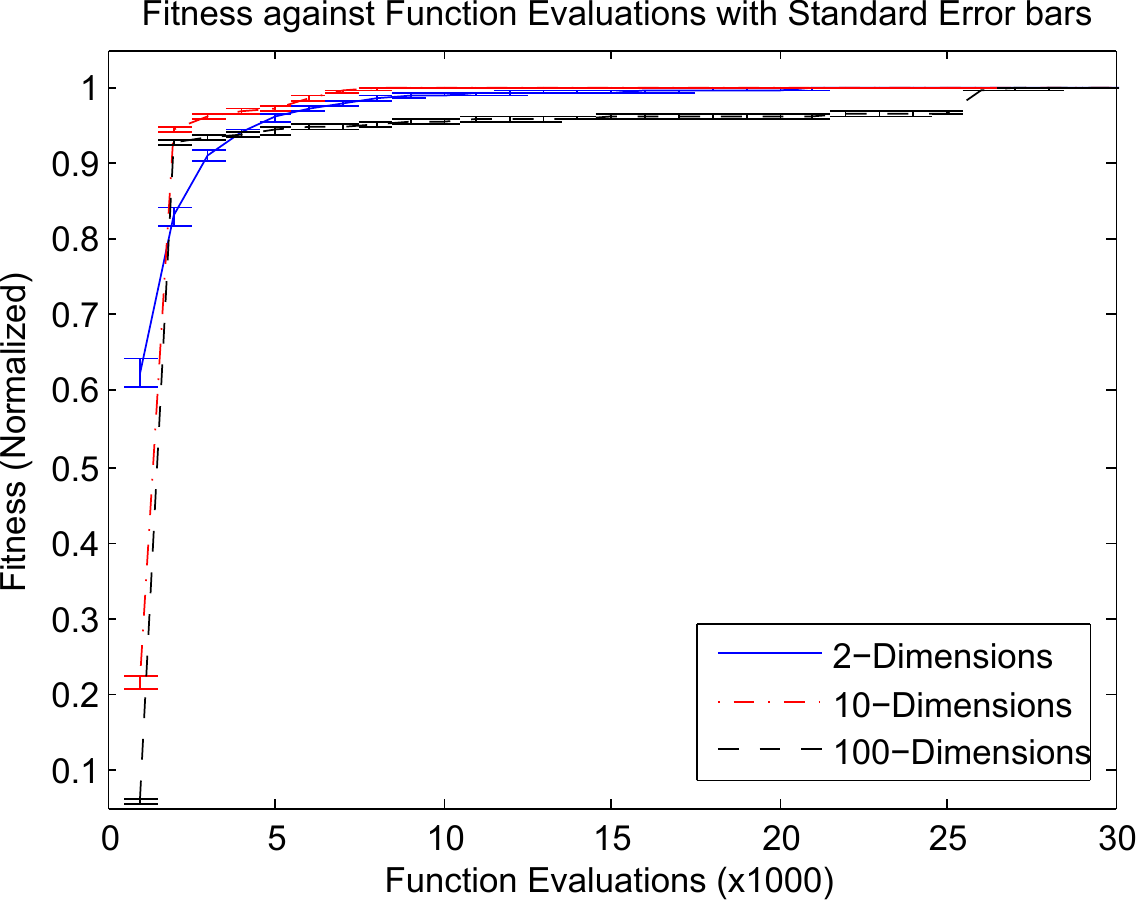}}\\
  \vspace*{0.20cm}
  \subfloat[Schwefel Function (2, 10 and 100-Dimensions)]{\label{fig:mouse}\includegraphics[width=0.53\textwidth]{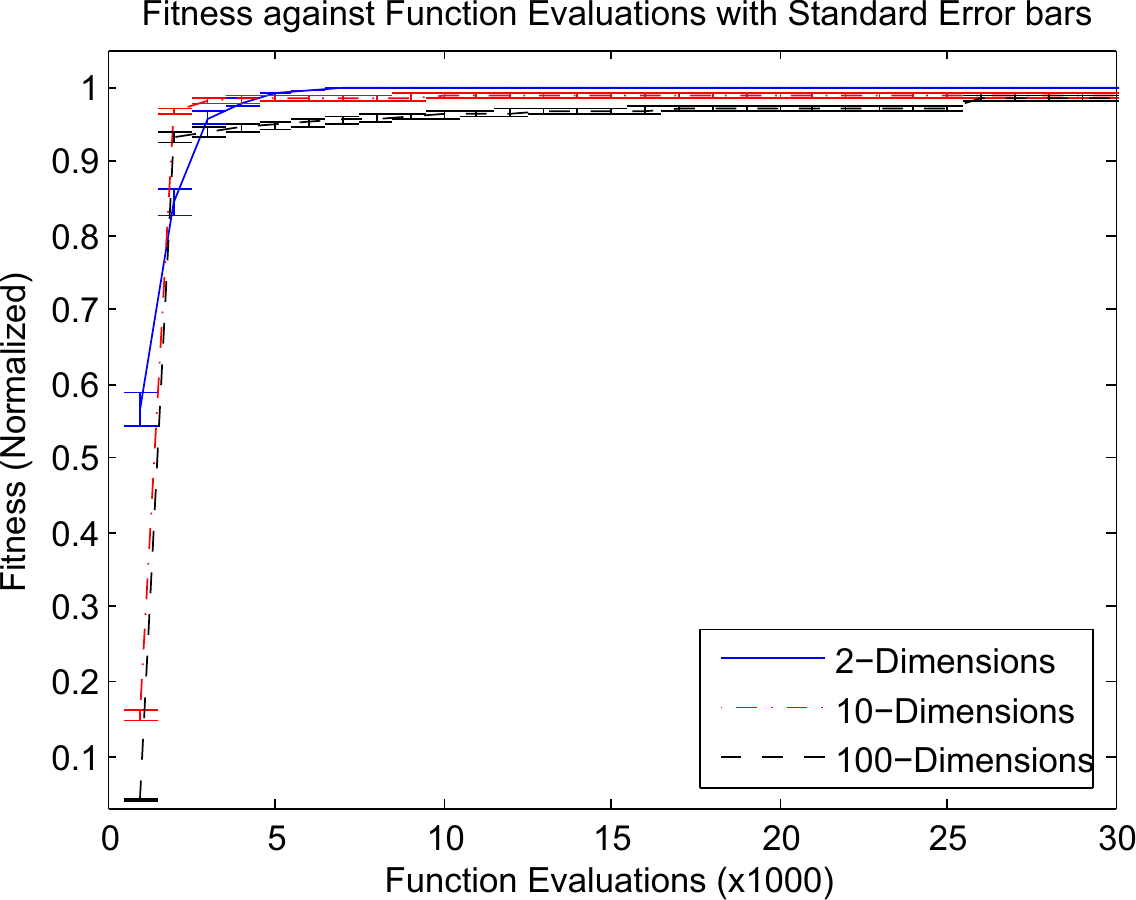}}  
  \caption[Hybrid EC/SQP hybrid algorithm: Scalability test]{Hybrid EC/SQP hybrid algorithm evaluated under increasing problem sizes $(2, 10, 100)$ of: (a) Ackley, (b) Rastrigin and (c) Schwefel test functions. The error bars $(\mathtt{I})$ represent the standard errors of the mean as all the results in these plots are averages of 100 independent runs. The plots in this figure show that the performance of the EC/SQP algorithm is fairly stable and immune to varying problem sizes.}
  \label{hybridScalabilityTestFig}
\end{figure}

With the aim of testing the robustness of the proposed method under increasing problem size, we seek to optimize the Ackley test problem using the EC/SQP algorithm with the problem size set to $2$, $10$ and $100$ dimensions. The results of this experiment are shown in figure \ref{hybridScalabilityTestFig}. The tests are repeated $100$ times with each having an entirely new randomly created initial population. The fitness plots for the averages obtained from the $100$ independent runs of the $2$, $10$ and $100$ dimensional Ackley function are plotted for comparison purposes. Similar test is carried out on the Rastrigin and Schwefel functions, see figure \ref{hybridScalabilityTestFig}.

\subsubsection*{Experiment 2: Performance Comparison Test}

\begin{figure}[htbp]
  \centering

\subfloat[Ackley Function (10-Dimensions)]{\label{fig:gull}\includegraphics[width=0.52\textwidth]{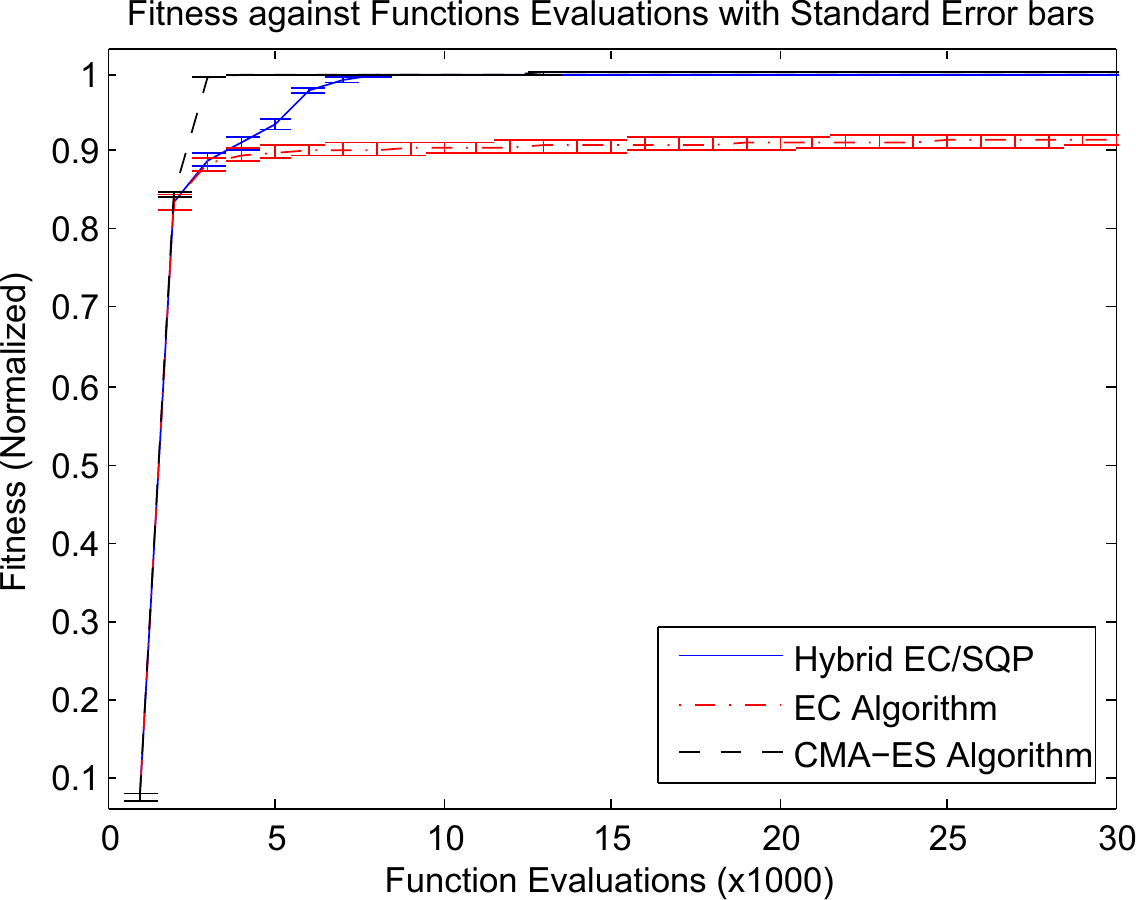}}                
  \hspace*{-0.20cm} \subfloat[Ackley Function (100-Dimensions)]{\label{fig:gull}\includegraphics[width=0.52\textwidth]{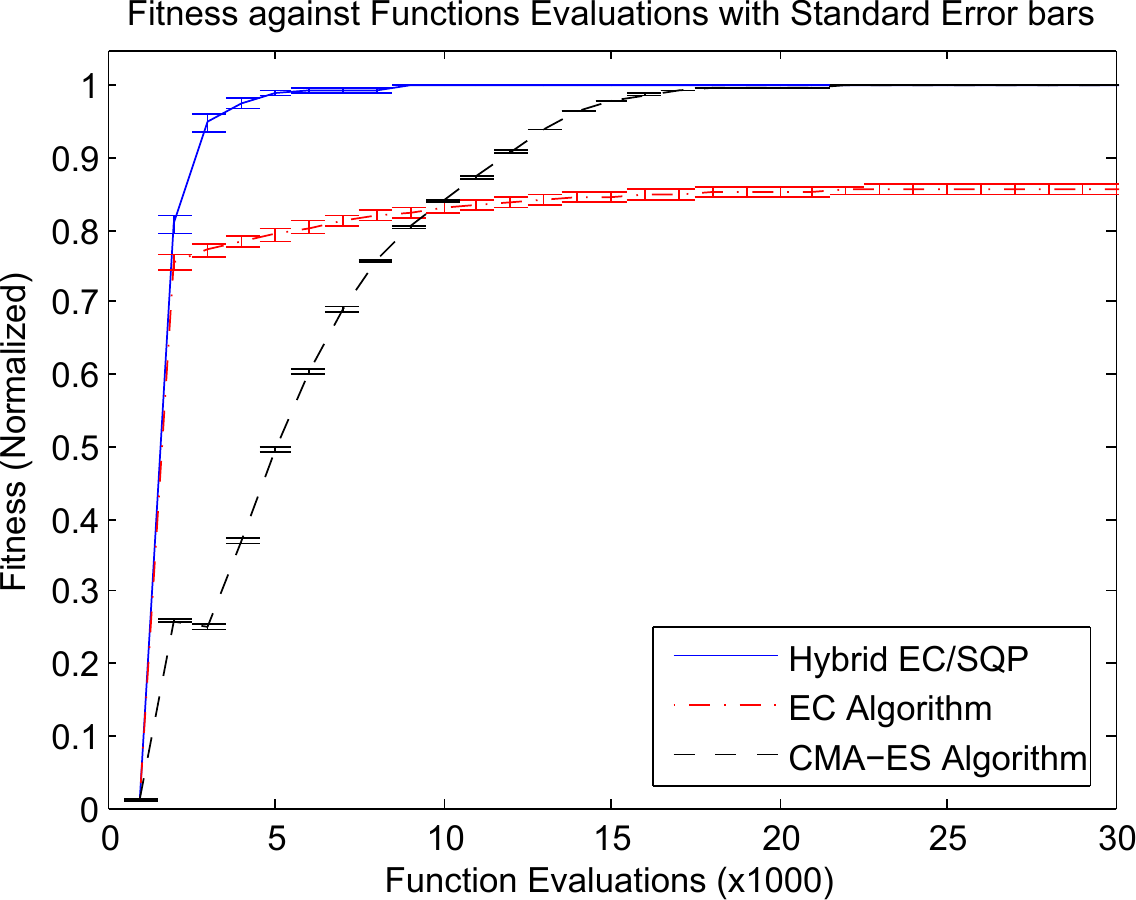}}\\
  \vspace*{0.20cm}
  \subfloat[Rastrigin Function (10-Dimensions)]{\label{fig:tiger}\includegraphics[width=0.52\textwidth]{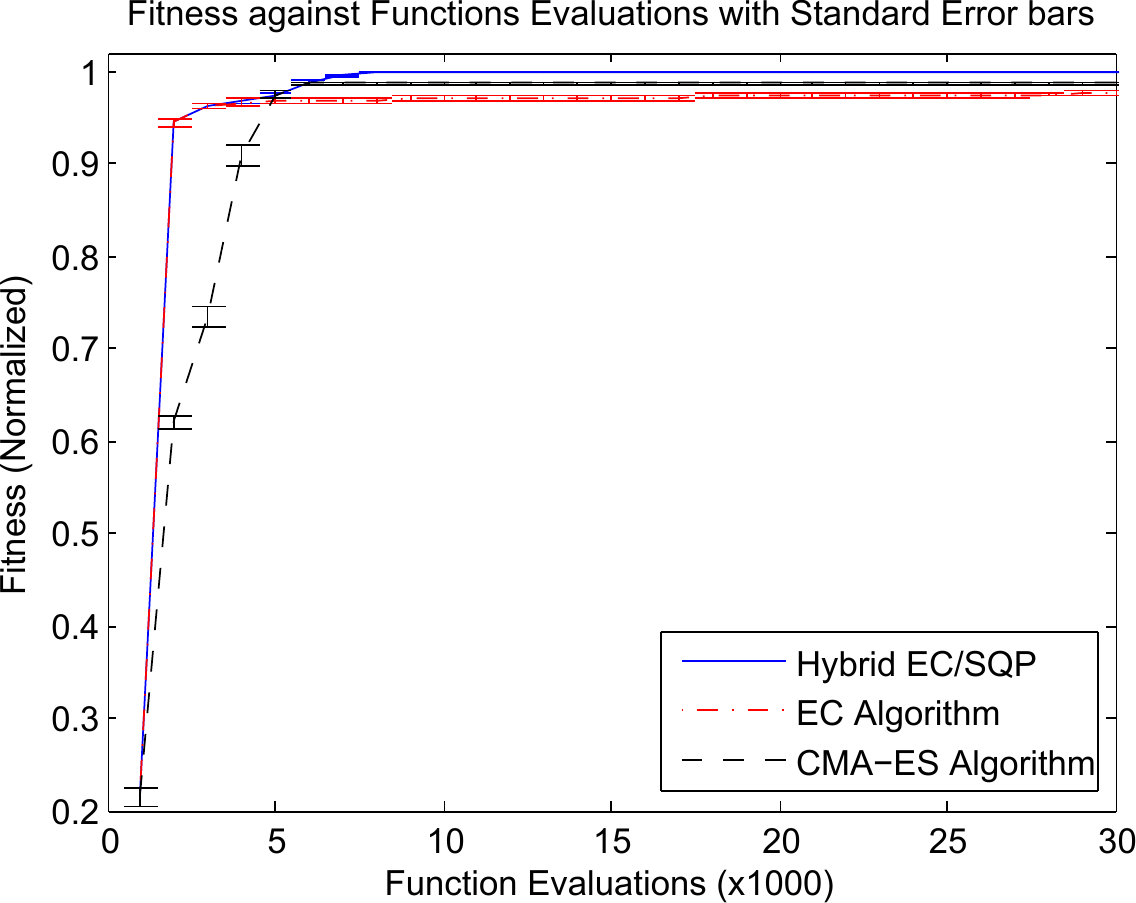}}
  \hspace*{-0.20cm} \subfloat[Rastrigin Function (100-Dimensions)]{\label{fig:tiger}\includegraphics[width=0.52\textwidth]{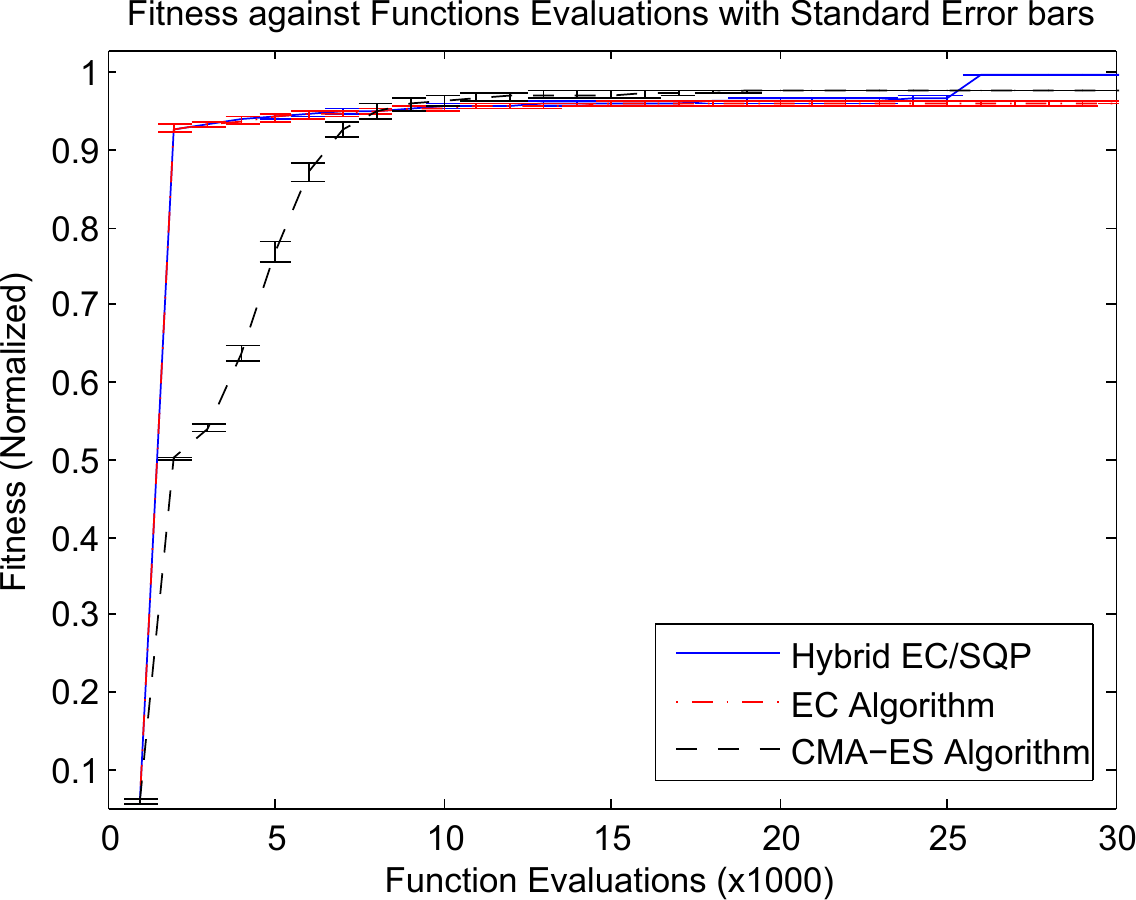}}\\
  \vspace*{0.20cm}
  \subfloat[Schwefel Function (10-Dimensions)]{\label{fig:mouse}\includegraphics[width=0.52\textwidth]{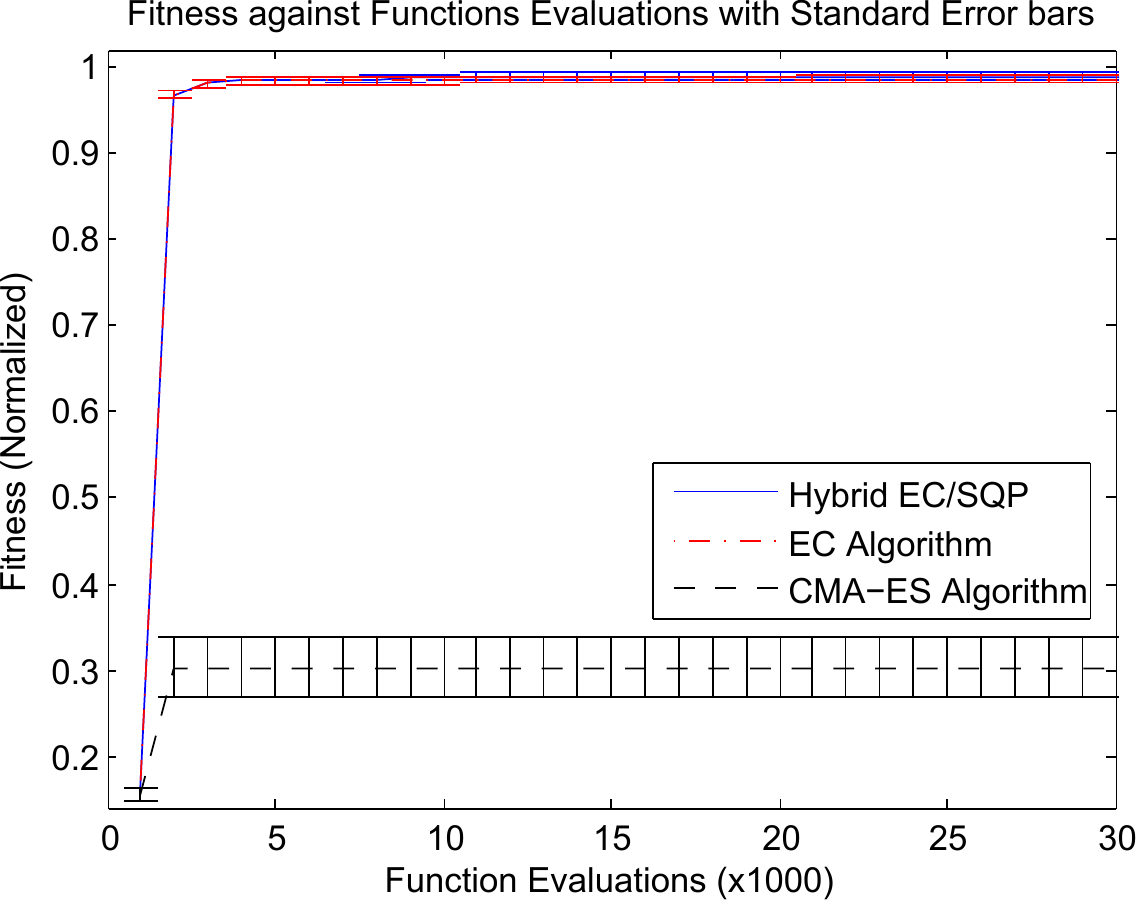}}
  \hspace*{-0.20cm} \subfloat[Schwefel Function (100-Dimensions)]{\label{fig:mouse}\includegraphics[width=0.52\textwidth]{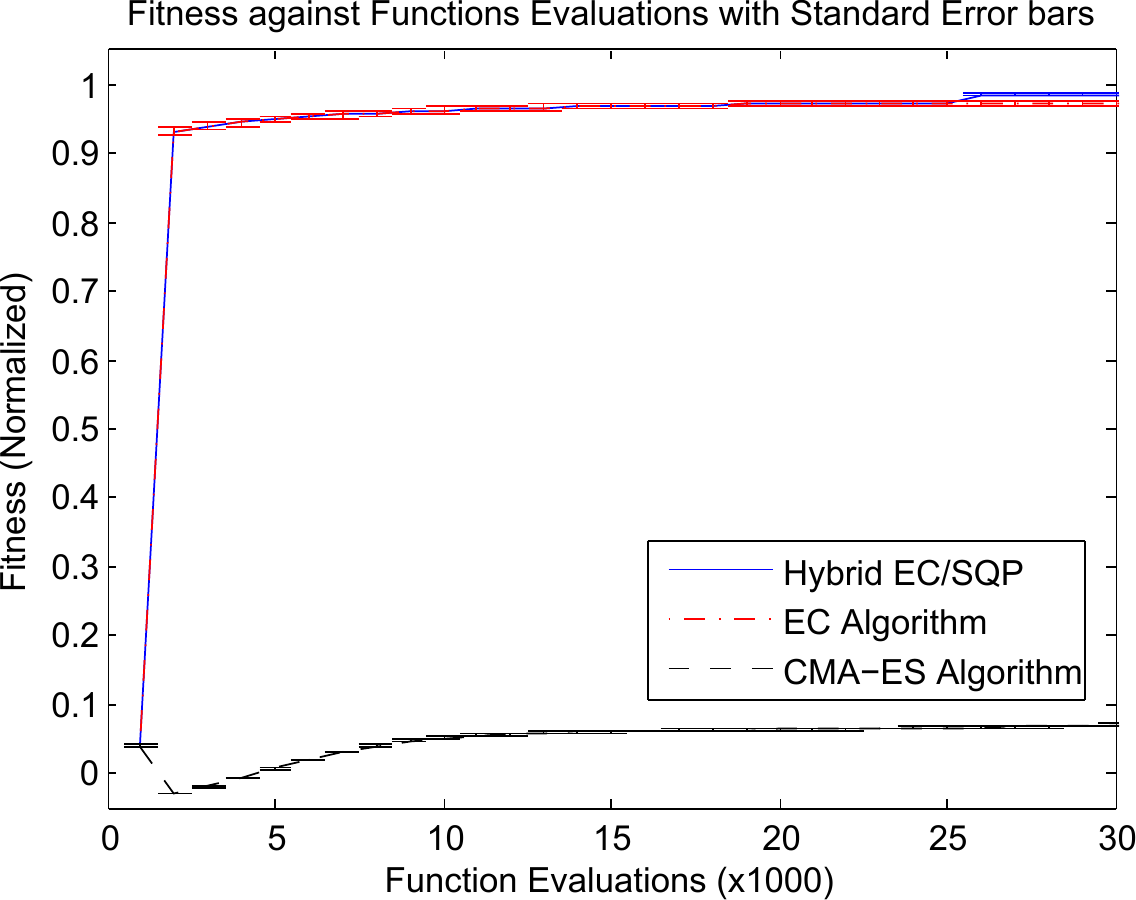}}
  
  \caption[Hybrid EC/SQP hybrid algorithm: Fitness comparison test]{Comparing fitness plots for Hybrid EC/SQP, standard EC and CMA algorithms\newline
  Test Problems (row-wise):  Top: Ackely, Middle: Rastrigin, Bottom: Schwefel\newline
Problem sizes (column-wise): Left: 10-Dimensions, Right: 100-Dimensions\newline
The error bars $(\mathtt{I})$ represent the standard errors of the mean as all the results in these plots are obtained  by averaging 100 independent runs. The plots in this figure show that for virtually all the test problems, the performance of the proposed EC/SQP algorithm is always as good if not better than any of the other two algorithms.}
  \label{HybridFitnessCompTestFig}
\end{figure}

This is a fitness comparison test where the proposed EC/SQP algorithm is compared to a standard evolutionary algorithm EC and the well known evolutionary strategy method called CMA-ES algorithm. The results of this experiment are shown in figure \ref{HybridFitnessCompTestFig}. Similar to experiment 1, we attempt to optimize the Ackley, Rastrigin and Schwefel benchmark functions with each of the three algorithms allowed to run for up to $30,000$ function evaluations. The parameters of both the EC and the CMA algorithms are tuned to optimality for the test problems.

\subsubsection*{Discussions}

 This study provided additional evidence on the fact that a skilful design of a hybrid system that combines the strengths of population-based and local search methods could yield an efficient and robust optimization system. The proposed hybrid EC/SQP algorithm combines a population-based global algorithm (EC) with a gradient-based local search method (SQP) in a collaborative manner. Results show that EC/SQP seems to be suitable for small, medium and large scale continuous optimization problems. This may be a manifestation of the well-known no free lunch theorem \cite{WolpertMacreadyNFL1997} which may simply be put as: \textit{two hands are better than one}. Recall that prior to the development of evolutionary methods and their hybrids, the notion of developing widely applicable optimization methods barely exists. Researchers were often focused on optimization methods that are mainly tailored to a specific problem category.  But the results from this investigation seem supportive of our earlier suppositions that the performance of hybrid methods scales well to varying problem sizes, categories and/or levels of complexities. 
 
A careful look at the results of experiment 1 (see figure \ref{hybridScalabilityTestFig}) reveals that for virtually all the varying sizes of the three different test problems, up to $90\%$ of the maximum attainable fitness level was reached within the first $5000$ functions evaluations.  This is indicative of the speed at which the proposed EC/SQP algorithm approaches the optimal solution point. Notice, however, that with the exception of the Ackley function in plot (a), increase in problem size leads to an increase in the number of function evaluations required to reach the optimum solution. This indicates that to some extent, larger problem sizes might inhibit the performance of the proposed hybrid algorithm. But this is consistent with the fact that increase in problem size widens the search space making it harder for algorithms to explore and narrow down to the region of the true optimum solution. Nevertheless, the decrease in performance is rather minimum considering the $10$ times increase in the problem size.

A possible explanation of the discrepancy shown in plot \ref{hybridScalabilityTestFig}(a) by the Ackley test function might be related to the fact that it is a pseudo-convex function with several local optima induced via its cosine component (see equation \eqref{AckleyFunctionEqn}). These local optimum valleys become shallow and smooth as its dimensionality increases. Thus higher dimensions of Ackley functions (i.e. with variables ranging from $10$ to $100$) appear easier to optimize than their lower dimensions counterparts (such as the $2$ variables case). A similar behaviour exhibited by Griewangk benchmark test function was reported by Whitley et al. \cite{Whitley1996paper108}.

Turning now to the findings from experiment $2$, the simulation results for the comparison tests are depicted in the six plots shown in figure \ref{HybridFitnessCompTestFig}. Notice that for the three test problems under consideration (see left column), in overall, the proposed hybrid EC/SQP algorithm shows quite a significant performance improvement over the standard EC and CMA-ES algorithms. Worth noting is that this remarkable performance seems to generalize over the varying problems' sizes (see right column).

However, for such carefully selected categories of test problems, a critical analysis of the performance of each of the three algorithms requires cautious treatment. This is necessary because although all the three test problems are nonlinear, scalable, and multimodal in nature, they are quite different in complexity when viewed on the basis of their dispersion.  In other words, a possible categorization of the plots in figure \ref{HybridFitnessCompTestFig} is to place the Rastrigin function (mid row) as a moderately dispersed problem, while the Ackley (top row) and Schwefel (bottom row) test functions are respectively the least and most dispersed problems. Consequently, they stand at the opposite extremes of complexity. 

Beginning with the Rastrigin function (figure \ref{HybridFitnessCompTestFig} mid row), the observed remarkable performance of all the three algorithms on this test problem may be attributed to its moderate level of dispersion. This could be why it seems easy to the three different global optimization approaches. Nonetheless, it can be noticed that the proposed EC/SQP algorithm appears to lead; the CMA-ES follows which is then followed by the standard EC algorithm.

With regard to the Ackley function (figure \ref{HybridFitnessCompTestFig} top row), one might notice that while the hybrid EC/SQP and CMA-ES algorithms clearly outperform the standard EC algorithm, the excellent efficiency exhibited by the CMA-ES algorithm (figure \ref{HybridFitnessCompTestFig}(a)) seem to deteriorate following the $10$ times increase in the problem size (figure \ref{HybridFitnessCompTestFig}(b)). This, however, is not the case with the EC/SQP algorithm. Hence, it could conceivably be hypothesised that the hybrid nature of the EC/SQP algorithm might be the reason behind its immunity to the increase in problem size.



It is interesting to note that for all the test cases considered in this experiment (figure \ref{HybridFitnessCompTestFig}), the standard EC algorithm always lags behind the other two algorithms except on the \mbox{Schwefel} function (bottom row). Surprisingly, this is where the CMA-ES algorithm exhibited least performance. This rather difficult to interpret result might have several possible explanations. It could be due to the fact that Schwefel function has the highest level of dispersion among all the three test cases. Therefore, it is a flat multimodal function that lacks any unimodal global topology (i.e., it does not exhibit any pseudo-convexity).

It remains unclear, though, why high dispersion problems like Shwefel function seem to be easy for the standard EC algorithm (see figure \ref{HybridFitnessCompTestFig} bottom row). But one possible reason for the sudden decline in the performance of the CMA-ES algorithm may be due to its excessive evaluation of infeasible solutions during the early stages of the search process. It is therefore thought that CMA-ES algorithm relies on exploiting global convexity to be successful. Hence, its remarkable performance on the low dispersion Ackley function (top row) deteriorates when faced with high dispersion problems like Schwefel. 

These results corroborate the findings of some previous studies in this field. An investigation on the CMA-ES algorithm by \cite{Lunacek2006paper106} revealed that the adaptive step-size heuristic, called \textit{cumulation}, does not function as intended when the best regions of the search space are too spread out (such as in a high dispersion problems like the Schwefel functions). Nonetheless, it was reported elsewhere \cite{HansenStefan2004paper105},  that although CMA-ES algorithm may need more than $10^5$ function evaluations for such high dispersion problems, it will at some point converge to the true global solution if sufficient evaluations are granted.

Noteworthy, from virtually all the results in these experiments (see figure \ref{HybridFitnessCompTestFig}), the moment at which the switching from the global algorithm (EC) to the local algorithm (SQP) took place might easily be noticed. Thus, one could appreciate the remarkable contribution of the local algorithm towards the overall success of the hybrid EC/SQP algorithm under various categories of test problems. 

Ultimately, the results presented herein illustrate that the proposed hybrid EC/SQP optimization method is a robust and efficient novel approach that can effectively complement the traditional global optimization methods.

\section{Contribution}

This first attempt to realize the proposed hybrid optimization algorithm was possible following an intuitive amalgamation of the previously presented evolutionary (EC) algorithm and the gradient based (SQP) local search method. Initially, we focus the design of the individual algorithms in such a way that the EC algorithm is optimized for exploration of problem space while the SQP algorithm is optimized for rapid exploitation of the high quality regions of the problem space. Then, by further tuning of the combined parameters of the two algorithms, while in collaboration, the requisite balance in exploration and exploitation for robust optimization has been greatly enhanced. This would not have been possible without:
\begin{itemize}
	\item The new convergence detection method (proposed in chapter \ref{ConvergenceAnalysisChap}, section \ref{AutoConvergDetcSec}) that detects convergence of the EC algorithm, and in due course, shifts control from the EC to the SQP algorithm adaptively, and
	\item The incorporation of the novel validation routine (section \ref{ProposedHybridAlgoSec}) which essentially further substantiates the quality of the solution returned by the local algorithm.
\end{itemize}
But above all, is the fact that the experimental results presented in this chapter have given us a glimpse of what to expect in the future when the proposed method is further refined.

\section{Remarks}

This chapter has presented a broad review of various techniques of hybridizing evolutionary algorithms. Particularly, a multidisciplinary survey of their applications in the recent years was presented. At a glance, an investigation of some of the essential qualities of a good global optimization test problem was given. Some well known benchmark test problems were presented and used for the evaluation process. A novel hybridization approach that combines an evolutionary algorithm with a local search method was proposed and evaluated via series of numerical experiments. The performance of the proposed hybrid algorithm was compared to that of the standard evolutionary algorithm and another based on evolutionary strategy. The next chapter will recap, look into detailed plans for further work and finally conclude the report.

\singlespacing
\bibliographystyle{ieeetr}




\chapter{Discussion and Conclusions}

\section{Development of Ideas}

Robust optimization requires systems that can not only widely explore the problem space (such as global algorithms like EAs), but also effectively exploit the high quality regions. However, the struggle to establish \textit{optimum} balance between exploration and exploitation of the problem search space with a single algorithmic framework remains rather far away from been realistic. Thus, the use of hybrid algorithms has gradually become a widely acceptable practice.

Investigations revealed that \cite{Raidl2003paper92} hybridization of EAs with local search methods popularly known as \textit{memetic} algorithms often yield systems that are more robust than a mere combination of approximate algorithms. The proposal made in chapter 5 is an attempt to take a step further in this direction.

It was noticed that the challenges in designing good and generic hybrid algorithms go much beyond the choice of the individual algorithms. And if given the required attention, improving the methodologies in which these algorithms individually operate and interact could certainly play a major role towards the overall success of the hybrid system.

It thus became apparent that there is a great deal of open research in this direction. Any further investigation with the aim of; sorting out the key features of the individual optimization algorithms that could be enhanced or added (as treated in chapter \ref{ECAlgoChap}); determining how best to improve the individual algorithms for the benefit of the hybrid system (which was the focus of chapters \ref{ConvergenceAnalysisChap} and \ref{LocalSearchAlgoChap}); and, designing an effective hybrid methodology (as investigated at in chapter \ref{ProposedHybridAlgoChap}) could lead to significant progress.

\section{Discussion}

Several convergence detection methods are available in the literature most of which are based on some distance measures such as the Hamming distance.  In a somewhat radical perspective, rather than directly monitoring the similarity among the solutions, the proposed convergence measure is designed to detect convergence by assessing the extent to which evolutionary forces continue to effect changes on the solution set.  This novel heuristic for adaptive convergence detection in evolutionary algorithm was developed in chapter \ref{ConvergenceAnalysisChap}. It was built based on the principle of Price's theorem \cite{Frank1997} in quantitative genetics. Our investigations reveal that convergence of evolution can effectively be measured via monitoring the effect of genetic operators (specifically, the crossover operator) on fitness progress in a population. Empirical results have shown that the value of the proposed convergence threshold parameter is more sensitive to the settings of the crossover and mutation rates but not population size. Based on the standard EA parameter settings that are empirically and theoretically proven by many researchers \cite{DeJong1975,Goldberg1989paper99,Goldberg1985GABookref93}, our investigation suggests that the new convergence threshold parameter could be bounded in $[0.001, 0.01]$ for crossover and mutation rates of $P_{c}=0.7$ to $1.0$ and $P_{m}=1/l$ or $0.01$ respectively; given the empirical investigation we have undertaken. Further investigations could provide more definitive evidence regarding any possible correlation between other EA parameters and this newly proposed convergence threshold parameter.

Turning to the local search method, gradient based local search algorithms are known for their suitability in deriving local optimal solution from virtually any given solution point. However, these algorithms rely on information about the slope of the problem under consideration to estimate their descent directions. Therefore, successful evaluations of search directions usually incur expensive derivative computations. Although $2^{\textrm{nd}}$ order gradient based algorithms such as Newton methods are among the most efficient local search algorithms, they require evaluation of $2^{\textrm{nd}}$ derivatives (i.e., Hessians). Various approximation procedures conventionally used to circumvent exact Hessian evaluations usually succeed in reducing computational cost only to the detriment of efficiency and solution quality. The proposed vector form of forward accumulation of derivatives (chapter 4) is an automatic differentiation method that facilitates evaluations of exact derivatives (including Hessians) of any differentiable function at reduced computational cost. Hence, the resulting local search algorithm (SQP) proposed in chapter \ref{LocalSearchAlgoChap} enjoys two major benefits: \textit{first}, it remains a true second order method taking Newton search directions and having greater chances of taking full step sizes (i.e. step length $\alpha=1$). \textit{Second}, it escapes the unavoidable restarts encountered when approximate Hessians are used for evaluation of the search directions. Thus, the proposed SQP algorithm remains a full $2^{\textrm{nd}}$ order globally convergent Newton method that enjoys quadratic rate of convergence.

Amongst others, the proposed way of hybridizing evolutionary computation (EC) algorithm with the sequential quadratic programming (SQP) local search method (chapter \ref{ProposedHybridAlgoChap}) would lessen the strong dependence of global search methods on a particular parameter setting for each problem category. In other words, the proposed method is designed such that it will operate optimally on wide range of problems without the need for further parameter fine tuning. The evaluation of the results show that compared to other state-of-the-art methods, the proposed system seems to be robust on medium to large scale multimodal, non-separable, non-symmetric, multidimensional problems having low to moderate levels of dispersion.  Moreover, the method can yield high quality solutions for small scale problems, albeit the required computational effort could be considerably high. Therefore, a tradeoff is necessary between high solution quality and available computation.

\section{Conclusions}

This report has investigated the reasons for the recent wide acceptance of hybrid algorithms in the field of global optimization. Particularly, the use of evolutionary algorithms together with gradient based local search method has been thoroughly examined. Then, a hybrid system composing of an evolutionary algorithm and the sequential quadratic programming algorithm has been proposed.

Prior to that an investigation on the parameterization aspect of the evolutionary algorithm was conducted and a new convergence threshold parameter was proposed for the EAs. Appropriate bound for the new convergence parameter was determined with regards to standard settings for other important EA parameters like crossover and mutation types and rates.

Thereafter, a Newton based local search algorithm (SQP) was modified to use interior point method (IPM) instead of the conventional active set method (ASM)\footnote{Active set methods are convex optimization techniques that solves quadratic programming subproblems by respectively categorizing the equality constraints in the feasible region as active and the inequality constraints as inactive sets.} for solving its quadratic programming subproblems. Unlike in ASM approach where the search follows the boundary of the feasible region, IPM progresses by searching through the interior of the feasible region. Thus, IPM aids the proposed SQP algorithm to converge to the local optimal solution in remarkably few steps. Moreover, the SQP algorithm is designed to utilize exact Hessians obtained via vectorized forward mode of automatic differentiation to derive its search directions. These upgrades boost the effectiveness of the search directions and step sizes taken by the SQP local algorithm and ultimately improves its overall convergence characteristics.

Returning to the hypotheses presented at the beginning of this study, based on the preliminary results obtained so far, it seems justifiable to infer that a skilful hybridization of evolutionary algorithms with a suitable local search method could yield a robust and efficient means of solving wide range of global optimization problems. 

\section{Research Contributions}

This study provides new understandings of the concepts and challenges behind global optimization approaches. Thus far, the following additions are made to the fast growing literature in the areas of local, global and hybrid optimization.

\begin{enumerate}
	\item \textbf{An Adaptive Elitist Strategy:} The broad survey conducted on various aspects of evolutionary algorithms gives valuable insight into the key parameterization issues upon which development of successful global optimization methods relied. Most importantly, it leads to the development of new ideas that give birth to the proposed new replacement technique (adaptive elitism, see section \ref{ProposedAdaptElitism}). The adaptive elitist strategy will aid efficient exploitation of the promising areas of the search space without compromising exploration of other potentially viable regions.
	\item \textbf{A Convergence Detection via Monitoring Crossover:} Using extended Price's theorem \cite{Frank1997}, a critical convergence analysis of evolutionary algorithm that aids understanding of the interactions among genetic operators in an evolutionary search was conducted. As a result, a visual means for investigating the individual roles of genetic operators on fitness progress was developed. It thus became possible to assess and utilize the effect of crossover operator on fitness progress as a means of automatic convergence detection in evolutionary algorithms (see section \ref{AutoConvergDetcSec}).
	\item \textbf{A Vectorized Forward Accumulation AD:} The gradient based local search algorithms, particularly the standard sequential quadratic programming (SQP) algorithm were thoroughly analyzed. Then, the convergence characteristics of the SQP algorithm were upgraded to that of a $2^{\textrm{nd}}$ order algorithm that searches by taking full Newton steps. The improvement was achieved via an automatic differentiation (AD) tool for evaluation of accurate derivatives. Built based on a vectorized forward accumulation method (section \ref{PropsedMatlabADImplementationSec}), the AD tool allows evaluation of the entire derivatives and Hessians in a single forward sweep. Consequently, accurate derivatives (to machine precision) for every differentiable function can be obtained at a computational cost comparable to that of evaluating the function itself.
	\item \textbf{A Validation Strategy:} As a first attempt, we combine the global and local algorithms in a collaborative manner and the design seems to greatly enhance the balance in exploration and exploitation of search space which is a necessary ingredient for robust optimization. Nonetheless, the novel validation routine, amongst others (see section \ref{ProposedHybridAlgoSec}), has also contributed to the performance improvement achieved by the hybrid system.
\end{enumerate}

\section{Further work}

The results of the investigations undertaken have so far supported the core objective of this work. They have, in essence, opened several avenues for further investigation. Therefore, the other key objectives of this work can in no way be met without broadening the research in the following areas:

\begin{enumerate}
	\item \textbf{Sensitivity Analysis:} Further sensitivity analysis on the newly proposed automatic convergence detection parameter is necessary to determine all sorts of parameter settings where its application will be suitable in addition to the standard EA parameter settings (i.e. population size, crossover and mutation) explored so far. We will need to investigate if there is any correlation between the measure of the effect of the crossover operator, the population's fitness progress and the similarity among the candidate solutions.
	\item \textbf{Hamming distance $\delta$:} Hamming distance between individual solutions should be estimated and the plots for the delta $(\delta)$ changes in the Hamming distance can be compared side-by-side to the Price's plots so as to validate the current findings. Use of cross correlation measurements could aid verify the hypothesis that fitness progress, delta changes in Hamming distance and the proposed measure of crossover's effect on the fitness all enable evaluation of convergence in evolution. The cross correlation measurements/graphs may yield useful insights as to the degree to which these measures are correlated to convergence.
	\item \textbf{Other Seeding Potentials:} Pertaining to the newly introduced validation routine, which is invoked at the end of every run of the local algorithm, the reported improvement is as a result of using single copies of the best solution and its inverted version to seed the creation of a new population which will mainly consist of random individuals. Further empirical experimentation might reveal whether or not changing the proportion of the seed (i.e. using several copies of the best solution and its mutated versions) will have any positive impact on the overall gain derived from the validation procedure.
	\item \textbf{Extension to Algorithmic Portfolio:} It would be interesting to assess the possibility of extending the proposed hybridization system to mimic the design of portfolios of algorithms \cite{Gomes1997paper78}. Thus, instead of forcing a halt at the end of every validation routine, the system might be left to decide whether to embark on a fresh run of the global and local algorithms. Although from the design of the proposed hybrid system the schedule for the portfolios can be self-adapted, the implications of utilizing best so far solutions as seeds and the appropriate proportion of the seeds is subject to further investigations.
	\item \textbf{Representation Sensitivity Considerations:} The EC algorithm employed in this investigation utilizes binary representation of candidate solutions. Considering the implications of the role of representation on the overall performance of any evolutionary algorithm \cite{Atmar1994paper17}, a more equitable performance evaluation with other global optimization methods that are based on real valued representation may require restructuring the proposed method to use real-valued representation. This will aid further comparison with other well known hybrid metaheuristics.
	\item \textbf{Accounting for Constraints and Dynamism:} Further research needs to be done to enhance the proposed system to handle global optimization problems having not only parameter bounds, but also constraints. This is crucial to facilitate further investigations on its applicability to practical optimization problems that may not only be stochastic but also dynamic in nature (i.e., in dynamic optimization). 
	\item \textbf{Mapping to Feedback Control Systems:} In the long run and if time permits, the proposed hybrid system could be utilized in online and offline parameter optimization of feedback control systems (such as in parameter tuning of the proportional-integral-derivative (PID) controllers) which are widely used in industries.
	\item \textbf{Mapping to ANN Problem Domain:} Additionally, we could map the proposed hybrid optimization technique to evolving-then-training ANNs. The EC algorithm could conduct initial search of the weights state-space and then switch to the local algorithm which would then do the final tuning of the weights. The system could then be used for multi-mode learning in ANN.

\end{enumerate}


\singlespacing
\bibliographystyle{ieeetr}





\singlespacing
\bibliographystyle{ieeetr}


\bibliography{GAOptimLibrary,HybridGALibrary,LocalOptimLibrary}

\end{document}